\documentclass[11pt]{wlscirep}
\usepackage[utf8]{inputenc}
\usepackage[T1]{fontenc}
\usepackage{booktabs}
\usepackage{rotating}
\usepackage{marvosym}
\usepackage{lineno}
\usepackage{cleveref}
\usepackage{multirow}
\usepackage{placeins}
\usepackage{fancyhdr}

\usepackage{bibunits} % <-- 添加这一行
\usepackage{graphicx} % 
\usepackage{adjustbox}
\usepackage{makecell}
\usepackage{graphicx}      % 用于插入图片
\usepackage{subcaption}    % 用于创建子图环境
\usepackage{xcolor}  % Add this line to include the xcolor package
\usepackage{longtable}  % 用于创建可以跨页的长表格
\usepackage{siunitx}    % 用于精确格式化和对齐数字列
\usepackage{graphicx}
\usepackage{helvet}
\usepackage{authblk}
\usepackage{hyperref}
\usepackage{amsmath} 
\usepackage{amssymb} 
\usepackage{orcidlink} 
\usepackage{hyphenat}
\usepackage{booktabs}
\usepackage{multirow}
\usepackage{colortbl}
\usepackage{adjustbox}

\usepackage{booktabs}
\usepackage{tabularx}
\usepackage{adjustbox}
\usepackage{geometry}
\usepackage{xcolor}
\usepackage{longtable}
\usepackage{array}
\usepackage{booktabs}
\usepackage{geometry}
\usepackage{adjustbox}
\usepackage{ltablex} % longtable + tabularx
\usepackage{booktabs} % 更漂亮的表格线
\usepackage{geometry} % 页边距控制
\usepackage{array} % 自定义列
\usepackage{ragged2e} % 自动换行对齐
\usepackage{sansmath}
\usepackage{caption}

\defaultbibliographystyle{naturemag-doi} 
\defaultbibliography{main} % <-- 替换成你的 .bib 文件名 (main)

\definecolor{ColorProprietary}{HTML}{d6e6ff}
\definecolor{ColorGeneral}{HTML}{ffffea}
\definecolor{ColorMedical}{HTML}{f2eaf7}
\usepackage{enumitem} 
\definecolor{YesGreen}{RGB}{34,139,34}
\definecolor{NoRed}{RGB}{178,34,34}
\definecolor{InDomainBlue}{RGB}{30,144,255}
\definecolor{OODOrange}{RGB}{255,140,0}

\newcolumntype{Y}{>{\RaggedRight\arraybackslash}X}

\definecolor{SyntheticColor}{HTML}{2A9D8F} % Teal
\definecolor{PublicColor}{HTML}{E76F51}  % Burnt Orange
\definecolor{HeaderColor}{HTML}{4A5568}
\definecolor{SubtotalColor}{gray}{0.9}
\definecolor{MedText}{HTML}{E76F51}
\definecolor{MedMulti}{HTML}{E76F51}
\definecolor{GenMulti}{HTML}{2A9D8F}
\definecolor{GenText}{HTML}{2A9D8F}
\definecolor{HeaderColor}{HTML}{1A2B3A}
\definecolor{SubtotalColor}{HTML}{E8E8E8}
\definecolor{Stage0Color}{HTML}{0077B6}

\crefname{figure}{Fig.}{Figs.} % 单数Fig., 复数Figs.
\crefname{table}{Tab.}{Tabs.}
\crefname{section}{Sec.}{Secs.}
\crefname{extendedfigure}{Extended Fig.}{Extended Data Figs.}
\crefname{extendedtable}{Extended Tab.}{Extended Data Tabs.}

\title{Hulu-Med: Scaling Medical Vision-Language Model for Universal Visual Understanding}
\title{Hulu-Med: An Open-source and Open-data Medical Generalist  Vision-Language Model for Universal Visual-Text Understanding}
\title{Hulu-Med: A Transparent Generalist Model towards Holistic Medical Vision-Language Understanding}
\author[1,2]{Songtao Jiang}
\author[1,2]{Yuan Wang}
\author[3]{Sibo Song}
\author[1,2]{Tianxiang Hu}
\author[1]{Chenyi Zhou}
\author[4]{Bin Pu}
\author[1]{Yan Zhang}
\author[3]{Zhibo Yang}
\author[5]{Yang Feng}
\author[6]{Joey Tianyi Zhou}
\author[7]{Jin Hao}
\author[8]{Zijian Chen}
\author[9]{Ruijia Wu}
\author[10]{Tao Tang} 
\author[11]{Junhui Lv} 
\author[11]{Hongxia Xu} 
\author[1]{Hongwei Wang}
\author[1]{Jun Xiao}
\author[2]{Bin Feng}
\author[2]{Fudong Zhu}
\author[4]{Kenli Li} 
\author[12, \Letter]{Weidi Xie}
\author[8, \Letter]{Jimeng Sun}
\author[1,11, \Letter]{Jian Wu} 
\author[1,2,11, \Letter]{Zuozhu Liu}

\affil[1]{College of Computer Science and Technology, Zhejiang University-University of Illinois Urbana-Champaign Institute, Zhejiang University, Hangzhou 310027, Zhejiang, China.}
\affil[2]{Stomatology Hospital, School of Stomatology, Zhejiang University School of Medicine, Zhejiang University, Hangzhou 310016, Zhejiang, China.}
\affil[3]{Alibaba Inc, Hangzhou 310023, China.}
\affil[4]{College of Computer Science and Electronic Engineering, Hunan University, Changsha 410082, China.}
\affil[5]{Angelalign Technology Inc., Shanghai 200082, China.}
\affil[6]{CFAR \& IHPC, Agency for Science, Technology and Research, 138632, Singapore.}
\affil[7]{Department of Orthodontics, Shanghai Ninth People's Hospital, College of Stomatology, Shanghai Jiao Tong University School of Medicine, Shanghai 200011, China.}
\affil[8]{Siebel School of Computing and Data Science, University of Illinois Urbana-Champaign, Urbana, IL 61801, USA.}
\affil[9]{Antai College of Economics and Management, Shanghai Jiao Tong University , Shanghai 200030, China.}
\affil[10]{China Mobile Group Zhejiang Company Limited, Hangzhou 310016, Zhejiang, China.}
\affil[11]{Zhejiang Key Laboratory of Medical Imaging Artificial Intelligence, Haining 314400, Zhejiang, China.}
\affil[12]{School of Artificial Intelligence, Shanghai Jiao Tong University, Shanghai 200030, China.}

\affil[\Letter]{Corresponding author}

\begin{abstract}

Real-world clinical decision-making requires integrating heterogeneous data, 
including medical text, 2D images, 3D volumes, and videos, 
while existing AI systems fail to unify all these signals, limiting their utility.
In this paper, we introduce \textbf{Hulu-Med}, 
a transparent, generalist medical Vision–Language Model (VLM) designed to unify language-only, 2D/3D vision–language, and video understanding within a single architecture.
Hulu-Med is trained on a curated corpus of 16.7 million samples, comprising exclusively public or synthetic data, spanning 12 major anatomical systems and 14 medical imaging modalities. 
Hulu-Med employs a medical-aware token-reduction strategy that prunes redundant visual tokens, achieving up to a 55\% reduction for 3D and video inputs, improving cross-modal efficiency, and enabling training at 7B–32B parameter scales in approximately 4,000–40,000 GPU hours. Across 30 public in-domain and out-of-domain medical benchmarks—covering text reasoning, visual question answering, report generation, multilingual dialogue, video understanding, and rare disease diagnosis—\textbf{Hulu-Med} surpasses existing open-source models on 27 of 30 benchmarks and outperforms proprietary systems such as GPT-4o on 16 benchmarks. 
Despite being a VLM, Hulu-Med outperforms GPT-4o and matches GPT-o1 on the text-only HealthBench. For the first time in the community, we provide a fully transparent, reproducible and cost-effective pipeline for holistic medical vision-language understanding by releasing our end-to-end data curation, training procedures, and model parameters. Code and models are available at \href{https://github.com/ZJUI-AI4H/Hulu-Med}{https://github.com/ZJUI-AI4H/Hulu-Med}.
\end{abstract}

\begin{document}
% \linenumbers 
\flushbottom
\maketitle

 \thispagestyle{empty}
\pagestyle{fancy}
\fancyhf{}
\fancyfoot[R]{\scriptsize \thepage}
\newpage

% ============== 主论文部分 ================
\begin{bibunit}

\section*{Introduction}

Clinical decision-making is inherently multimodal, requiring the integration of diverse data sources such as free-text notes, structured records, and visual inputs—including 2D images, 3D scans, and videos—spanning a patient’s care journey~\cite{moor2023foundation,topol2019high}. While clinicians must synthesize these signals over time, current healthcare AI systems remain fragmented and task-specific, leading to inefficiencies and missed cross-modal insights~\cite{rajkomar2019machine,lambin2017radiomics}. A generalist medical vision–language model (VLM) capable of processing both visual and textual data natively could streamline workflows, reduce diagnostic errors, and extend the reach of multimodal analysis~\cite{moor2023foundation,tu2025towards,tiu2022expert,maier2017surgical,mckinney2020international,vasey2022reporting}.

Despite the rapid advances in general-purpose VLMs, their deployment in the medical sector is impeded by substantial limitations in technical unification and transparency~\cite{brown2020language,radford2019language,wang2024qwen2,comanici2025gemini,lillava,luunified,jin2024chat}. Current models, which rely heavily on instruction tuning to bridge the pretrained vision and language encoders with image-text pairs, excel in specific tasks but fail to comprehensively cover the full spectrum of clinical needs, for example, 
language-based tasks such as diagnosis and medical reporting, 
2D/3D image analysis in radiology and pathology, complex video analysis in endoscopy and surgery~\cite{lu2024multimodal,yan2025multimodal,zhang2024generalist,qiu2024development,li2023llava,wu2025towards,chen2024huatuogpt,li2024llavasurg}. Moreover, the development of these specialized medical AI tools is often opaque, dominated by proprietary datasets and non-transparent curation processes, which hinder community scrutiny, reproducibility, and, critically, clinical adoption~\cite{kim2025transparency,mahmood2025benchmarking,shick2024transparency,saenz2024maida,guo2025deepseek}. To overcome these limitations, a new approach is necessary: a unified, multimodal architecture that integrates text, 2D/3D images, and video data with full transparency in its development and training processes~\cite{lillava,zhang2024generalist,ma2024evolution,huang2024position}.

In this paper, we present \textbf{Hulu-Med}, a generalist medical VLM that achieves holistic multimodal coverage by unifying text, 2D images, 3D volumes, and video understanding within a single architecture~(\textcolor{blue}{Fig.~\ref{fig:overview}a}, \textcolor{blue}{Extended Tab.~\ref{tab:data_composition}-\ref{tab:benchmark}}). Hulu-Med is built upon three core design principles: \textbf{holistic coverage}, \textbf{efficiency at scale}, 
and \textbf{end-to-end transparency}. Hulu-Med is trained on a comprehensive corpus of 16.7 million samples which are sourced from publicly available data and synthesized by us, spanning 12 anatomical systems and 14 medical imaging modalities, including CT, MRI, X-ray, endoscopy, and histopathology, as shown in \textcolor{blue}{Fig.~\ref{fig:overview}b}, \textcolor{blue}{Extended Tab.~\ref{tab:modality_coverage}}. The corpus integrates 9M medical multimodal samples, 4.9 M medical text samples, 1.3M general multimodal samples, and 1.5M general text samples, balancing domain specialization with broad linguistic and visual competence (\textcolor{blue}{Extended Tab.~\ref{tab:stage0_data}-\ref{tab:final_definitive_verified}}). We provide access to our complete pipeline-including a 3-stage training regime (medical alignment, continuous medical pretraining, and mixed modality finetuning), detailed data curation documentation, training code, evaluation scripts, and model weights—to ensure full reproducibility and auditability. This transparent approach facilitates understanding from 2D to complex 3D and video data, enhancing textual reasoning capabilities (\textcolor{blue}{Extended Fig.~\ref{fig:data_syn}-\ref{fig:framework}, Extended Tab.~\ref{tab:prompt_templates}-\ref{tab:dataset_license_full}, Methods})~\cite{kim2025transparency}.

Technically, the model introduces a novel stack that integrates a SigLIP-based vision encoder with two-dimensional rotary position embeddings (2D RoPE)—seamlessly extended to 3D and video data—and a large language model~(LLM) decoder through a unified patch-based encoding strategy. This strategy treats visual patches as universal input units across all modalities, eliminating the need for separate encoders. Our architecture supports arbitrary medical image resolutions and spatio-temporal understanding, enabling the flexible combination of LLMs and ViTs for continuous pretraining and eliminating the need for pretrained VLMs. Moreover, we develop a progressive three-stage training curriculum, which scales understanding from 2D images to complex 3D volumes and videos, demonstrates emergent cross-modal generalization.
Crucially, to manage the high computational demands typical of medical data processing, we introduce a medical-aware token reduction mechanism that reduces visual tokens by approximately $55\%$. This significant reduction enhances computational efficiency and supports extended context processing across 3D volumes and videos without compromising data fidelity. Ultimately, Hulu-Med scales efficiently from 7 billion to 32 billion parameters, maintaining exceptional performance with low compute requirements—ranging from 4,000 to 40,000 GPU hours for the largest model variant. This scalability makes advanced medical VLM capabilities more accessible to the broader research community.

We thoroughly evaluate Hulu-Med across 30 public medical benchmarks covering diverse tasks, including language-based reasoning, 2D/3D/video question answering (VQA), medical report generation (MRG), etc. 
Hulu-Med consistently delivers \textbf{leading performance} among open medical and general VLMs, and is highly competitive with proprietary systems across diverse tasks. It demonstrates robust generalization across modalities, anatomies, and task formats, particularly those requiring intensive knowledge and textual reasoning, showing robust generalization on challenging benchmarks for multilingual understanding, rare disease diagnosis, and clinical dialogue. 
This high performance, achieved with calibrated compute and full transparency, validates the feasibility of a unified, accessible, and high-performing medical generalist model. 

Our work makes three key contributions:
\begin{enumerate}
    \item \textbf{Architectural Unification:} We introduce the first medical VLM capable of natively processing text, 2D images, 3D volumes, and video within a single, unified architecture, solving the long-standing challenge of modality-specific encoders.
    
    \item \textbf{Efficiency and Transparency:} We provide a fully transparent and highly efficient training pipeline, demonstrating the feasibility of scaling generalist medical VLMs (7B–32B parameters) within realistic, accessible compute budgets. 
    Additionally, the privacy and copyright concerns inherent in proprietary systems can be mitigated, empowering the development of customized trustworthy models.
    
    \item \textbf{Strong Performance:}  We demonstrate strong performance in 30 public medical benchmarks, encompassing language-based reasoning, 2D/3D/video question answering, report generation, etc. To our knowledge, this represents the first systematic benchmarking of a medical VLM at this scale and diversity. This work marks a significant step toward a holistic understanding of medical data and fostering greater accessibility for the broader community.
    
    % We demonstrate strong performance across 30 public medical benchmarks, spanning across language-based reasoning, 2D/3D/video question answering, report generation, etc (Actually we are the first to do this diverse evaluation). This work represents a step towards holistic understanding of medical data and fostering greater accessibility to the broader community. 

\end{enumerate}

\section*{Results}

\subsection*{Overview of Hulu-Med}

Hulu-Med refers to a versatile multimodal model adept at navigating an extensive range of medical tasks, encompassing from answering complex language-only queries, to performing sophisticated analyses involving 2D and 3D medical images, to interpreting long-duration medical videos.

\paragraph{Problem Formulation.} 
Hulu-Med is designed to be a generalist medical VLM capable of processing heterogeneous inputs and generating textual responses for diverse clinical tasks (\textcolor{blue}{Fig.~\ref{fig:overview}a}). Formally, given a textual instruction $\mathbf{t}$ and optional visual input $\mathbf{v} \in \{\mathbf{v}_{2D}, \mathbf{v}_{3D}, \mathbf{v}_{video},\varnothing \}$ (where $\mathbf{v}$ can be a 2D image, 3D volume, video sequence, or absent), 
the model generates a textual response $\mathbf{y}$ as: 
$$\mathbf{y} = \Phi([g ( f_v(\mathbf{v}));  f_t(\mathbf{t})]),$$
where $f_v(\cdot)$ and $f_t(\cdot)$ denote the visual encoder and text tokenizer, respectively, which transform inputs into variable-length sequences of visual tokens $ f_v(\mathbf{v}) \in \mathbb{R}^{N_v \times d_v}$ and text tokens $f_t(\mathbf{t}) \in \mathbb{R}^{N_t \times d_t}$, where $N_v$ and $N_t$ represent the number of visual and text tokens, $d_v$ and $d_t$ denote the visual and text feature dimensions, respectively. A projection layer $g(\cdot)$ aligns visual features to the LLM's embedding space: $ g( f_v(\mathbf{v})) \in \mathbb{R}^{N_v \times d}$, where $d$ is the hidden dimension of the language model. 
The notation $[\cdot; \cdot]$ indicates concatenation along the sequence dimension. The language model decoder $\Phi(\cdot)$ then autoregressively generates responses conditioned on the concatenated token sequence. Critically, when visual input is absent ($\mathbf{v} = \varnothing$), the model seamlessly operates in text-only mode, where text tokens are directly fed into the LLM backbone for autoregressive generation: $\mathbf{y} = \Phi(f_t(\mathbf{t}))$ (\textcolor{blue}{Fig.~\ref{fig:overview}b}).

% $$\mathbf{y} = \Phi([\mathbf{T}_v; \mathbf{T}_t])$$
% where $f_v(\cdot)$ and $f_t(\cdot)$ denote the visual encoder and text tokenizer, respectively, which transform inputs into variable-length sequences of visual tokens $\mathbf{T}_v = f_v(\mathbf{v}) \in \mathbb{R}^{N_v \times d_v}$ and text tokens $\mathbf{T}_t = f_t(\mathbf{t}) \in \mathbb{R}^{N_t \times d_t}$, where $N_v$ and $N_t$ represent the number of visual and text tokens, $d_v$ and $d_t$ denote the visual and text feature dimensions, respectively. A projection layer $g(\cdot)$ aligns visual features to the LLM's embedding space: $\mathbf{T}'_v = g(\mathbf{T}_v) \in \mathbb{R}^{N_v \times d}$, where $d$ is the hidden dimension of the language model. 
% The notation $[\cdot; \cdot]$ indicates concatenation along the sequence dimension. The language model decoder $\Phi(\cdot)$ then autoregressively generates responses conditioned on the concatenated token sequence: $\mathbf{y} = \Phi([\mathbf{T}'_v; \mathbf{T}_t])$. Critically, when visual input is absent ($\mathbf{v} = \varnothing$), the model seamlessly operates in text-only mode, where text tokens are directly fed into the LLM backbone for autoregressive generation: $\mathbf{y} = \Phi(\mathbf{T}_t)$ (\textcolor{blue}{Fig.~\ref{fig:overview}b}).
% \weidi{this paragraph is confusing, is $\Phi$ operating on $T$ or $T'$ ?}

% \weidi{I would directly write in this way, then say, $f_v(\mathbf{v}) \in \mathbb{R}^{N_v \times d_v}$ denotes .... }
% $$\mathbf{y} = \Phi([g \circ f_v(\mathbf{v});  f_t(\mathbf{t})])$$

This unified formulation enables Hulu-Med to flexibly handle various input configurations: 
(i) text-only queries for medical knowledge reasoning and clinical dialogue; 
(ii) vision-language tasks, for example, medical images/videos with textual instructions for visual question answering, report generation, and diagnostic reasoning; and (iii) interleaved multimodal inputs, 
where diverse visual modalities (2D images, 3D volumes, videos) and text can be arbitrarily interspersed within a single context. 
The model's architecture, as detailed in \textcolor{blue}{Extended Fig.~\ref{fig:framework}}, encapsulates the ability to integrate and transition between various data modalities and analytical demands within a single, unified system, supporting diverse clinical applications from radiological interpretation to surgical video analysis.

\paragraph{Training Dataset.}
To support broad generalist capabilities and promote transparency, we curated an unprecedented multimodal dataset of 16.7 million samples—the largest publicly available to our knowledge—compiled from open sources and augmented with synthetic data (see \textcolor{blue}{Extended Fig.\ref{fig:data_syn}; Extended Tab.~\ref{tab:stage0_data}-\ref{tab:final_definitive_verified}}). The corpus comprises 9 million multimodal medical samples, 4.9 million medical text QA pairs, 1.3 million multimodal general samples, and 1.5 million general text QA pairs. The medical subset spans 12 major anatomical systems ({\hyperref[fig:overview]{\Cref{fig:overview}c}}) and 14 distinct imaging modalities ({\hyperref[fig:overview]{\Cref{fig:overview}d}}), covering more than 60 specific types and a broad range of clinical tasks~(\textcolor{blue}{Extended Tab.~\ref{tab:modality_coverage}}).

Raw public datasets typically suffer from limited modality coverage, suboptimal alignment between text and visual data, and pronounced long-tail distributions, all of which can hinder both model performance and generalizability. To address these challenges, we developed \textbf{five dedicated synthesis pipelines} to generate high-quality, instruction-aligned visual–text pairs. 
These pipelines encompass: 
\textbf{(i)} rewriting brief captions into detailed descriptions; 
\textbf{(ii)} generating novel, long-form medical image captions; 
\textbf{(iii)} constructing medical VQA pairs; 
\textbf{(iv)} producing multilingual Chain-of-Thought (CoT) reasoning data; 
and \textbf{(v)} annotating surgical videos~(\textcolor{blue}{Methods}, \textcolor{blue}{Extended Fig.~\ref{fig:data_syn}}). The resulting synthetic data proved instrumental in the multi-stage training of Hulu-Med.

% zuozhu stops here. Oct 21st. 

\paragraph{Model Architecture and Training.} 
Hulu-Med consists of four core components: 
a rotary position-adaptive visual transformer (ViT) encoder, a multimodal projector, a language tokenizer, and an LLM decoder~(\textcolor{blue}{Fig.~\ref{fig:overview}b, Extended Fig.~\ref{fig:framework}}; for details, see \textcolor{blue}{Methods}). 

For visual encoding, we adopt image patch as a universal processing unit, that allows 2D images, 3D volumes, and videos to be handled as variable-length patch sequences by a single encoder, obviating the need for modality-specific architectures. In particular, we adapt a pre-trained SigLIP model, enhancing it with 2D RoPE to extend compatibility with 3D and video data~\cite{zhai2023sigmoid,su2024roformer}.
To demonstrate scalability and address varying computational constraints, we developed three model variants: \textbf{Hulu-Med-7B}, \textbf{Hulu-Med-14B}, and \textbf{Hulu-Med-32B}. Their respective LLM decoders were continuously pretrained from Qwen2.5-7B, Qwen3-14B, and Qwen2.5-32B~\cite{yang2025qwen3}. 
To efficiently manage the substantial computational demands imposed by long sequences of 3D and video patches, we devised a medical-aware token reduction strategy that enables holistic and efficient training.

Hulu-Med is trained using a progressive, three-stage curriculum~(\textcolor{blue}{Fig.~\ref{fig:overview}e}). 
At Stage-1, the model establishes the medical vision–language alignment, 
with only the visual encoder and multimodal projector being trained on concise 2D medical image–caption pairs~(\textcolor{blue}{Extended Tab.~\ref{tab:stage0_data}}). 
At Stage-2, Hulu-Med undergoes continuous training on large-scale, long-form medical image–caption pairs (2D images), supplemented by mixed general data~(\textcolor{blue}{Extended Tab.~\ref{tab:stage2_data_corrected}}). 
Stage-3 comprises comprehensive finetuning on an extensive multimodal dataset encompassing both medical and general domains, spanning diverse downstream tasks across text, 2D, 3D, and video modalities~(\textcolor{blue}{Extended Tab.~\ref{tab:final_definitive_verified}}). 
Throughout Stage-2 and Stage-3, all model parameters—including the LLM decoder, visual encoder, and multimodal projector—remain fully trainable to maximize performance and generalization. This training curriculum leverages the abundance of 2D data to cultivate robust visual representations, enabling the model to excel on complex 3D and video tasks with comparatively less specialized data.

\paragraph{Evaluation Protocols.} 
We conducted a comprehensive evaluation of Hulu-Med across 30 diverse benchmarks spanning language, 2D and 3D images, and video modalities~(\textcolor{blue}{Fig.~\ref{fig:overview}f}), rigorously assessing both in-distribution (ID) and out-of-distribution (OOD) tasks to evaluate generalization. 
Our comparisons encompass 46 state-of-the-art models, including leading proprietary systems~({\em e.g.}, GPT-4.1, Claude Sonnet 4, Gemini-2.5-Flash), large-scale general-purpose vision–language models~({\em e.g.}, Qwen2.5VL-7B/72B, InternVL3-8B/38B)~\cite{bai2025qwen2,zhu2025internvl3}, medical generalist VLMs ({\em e.g.}, Lingshu-7B/32B, MedGemma-4B, HuatuoGPT-V-7B/34B), and specialized medical foundation models ({\em e.g.}, M3D series, RadFM, Surgical-LLaVA)~\cite{bai2024m3d,jin2024surgical,xu2025lingshu,sellergren2025medgemma,wu2025towards,bai2024m3d}.

To further probe real-world utility, we also include more comprehensive evaluation for language-only tasks, including multilingual medical understanding~(MMedBench), rare disease diagnosis~(RareBench), and multi-turn clinical dialogue (HealthBench)~\cite{qiu2024towards,chen2024rarebench,arora2025healthbench}. 
Standard evaluation metrics were employed for each benchmark and task, 
with detailed protocols provided in the \textcolor{blue}{Methods}.

\subsection*{Evaluation on 2D Medical Vision–Language Understanding}

We systematically evaluated Hulu-Med’s 2D medical image understanding across 11 established benchmarks, comprising seven medical VQA datasets, three MRG benchmarks, and the MedMNIST classification task. 
Across these benchmarks, Hulu-Med surpassed all open-source models (medical or general) in 10 of 11 benchmarks. 
It also outperforms the leading proprietary models in 8 benchmarks~(\textcolor{blue}{Tab.~\ref{tab:mm_result}, Fig.~\ref{fig:2d_results}}). 

% all Hulu-Med variants established new state-of-the-art performance, \weidi{the sota performance only refers to the comparison with open-source models, right ? be clear on that, as it seems not always leading proprietary models} consistently surpassing both leading proprietary systems ({\em e.g.}, GPT-4.1) and open-source general and medical VLMs~(\textcolor{blue}{Tab.~\ref{tab:mm_result}}). 
% %These results demonstrate that systematically leveraging large-scale public and synthetic data can deliver superior models, even substantially outperform larger proprietary models.
%\weidi{can we make this section more clear, imo, we are evaluating on 11 benchmarks, across VQA, classification, MRG, then each of the following paragraphs should discuss one type of tasks.}

In particular, the VQA suite encompasses multi-modal understanding (OmniMedVQA, PMC-VQA)~\cite{hu2024omnimedvqa,zhang2023pmc}, modality-specific reasoning~(VQA-RAD, SLAKE, PathVQA), advanced clinical reasoning~(MedXQA), and knowledge-intensive tasks~(MMMU-Med)\cite{lau2018dataset,liu2021slake,he2020pathvqa,zuo2025medxpertqa}, as shown in \textcolor{blue}{Tab.~\ref{tab:mm_result}}. 
Hulu-Med-7B/32B set new state-of-the-art performance on the multi-modal understanding and modality-specific reasoning benchmarks, spanning ID and OOD tasks. On MedXQA, Hulu-Med-7B/32B outperformed all open-source VLMs of comparable scale (both below and above 10B parameters), though they remained behind proprietary models ({\em e.g.}, 34\% of Hulu-Med-32B versus 45.2\% of GPT-4.1). 
%\weidi{for example, .....}.
We attribute this performance gap primarily to the text-based reasoning requirements of MedXQA, which favor models with more powerful LLMs. Similarly, on the knowledge-intensive benchmark~(MMMU-Med), Hulu-Med surpassed other medical VLMs and most generalist models, although trailed the strongest open model InternVL-38B, as this benchmark requires extra capabilities like optical character recognition (OCR), which is not a central focus of our architecture. 
% \weidi{While on multimodal understanding and modality-specific reasoning benchmarks, ....... }
To validate the robustness of these findings, we performed statistical significance tests across three independent runs of Hulu-Med-7B, which demonstrated consistently superior performance~(p < 0.001 for PMC-VQA, VQA-RAD, and MedXQA; p < 0.05 for OmniMedVQA, SLAKE, and PathVQA; \textcolor{blue}{Extended Fig.~\ref{extendedfigure:mm_bench_pvalue}}).

On MRG, we assessed Hulu-Med on three standard benchmarks—MIMIC-CXR, CheXpert, 
and IU X-ray—using both conventional natural language metrics (BLEU, ROUGE, METEOR) and the clinically oriented RaTEScore~\cite{demner2015preparing,zhao2024ratescore,irvin2019chexpert,johnson2019mimic} (\textcolor{blue}{Fig.~\ref{fig:2d_results}a-b}). All Hulu-Med variants established new state-of-the-art results. 
Notably, as shown in \textcolor{blue}{Fig.~\ref{fig:2d_results}b},
Hulu-Med-7B achieved a RaTEScore of 57.0 on MIMIC-CXR, substantially outperforming MedGemma-4B/27B (RaTEScore 51.3). This improvement is clinically meaningful, as MedGemma’s score corresponded to 81\% of reports leading to the same or superior clinical decisions as original reports, as assessed by board-certified radiologists~\cite{sellergren2025medgemma}. 
Our results further reveal that larger model size does not guarantee superior MRG performance: Hulu-Med-7B occasionally surpassed its 32B counterpart, mirroring trends observed with MedGemma. This underscores that domain-specific pretraining is paramount for specialized tasks such as MRG, reaffirming the necessity of dedicated medical VLMs.

Hulu-Med’s 2D medical image understanding was further validated on the MedMNIST-2D benchmark, which spans seven distinct domains~\cite{yang2021medmnist}. 
Hulu-Med achieved a leading average accuracy exceeding 85\%, dramatically outperforming all baselines—including proprietary models such as GPT-4o, which attained less than 40\%~(\textcolor{blue}{Fig.~\ref{fig:2d_results}c}). Hulu-Med’s robust performance across diverse data modalities and task types—including binary, multi-class, and multi-label classification—further highlights the critical importance of domain-specific medical training.

\subsection*{Evaluation on 3D Medical Vision–Language Understanding}

We systematically evaluated Hulu-Med’s 3D medical image understanding on VQA and MRG benchmarks, including M3D, 3D-RAD, and AMOS-MM ~\cite{ji2022amos,bai2024m3d,gai20253d}. For comprehensive comparison, we benchmarked our model against both medical foundation models specialized for 3D data~({\em e.g.}, RadFM, M3D-Llama2/Phi/Mistral) 
and adapted generalist models ({\em e.g.}, Lingshu, Qwen2.5-VL). As these generalist models do not natively support 3D volumetric data, we enabled 3D evaluation by slicing each volume into a sequence of images, thus treating it as a multi-image task~(\textcolor{blue}{Methods}).

On 3D VQA tasks, Hulu-Med achieved state-of-the-art performance for both open- and closed-ended question answering~(\textcolor{blue}{Fig.~\ref{fig:3d_video_bench}a-b}). 
On the M3D benchmark, which assesses anatomical understanding, Hulu-Med outperformed all specialized 3D models and general-purpose VLMs~(\textcolor{blue}{Fig.~\ref{fig:3d_video_bench}a}). Hulu-Med further excelled at complex 3D reasoning tasks on the 3D-RAD benchmark~(\textcolor{blue}{Fig.~\ref{fig:3d_video_bench}b}). The performance advantage was particularly pronounced on challenging tasks requiring multi-step inference, such as problems involving biomarker characteristics ({\em e.g.}, size, thickness, and shape) and static/longitudinal temporal diagnosis. Hulu-Med-7B exceeded the best baseline by 22.8\% on longitudinal temporal diagnosis, a task demanding comprehensive understanding of disease progression across multiple time points. The consistent, superior performance of Hulu-Med across diverse 3D tasks underscores the effectiveness of a unified architecture for the nuanced interpretation of volumetric medical data.

For 3D MRG tasks on the AMOS-MM benchmark, all Hulu-Med variants demonstrated leading performance on conventional natural language generation metrics~(BLEU, ROUGE-L) and exhibited clear superiority on the clinically oriented RaTEScore, underscoring the model’s ability to generate comprehensive and clinically accurate radiology reports from volumetric scans (\textcolor{blue}{Fig.~\ref{fig:3d_video_bench}c}, \textcolor{blue}{Extended Fig.~\ref{fig:cas4}}, \textcolor{blue}{Extended Fig.~\ref{fig:cas9}}). Their performance on METEOR was also competitive to models trained for MRG on this dataset, validating the effectiveness of our unified training approach.

% \zuozhu{the order of fig 3b and 3c is changed, please revise the fig accordingly. \st{(Done,now 3d rad is 3b, amos is 3c)}}
% \weidi{The first paragraph in this section has already discussed 3D-RAD ? why again ? merge them ?}

% On 3D-RAD, our model demonstrated robust capabilities in core 3D understanding tasks—such as anomaly detection, effectively interpreting complex volumetric structures across diverse anatomical regions.

\subsection*{Evaluation on Medical Video Benchmarks}

We evaluated Hulu-Med variants on multi-frame temporal reasoning and surgical video analysis~(\textcolor{blue}{Methods}).
As shown in~\textcolor{blue}{Fig.~\ref{fig:3d_video_bench}d},
for multi-frame temporal reasoning, we assess zero-shot performance on MedFrameQA~\cite{yu2025medframeqa}—{\em i.e.}, without any task-specific training or fine-tuning. In this OOD setting, Hulu-Med markedly outperforms the leading proprietary models reported in the original study, achieving higher accuracy and lower variance as the number of frames increases (\textcolor{blue}{Extended Tab.~\ref{tab:model_accuracy_updated}}). 
This stability under growing temporal complexity underscores robust temporal reasoning. The radar plot further illustrates the unified understanding of Hulu-Med across modalities.

In specialized surgical video benchmarks: Cholec80-VQA~\cite{nwoye2023cholectriplet2021}, EndoVis18-VQA~\cite{allan20202018roboticscenesegmentation}, and PSI-AVA-VQA~\cite{10.1007/978-3-031-16449-1_42}, Hulu-Med was compared with proprietary systems, general and medical VLMs, and surgical video foundation models~(\textcolor{blue}{Fig.~\ref{fig:3d_video_bench}e}). It achieves superior accuracy and recall to the video foundation models on Cholec80-VQA and EndoVis18-VQA, and delivers competitive results on PSI-AVA-VQA, given that several baselines are tailored for video data. For VLM baselines lacking reported quantitative metrics, we used ChatGPT-4o-latest as an automated judge, 
Hulu-Med consistently surpasses all baselines across the three benchmarks~(\textcolor{blue}{Fig.~\ref{fig:3d_video_bench}f}).

Additionally, the SurgeryVideoQA~\cite{thapa2025well} presents a distinct OOD challenge, drawing on educational video content that integrates medical images, diagrams, and narrative explanations—unlike conventional surgical footage. Here, Hulu-Med-32B led all open-source models, achieving a score of 30.1\% and outperforming other specialized medical VLMs such as Lingshu-32B~(29.9\%), while proprietary models like GPT-4o attained the highest score at 44.8\%~(\textcolor{blue}{Fig.~\ref{fig:3d_video_bench}g}). Overall, Hulu-Med demonstrated competitive performance in this complex, educationally-focused benchmark while maintaining strong advantages on specialized surgical video analysis.

\subsection*{Evaluation on Medical Text-Only Benchmarks}

We evaluated Hulu-Med on eight medical text understanding benchmarks, assessing capabilities in complex reasoning, textual comprehension, and medical examination~\cite{wang2024mmlu,zuo2025medxpertqa,chen2025benchmarking,du2025supergpqa,jin2019pubmedqa,pal2022medmcqa,hendrycks2020measuring}~(\textcolor{blue}{Tab.~\ref{tab:text_result}}). 
Additionally, we assessed the model's generalization capability on challenging real-world tasks including multilingual reasoning (MMedBench), rare disease diagnosis (RareBench), and realistic clinical dialogues (HealthBench)~\cite{qiu2024towards,chen2024rarebench,arora2025healthbench}. 

%Hulu-Med-7B and Hulu-Med-32B outperformed all general and medical VLMs of comparable scale across all eight standard benchmarks.

\vspace{3pt}\noindent\textbf{State-of-the-art performance among open-source models.} Hulu-Med delivered substantial gains over prior open-source systems. Against Lingshu-7B—the strongest medical VLM baseline under 10B parameters—Hulu-Med-7B improves by 10.2 points on MMLU-Pro-Med, 3.1 on MedXQA, 5.3 on Medbullets, 4.8 on SGPQA, 11.7 on MedMCQA, 10.2 on MedQA, and 5.0 on MMLU-Med. For larger models, Hulu-Med-32B similarly surpasses Lingshu-32B with gains of 2.7 points on MMLU-Pro-Med, 3.4 on Medbullets, 6.7 on MedMCQA, and 5.7 on MedQA. Notably, Hulu-Med is competitive even against larger open-source models: Hulu-Med-7B attains an average accuracy of 58.9\%, exceeding InternVL3-8B (52.9\%) by 6.0 points, while Hulu-Med-32B reaches 65.9\%, outperforming InternVL3-38B (60.1\%) by 5.8 points. Statistical testing on three independent runs confirms these improvements; Hulu-Med demonstrated superior performance ($p < 0.001$) on seven benchmarks (\textcolor{blue}{Extended Fig.~\ref{extendedfigure:text_bench_pvalue}}).

\vspace{3pt}\noindent\textbf{Competitive with proprietary models.} 
Hulu-Med-32B attains state-of-the-art performance on PubMedQA, surpassing proprietary systems including Claude Sonnet 4, DeepSeek-V3, GPT-4.1, and Gemini 2.5 Flash. This strength in biomedical literature comprehension likely reflects the extensive PubMed content used in continuous pretraining (\textcolor{blue}{Extended Tab.~\ref{tab:data_composition}}). On complex reasoning benchmarks such as MMLU-Pro-Med, Hulu-Med-32B outperforms Gemini 2.5 Flash and substantially narrows the gap to top-tier models, including Claude Sonnet 4 and GPT-4.1. More broadly, with an average accuracy of 65.9\% across eight benchmarks, Hulu-Med-32B exceeds DeepSeek-V3 (59.8\%) by 6.1 points and approaches o3-mini (67.9\%) within 2.0 points, demonstrating competitive performance overall against proprietary systems.

\vspace{3pt}\noindent\textbf{Performance gap on complex reasoning tasks.} On challenging reasoning-intensive benchmarks requiring multi-step clinical reasoning, Hulu-Med substantially narrowed but did not fully close the gap with top-tier proprietary models. On MedXpertQA, Hulu-Med-32B outperformed DeepSeek-V3 but trailed Claude Sonnet 4 by 9.4 points. Similarly, on Medbullets, while exceeding DeepSeek-V3, Hulu-Med-32B remained 11.4 points behind Claude Sonnet 4. The performance gap was most pronounced on MedQA, with an 11.7 point difference from Claude Sonnet 4. These gaps primarily reflect the advantages of larger model capacity and more sophisticated reasoning capabilities in proprietary systems.

\vspace{3pt}\noindent\textbf{Scaling effects with model size.} We observed substantial performance improvements with increasing model scale from 7B to 32B parameters across all benchmarks (\textcolor{blue}{Tab.~\ref{tab:text_result}}). The gains were particularly pronounced on reasoning-intensive tasks, with improvements of 4.6 percentage points on MedXQA, 7.3 points on Medbullets, 10.7 points on SGPQA, and 6.9 points on MedQA. This scaling effect indicates that textual reasoning capabilities strongly depend on the capacity of the underlying language model, explaining the remaining performance disparity with even larger-scale proprietary models and suggesting that further scaling combined with enhanced reasoning-focused training represents a promising direction for closing this gap.

\vspace{3pt}\noindent\textbf{Strong generalization to real-world scenarios.}
Beyond standard benchmarks, we evaluate Hulu-Med on three challenging, real-world tasks. On MMedBench, a multilingual benchmark spanning six languages (English, Chinese, Japanese, French, Russian, Spanish), Hulu-Med-32B attains 75.13\% average accuracy, surpassing GPT-4 (74.27\%), while Hulu-Med-7B (67.81\%) exceeds the prior state-of-the-art, MMed-Llama 3-8B (67.75\%)~(\textcolor{blue}{Fig.~\ref{fig:generalization_bench}a}). 
With CoT prompting, Hulu-Med-32B-Thinking reaches 78.41\%, outperforming all tested proprietary models, including Gemini 2.5 Flash (77.63\%) and Claude Sonnet 4 (76.04\%), underscoring strong multilingual reasoning and practical utility.

On HealthBench—which assesses clinical conversation quality and safety using physician-defined rubrics—Hulu-Med-32B achieves an overall score of 41.6, outperforming general-purpose models such as GPT-4o (32.0) and matching GPT-o1 (\textcolor{blue}{Fig.~\ref{fig:generalization_bench}b}, \textcolor{blue}{Extended Tab.~\ref{tab:bench_all}}). Hulu-Med consistently surpasses specialized medical VLMs across all seven conversational themes; notably, Hulu-Med-7B (38.3) more than doubles the scores of HuatuoGPT-Vision-34B (17.2) and Lingshu-7B (15.9).

Finally, on RareBench Task 4, containing 1,114 rare-disease cases (\textcolor{blue}{Fig.~\ref{fig:generalization_bench}c})—standard Hulu-Med-7B/32B performs modestly on this OOD, long-tail task. However, with explicit CoT prompting (``Please reason step by step'') (\textcolor{blue}{Extended Fig.~\ref{tab:prompt_templates}}), Hulu-Med surpasses all proprietary models in 7 of 8 testing scenarios, indicating strong latent reasoning for complex, low-prevalence conditions.

\subsection*{Analysis of Model Design and Data Curation}

To elucidate the principles underlying Hulu-Med's performance, 
we conducted a series of analytical studies on model architecture, data composition, data enhancements and training efficiency.

\vspace{3pt}\noindent \textbf{Expert vs.~generalist model.}
To assess the value of a unified multimodal architecture versus specialized systems, we trained five modality-specific models (ultrasound, Optical Coherence Tomography (OCT), fundus, microscopy, dermoscopy). Hulu-Med, trained on a mixed dataset spanning these and additional modalities, consistently outperforms all specialized counterparts~(\textcolor{blue}{Fig.~\ref{fig:efficiency}a}). These results indicate that a single unified model not only achieves broad competency but also delivers superior cross-modal transfer and understanding.

\vspace{3pt}\noindent \textbf{Data scale and mixture strategy.}
We next investigated the impact of data scale and mixture composition. Performance on both text and multimodal tasks improved monotonically with training volume, consistent with established scaling laws in LLMs and VLMs~\cite{hoffmann2022training,henighan2020scaling} (\textcolor{blue}{Fig.~\ref{fig:efficiency}b}). 
%\textcolor{blue}{DONE}\weidi{cite general scaling law and paper}. 
Ablations further show that two axes of diversity: domain (medical vs.~general) and modality (text-only vs.~multimodal), are critical: removing any single component degrades performance (\textcolor{blue}{Fig.~\ref{fig:efficiency}c}). 
% Exploring mixing ratios along these axes, we find that a 3:1 medical-to-general balance and a 1:1 text-to-multimodal balance yield the best results
Further analysis of the mixing ratios showed optimal performance at a 3:1 medical-to-general and a 1:1 text-to-multimodal balance(\textcolor{blue}{Fig.~\ref{fig:efficiency}d-e}). 
These findings indicate that performance depends not only on scale but also on domain specificity and modality composition, providing practical guidance for future data mixture design.

\vspace{3pt}\noindent\textbf{Effectiveness of synthetic data.}
Adding synthetically generated long captions increases accuracy on OmniMedVQA (\textcolor{blue}{Fig.~\ref{fig:efficiency}f}). Likewise, incorporating generated CoT data boosts both textual and multimodal reasoning (MedXpert-Text, MedXpert-Multimodal), with especially pronounced gains in multimodal settings (\textcolor{blue}{Fig.~\ref{fig:efficiency}g-h}). These results indicate that synthetic data provides valuable supervision for complex tasks beyond what public datasets afford.

\vspace{3pt}\noindent\textbf{Token reduction for efficiency.}
We further validated the medical-aware token reduction strategy for 3D and video inputs. With approximately 55\% fewer visual tokens, performance on surgical video benchmarks was effectively unchanged, and degradation on 3D benchmarks (M3D, 3D-RAD) was minimal (\textcolor{blue}{Fig.~\ref{fig:efficiency}i, Extended Fig.~\ref{fig:enxtended3_aba_efficiency}a-b}). This efficiency was critical for practical training: Hulu-Med-7B/32B required roughly 4,100/38,000 GPU-hours on 80G memory GPUs (\textcolor{blue}{Extended Fig.~\ref{fig:enxtended3_aba_efficiency}c}), improving accessibility for both academic and industrial use.

% zuozhu continue here

\section*{Discussion}

%\noindent \textbf{Overview and contribution.}

We present Hulu-Med, the first transparent generalist medical VLM for holistic understanding of medical text, 2D/3D images, and videos. Hulu-Med is trained in a three-stage pipeline with the ever-largest open corpus of 16.7M samples, showing a cost-effective strategy with affordable computing. Hulu-Med's leading performance across 30 benchmarks, combined with its openness and unified understanding, establishes it as a foundational resource for medical AI.

\vspace{3pt}\noindent \textbf{Superior performance through guaranteed transparency and reproducibility.}
Hulu-Med was developed exclusively from open-access datasets using a fully transparent workflow, from data curation to model release. Its training corpus is unprecedented in scale and diversity, integrating clinical and literature data across broader modalities than existing medical VLMs, such as RadFM and Lingshu ~\cite{li2023llava,chen2024huatuogpt,wu2025towards,xu2025lingshu} (\textcolor{blue}{Extended Tab.~\ref{tab:data_composition}}). Hulu-Med's superior performance demonstrates that the systematic consolidation of public data is a viable pathway to state-of-the-art medical VLMs. To ensure full reproducibility and establish a trustworthy foundation for clinical application, we publicly release all data pipelines, training code, and model parameters (\textcolor{blue}{Extended Tab.~\ref{tab:stage2_data_corrected}-~\ref{tab:model_accuracy_updated}, Methods}). This open approach mitigates the privacy and copyright risks inherent in proprietary models and private data~\cite{4-lu2024multimodal,price2019privacy}.

% zuozhu continue here

\vspace{3pt}\noindent \textbf{Technical novelty for holistic medical understanding.} Hulu-Med introduces the first unified architecture that concurrently achieves state-of-the-art performance across medical text, 2D/3D images and video understanding tasks, as demonstrated by leading results on 30 diverse benchmarks~\cite{xu2025qwen3,lillava,guo2025seed1} (\textcolor{blue}{Extended Tab.~\ref{tab:data_composition}}). This is enabled by three technical designs. First, a unified visual encoding strategy treats all visual patches as universal input units, employing 2D RoPE to natively represent multiple modalities and resolutions as dynamic, variable-length sequences within a single encoder (\textcolor{blue}{Methods}). Second, an adaptive token-reduction scheme—combining bilinear interpolation for shorter sequences with medical-aware pruning for longer ones—reduces 3D and video tokens by approximately 55\% with minimal accuracy loss, ensuring computational efficiency~(\textcolor{blue}{Fig.~\ref{fig:efficiency}i, Extended Fig.~\ref{fig:enxtended3_aba_efficiency}}). Third, a progressive training curriculum that establishes robust 2D understanding before introducing 3D and video modalities outperforms mixed-modality training, providing a more effective learning pathway~(\textcolor{blue}{Extended Fig.~\ref{fig:enxtended3_aba_training_strategy}a}).
Together, these designs offer a cost-efficient solution for holistic medical understanding, overcoming critical bottlenecks of limited data and high computational cost~\cite{saenz2024maida,acosta2022multimodal}.

\vspace{3pt}\noindent \textbf{Scalable training recipe.}
Hulu-Med offers a practical recipe for multimodal medical generalist models, grounded in extensive analysis. 
Our decoupled architecture—integrating a separate visual encoder with an LLM decoder—provides critical flexibility over methods that merely fine-tune general-purpose VLMs. This modular design facilitates tailored model configurations, allowing state-of-the-art components like Qwen LLMs and specialized ViTs to be matched to specific clinical needs (\textcolor{blue}{Extended Fig.~\ref{fig:abalation_llmbackbone}})~\cite{yang2025qwen3,guo2025seed1}. The progressive strategy yields emergent generalization: Hulu-Med-Image-7B, trained solely on 2D data, extrapolates strongly to 3D and video (\textcolor{blue}{Extended Fig.~\ref{fig:enxtended3_aba_training_strategy}b-c}), while adding 3D/video in the final stage further improves 2D performance (\textcolor{blue}{Extended Fig.~\ref{fig:enxtended3_aba_training_strategy}d}). 
%We show that scale, quality, and diversity of data are crucial for high performance~(\textcolor{blue}{Fig.~\ref{fig:efficiency}}); 
Detailed benchmarks further validate consistent gains from scaling both data and model size within the same family~(\textcolor{blue}{Extended Figs.~\ref{fig:rebalancing}, \ref{fig:abalation_llmsize}}). Together, our approach ensures scalable training and strengthens domain-specific reasoning.

\noindent \textbf{Real-world clinical utility.}
Hulu-Med demonstrates substantial real-world clinical utility through superior performance on widely used clinical benchmarks~(\textcolor{blue}{Tab.~\ref{tab:mm_result}-\ref{tab:text_result}, Fig.~\ref{fig:efficiency}}) and scenario-specific evaluations including HealthBench, MMedBench, and RareBench~\cite{qiu2024towards,chen2024rarebench,arora2025healthbench} (\textcolor{blue}{Fig.~\ref{fig:generalization_bench}}). 
Extensive case studies across text, 2D/3D images, and video modalities confirm robust understanding and reasoning capabilities~(\textcolor{blue}{Extended Fig.~\ref{fig:case1}-\ref{fig:cas2}}). 
In 2D/3D medical report generation, Hulu-Med produces more accurate findings with fewer hallucinations than Med-Gemma~(\textcolor{blue}{Extended Fig.~\ref{fig:cas3}–\ref{fig:cas4}}), exhibits step-by-step diagnostic reasoning~(\textcolor{blue}{Extended Fig.~\ref{fig:cas5}–\ref{fig:cas6}}), and generates detailed surgical video captions~(\textcolor{blue}{Extended Fig.~\ref{fig:cas7}}). 
Notably, it efficiently processes hour-long videos while pruning 55\% of tokens and reducing GPU memory by 43\%~(\textcolor{blue}{Extended Fig.~\ref{fig:cas8}}), demonstrating suitability for resource-constrained environments. 
The model shows strong multilingual capability, rare disease diagnostics~(\textcolor{blue}{Extended Fig.~\ref{fig:cas9}–\ref{fig:cas10}}), and robust multi-turn dialogue performance~(\textcolor{blue}{Extended Fig.~\ref{fig:cas15}-~\ref{fig:cas17}}). 
Without reinforcement learning, it performs reflective reasoning with self-correction when prompted~(\textcolor{blue}{Extended Fig.~\ref{fig:cas11}–\ref{fig:cas14}}), particularly valuable for complex, low-prevalence conditions. 
With its transparent pipeline and cost-effective training, Hulu-Med provides a credible foundation for real-world clinical deployment.

\vspace{3pt} \noindent \textbf{Limitations and future directions.}
Hulu-Med has limitations that chart a course for future work. First, the model's input is presently restricted to medical text and visual data. A critical next frontier involves integrating genomic and molecular data to enable a truly multi-scale understanding of disease, moving towards predictive and personalized medicine. Furthermore, the landscape of public data remains underutilized; a more exhaustive aggregation of global datasets represents a straightforward path to further scale model performance and generalizability. Second, the reasoning capabilities of medical VLMs are not fully realized. Future work could leverage advanced training paradigms, such as large-scale reinforcement learning on diverse long CoT and long-horizon data to better capture the nuanced logic of clinical reasoning. This would enhance both the interpretability and reliability. Concurrently, establishing efficient continual pretraining mechanisms will be crucial for the model to remain current with the rapid evolution of medical knowledge. Finally, although Hulu-Med has been comprehensively evaluated on established benchmarks, further integration into specialist models and multi-agent systems for  clinical validation is of high necessity to ensure safe and effective workflows.

\clearpage
\addtocontents{toc}{\protect\setcounter{tocdepth}{-10}}
% \bibliography{main}
\putbib 

 % <-- 在主论文参考文献后结束
\addtocontents{toc}{\protect\setcounter{tocdepth}{1}}
\clearpage

% \begin{figure*}[!t]
% \centering
% \includegraphics[width=1\linewidth]{Figure_1.pdf}
% \caption{\textbf{Overview of the Hulu-Med architecture, data composition, and training strategy.} 
%     \textbf{a}, The unified encoder-decoder framework of Hulu-Med, capable of processing a wide array of medical inputs including 2D/3D images, videos, and complex textual queries. The model employs a Medical-Aware Token Pruning mechanism to efficiently process diverse inputs and supports a comprehensive suite of downstream tasks, from report generation and visual QA to complex reasoning and multi-turn diagnosis.
%     \textbf{b}, Quantitative distribution of the pretraining data across major body organs. This comprehensive coverage is foundational for building the model's ability to generalize across diverse anatomical regions.
%     \textbf{c}, Modality distribution of the pretraining image corpus. The chart highlights the extensive diversity of our dataset, which encompasses 29 distinct medical imaging modalities ranging from common radiological scans to specialized microscopic and endoscopic imagery.
%     \textbf{d}, The multi-stage data curation strategy for model training. The process begins with a large set of short medical image-caption pairs, progressively enriched with long-form medical captions, high-quality general domain data, and a massive-scale mixture of text and diverse multimodal task data in subsequent stages.
% }
%     \label{fig:overview}
% \end{figure*}

\begin{figure*}[!p]
    \centering
    \includegraphics[width=1\linewidth]{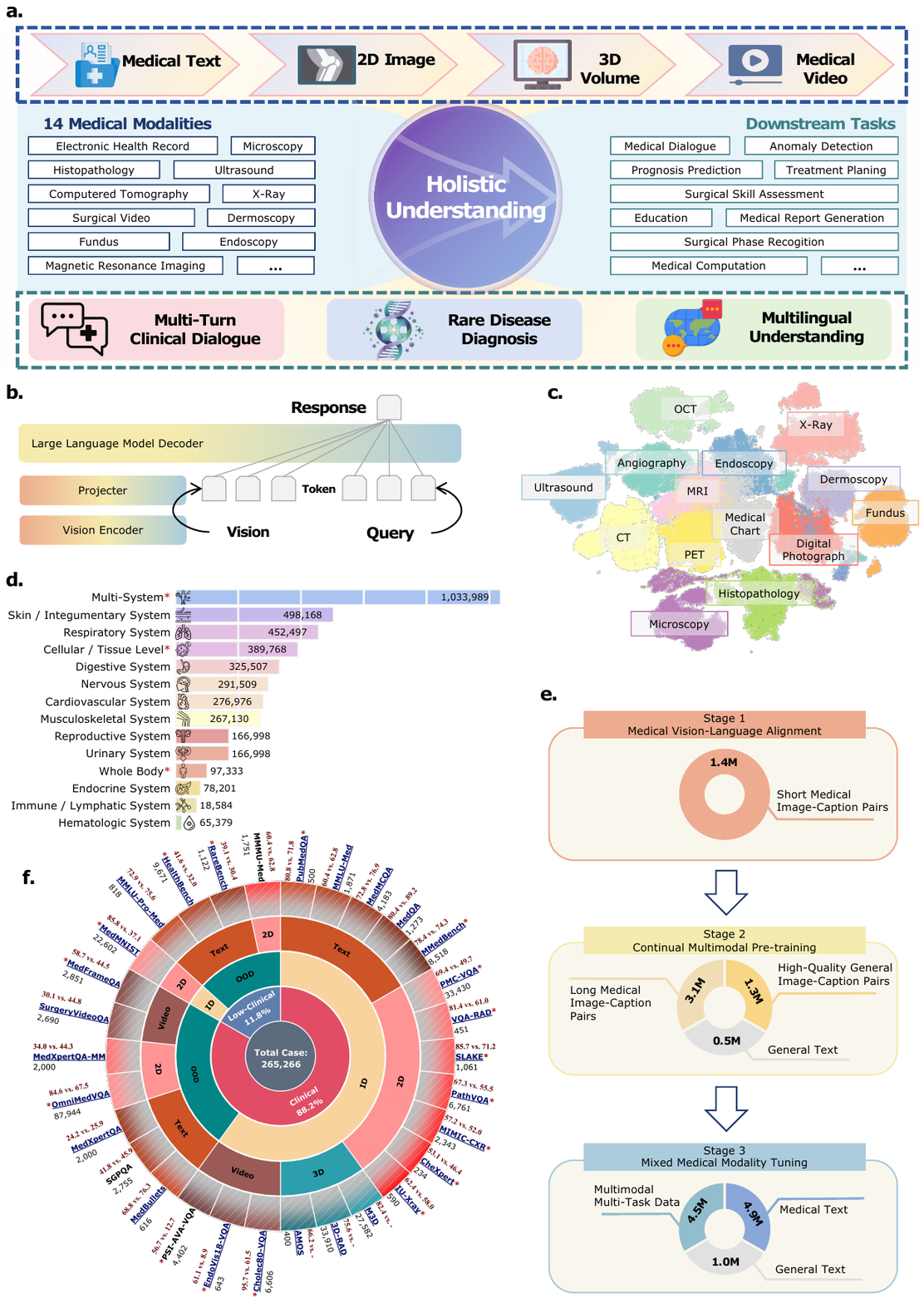} 
\end{figure*}

\clearpage

\begin{center}
    \parbox{0.95\textwidth}{ 
        \captionof{figure}{\textbf{Overview of the Hulu-Med architecture, data composition, training strategy and Evaluation.}
        \textbf{a}, The model's unified architecture is designed to holistically process a diverse spectrum of medical inputs—spanning text, 2D images, 3D volumes, and video—to support a wide array of downstream clinical tasks.
         \textbf{b}, A schematic of the core model components, including the vision encoder, projector, and LLM decoder, is presented.
         \textbf{c,d}, The training corpus spans 12 major anatomical systems and 14 imaging modalities, forming a comprehensive basis for the model’s generalist reasoning and understanding capabilities.
        \textbf{e}, The progressive three-stage training curriculum is detailed, beginning with foundational vision-language alignment, advancing to continual pre-training with enriched data, and culminating in mixed-modality instruction tuning. 
        \textbf{f}, Comprehensive benchmark evaluation overview. This visualization encompasses 30 medical benchmarks, hierarchically organized by clinical domain and task complexity, totaling 265,266 test cases. The scores in the outermost ring denote performance of Hulu-Med (left score) vs. GPT-4o (right score). The \textcolor{red}{red asterisks (*)} indicates 17 benchmarks where Hulu-Med-32B outperforms GPT-4o; \textcolor{blue}{\underline{blue underlined text}} indicates 27 benchmarks where Hulu-Med attains the best open-source performance; and \textbf{black bolded text} indicates  3 benchmarks where Hulu-Med is the second-best. GPT-4o's performance is marked as "–" for 3D benchmarks due to its inability to process 3D inputs and limitations on the number of image inputs.}
        \label{fig:overview} 
    }
\end{center}

\clearpage

\begin{table*}[t]
\centering
\sansmath
\caption{Performance comparison among three categories of VLMs (Proprietary, General-purpose, and Medical) on 2D medical VQA benchmarks, with benchmarks categorized by task type. \textbf{bold} and \underline{underline} scores indicate the best and second-best medical VLMs in two subgroups with different model sizes, respectively.}
\label{tab:mm_result}
\adjustbox{max width=\textwidth}{
% 使用siunitx进行小数点对齐
\begin{tabular}{@{}l|cc|ccc|c|c@{}}
\toprule
% --- Header Row 1: Benchmark Categories ---
&
\multicolumn{2}{c|}{\makecell[c]{\textbf{Multi-modality} \\ \textbf{Benchmarks}}} &
\multicolumn{3}{c|}{\makecell[c]{\textbf{Specific-modality} \\ \textbf{Benchmarks}}} &
\multicolumn{1}{c|}{\makecell[c]{\textbf{Reasoning} \\ \textbf{Benchmark}}} &
\multicolumn{1}{c}{\makecell[c]{\textbf{Knowledge-intensive} \\ \textbf{Benchmark}}} \\
\cmidrule(lr){2-3} \cmidrule(lr){4-6} \cmidrule(lr){7-7} \cmidrule(lr){8-8}

% --- Header Row 2: Benchmark Names ---
\textbf{Models} &
{OM.VQA} & {PMCVQA} & {VQA-RAD} & {SLAKE} & {PathVQA} & {MedXQA} & {MMMU-Med} \\
\midrule
\multicolumn{8}{c}{\textit{Proprietary Models}} \\
\midrule
\rowcolor{ColorProprietary!30}
GPT-4.1          & 75.5 & 55.2 & 65.0 & 72.2 & 55.5 & 45.2 & 75.2  \\
\rowcolor{ColorProprietary!30}
GPT-4o  & 67.5 & 49.7& 61.0 &71.2  &55.5  &44.3  & 62.8  \\
\rowcolor{ColorProprietary!30}
Claude Sonnet 4  & 65.5 & 54.4 & 67.6 & 70.6 & 54.2 & 43.3 & 74.6 \\
\rowcolor{ColorProprietary!30}
Gemini-2.5-Flash & 71.0 & 55.4 & 68.5 & 75.8 & 55.4 & 52.8 & 76.9 \\
\midrule
\multicolumn{8}{c}{\textit{General-purpose Multimodal VLMs}} \\
\midrule
\addlinespace
\multicolumn{8}{l}{\textit{--- Models < 10B ---}} \\
\rowcolor{ColorGeneral!40}
Qwen2.5VL-7B    & 63.6 & 51.9 & 63.2 & 66.8 & 44.1 & 20.1 & 50.6 \\
\rowcolor{ColorGeneral!40}
Janus-Pro-7B    & 59.6 & 50.1 & 49.7 & 55.2 & 35.4 & 18.4 & 36.1 \\
\rowcolor{ColorGeneral!40}
InternVL2.5-8B  & 81.3 & 51.3 & 59.4 & 69.0 & 42.1 & 21.7 & 53.5 \\
\rowcolor{ColorGeneral!40}
InternVL3-8B    & 79.1 & 53.8 & 65.4 & 72.8 & 48.6 & 22.4 & 59.2 \\
\addlinespace
\multicolumn{8}{l}{\textit{--- Models > 10B ---}} \\
\rowcolor{ColorGeneral!40}
Llama3.2-11B    & 43.8 & 48.1 & 58.8 & 65.8 & 32.9 & 20.1 & 51.0 \\
\rowcolor{ColorGeneral!40}
InternVL3-14B   & 78.9 & 54.1 & 66.3 & 72.8 & 48.0 & 23.1 & 63.1 \\
\rowcolor{ColorGeneral!40}
Qwen2.5V-32B    & 68.2 & 54.5 & 71.8 & 71.2 & 41.9 & 25.2 & 59.6 \\
\rowcolor{ColorGeneral!40}
InternVL2.5-38B & 79.9 & 57.2 & 61.4 & 70.3 & 46.9 & 24.4 & 61.6 \\
\rowcolor{ColorGeneral!40}
InternVL3-38B   & 79.8 & 56.6 & 65.4 & 72.7 & 51.0 & 25.2 & 65.2 \\
\midrule
\multicolumn{8}{c}{\textit{Medical Multimodal VLMs}} \\
\midrule
\addlinespace
\multicolumn{8}{l}{\textit{--- Models < 10B ---}} \\
\rowcolor{ColorMedical!50}
BiomedGPT$^\heartsuit$ & 27.9 & 27.6 & 16.6 & 13.6 & 11.3 & {-}  & 24.9 \\
\rowcolor{ColorMedical!50}
Med-R1-2B$^\diamondsuit$ & {-}  & 47.4 & 39.0 & 54.5 & 15.3 & 21.1 & 34.8 \\
\rowcolor{ColorMedical!50}
MedVLM-R1-2B    & 77.6 & 48.8 & 49.2 & 56.3 & 36.0 & 21.4 & 35.2 \\
\rowcolor{ColorMedical!50}
HealthGPT-M3    & 71.5 & 55.4 & 56.8 & 70.8 & 55.4 & 22.4 & 42.8 \\
\rowcolor{ColorMedical!50}
BioMediX2-8B    & 66.0 & 41.8 & 55.7 & 54.1 & 34.6 & 21.9 & 39.8 \\
\rowcolor{ColorMedical!50}
LLaVA-Med-7B    & 34.8 & 22.7 & 46.6 & 51.9 & 35.2 & 20.8 & 28.1 \\
\rowcolor{ColorMedical!50}
MedGemma-4B-IT  & 70.7 & 49.2 & \underline{72.3} & 78.2 & 48.1 & 25.4 & 43.2 \\
\rowcolor{ColorMedical!50}
HuatuoGPT-V-7B  & 74.3 & 53.1 & 67.6 & 68.1 & 44.8 & 23.2 & \underline{49.8} \\
\rowcolor{ColorMedical!50}
Lingshu-7B$^\dagger$ & \underline{82.9} & \underline{56.3} & 67.9 & \underline{83.1} & \underline{61.9} & \underline{26.7} & {-} \\
% \rowcolor{ColorMedical!50}
% \textbf{Hulu-Med-4B}  & {81.6} & {64.6} & {60.1} & {80.1} & {56.6} & {26.5} & 50.1 \\
\rowcolor{ColorMedical!50}
\textbf{Hulu-Med-7B}  & \textbf{84.2} & \textbf{66.8} & \textbf{78.0} & \textbf{86.8} & \textbf{65.6} & \textbf{29.0} & \textbf{51.4} \\
% \rowcolor{ColorMedical!50}
% \textbf{Hulu-Med-8B}  & 84.0 & {68.0} & 74.9 & 85.0 & 63.9 & 28.3 & 52.7 \\
% \addlinespace
\multicolumn{8}{l}{\textit{--- Models > 10B ---}} \\
\rowcolor{ColorMedical!50}
HealthGPT-14B   & 75.2 & 56.4 & 65.0 & 66.1 & 56.7 & 24.7 & 49.6 \\
\rowcolor{ColorMedical!50}
HuatuoGPT-V-34B & 74.0 & 56.6 & 61.4 & 69.5 & 44.4 & 22.1 & 51.8 \\
\rowcolor{ColorMedical!50}
Lingshu-32B$^\dagger$ & {83.4} & {57.9} & \underline{76.7} & \textbf{86.7} & \underline{65.5} & \underline{30.9} & {-} \\
\rowcolor{ColorMedical!50}
MedDr-40B$^\heartsuit$ & 64.3 & 13.9 & 65.2 & 66.4 & 53.5 & {-} & 49.3 \\
\rowcolor{ColorMedical!50}
\textbf{Hulu-Med-14B}  & \underline{85.1} & \underline{68.9} & {76.1} & \underline{86.5} & {64.4} & {30.0} & \underline{54.8} \\
\rowcolor{ColorMedical!50}
\textbf{Hulu-Med-32B}  & \textbf{84.6} & \textbf{69.4} & \textbf{81.4} & {85.7} & \textbf{67.3} & \textbf{34.0} & \textbf{60.4} \\
\bottomrule
\addlinespace
\multicolumn{8}{l}{\footnotesize $^\diamondsuit$Med-R1 trained on OmniMedVQA test set. $^\heartsuit$No multi-image support. $^\dagger$Lingshu trained on MMMU-Med val set.} \\
\end{tabular}}
\end{table*}

%第四段的clinical findings，第五段case study，第六段limitation，conclusion

\begin{table}[t]
\centering
\sansmath % 在表格环境内激活无衬线数学字体
\caption{Performance comparison among three categories of VLMs (Proprietary, General-purpose, and Medical) on medical text benchmarks. Within each open-source medical VLM subgroup, \textbf{bold} and \underline{underline} scores indicate the best and second-best methods, respectively. Note that MedQA, MedXQA, and SGPQA denote MedQA-USMLE, MedXpertQA-Text, and SuperGPQA-Medical benchmarks, respectively.}
\label{tab:text_result}
\renewcommand{\arraystretch}{1.1}
\adjustbox{max width=1.0\textwidth}{
\begin{tabular}{lcccccccc}
\toprule
% --- Header Row 1: Benchmark Categories ---
&
\multicolumn{4}{|c|}{\makecell[c]{\textbf{Complex Reasoning} \\ \textbf{Benchmarks}}} &
\multicolumn{1}{c|}{\makecell[c]{\textbf{Text Understanding} \\ \textbf{Benchmark}}} &
\multicolumn{3}{c}{\makecell[c]{\textbf{Medical Exam} \\ \textbf{Benchmarks}}} \\
\cmidrule(lr){2-5} \cmidrule(lr){6-6} \cmidrule(lr){7-9}

\textbf{Models} &
\textbf{MMLU-Pro-Med} &
\textbf{MedXQA} &
\textbf{Medbullets} &
\textbf{SGPQA} &
\textbf{PubMedQA} &
\textbf{MedMCQA} &
\textbf{MedQA} &
\textbf{MMLU-Med} \\
\midrule

\multicolumn{9}{c}{\textit{Proprietary Models}}\\
\midrule
\rowcolor{ColorProprietary!30}
GPT-4.1           & 78.0 & 30.9 & 77.0 & 49.9 & 75.6 & 77.7 & 89.1 & 89.6 \\
\rowcolor{ColorProprietary!30}
o3-mini           & 78.1 & 35.4 & 83.7 & 50.1 & 73.6 & 60.6 & 74.5 & 87.0 \\
\rowcolor{ColorProprietary!30}
GPT-4o            & 75.6 & 25.9 & 76.3 & 45.9 & 71.8 & 76.9 & 89.2 & 88.2 \\
\rowcolor{ColorProprietary!30}
Claude Sonnet 4   & 79.5 & 33.6 & 80.2 & 56.3 & 78.6 & 79.3 & 92.1 & 91.3 \\
\rowcolor{ColorProprietary!30}
Gemini-2.5-Flash  & 70.0 & 35.6 & 77.6 & 53.3 & 73.8 & 73.6 & 91.2 & 84.2 \\
\rowcolor{ColorProprietary!30}
Deepseek-V3       & 74.6 & 20.0 & 48.4 & 32.1 & 77.7 & 88.0 & 51.0 & 86.5 \\
\midrule

\multicolumn{9}{c}{\textit{General-purpose Multimodal VLMs}}\\
\midrule
\addlinespace
\multicolumn{9}{l}{\textit{--- Models \textless\ 10B ---}} \\
\rowcolor{ColorGeneral!40}
Qwen2.5VL-7B      & 50.5 & 12.8 & 42.1 & 26.3 & {76.4} & 52.6 & 57.3 & 73.4 \\
\rowcolor{ColorGeneral!40}
Janus-Pro-7B      & 20.2 & 10.0 & 30.2 & 14.8 & 72.0 & 37.5 & 37.4 & 46.4 \\
\rowcolor{ColorGeneral!40}
InternVL2.5-8B    & 50.6 & 11.6 & 42.4 & 26.1 & {76.4} & 52.4 & 53.7 & 74.2 \\
\rowcolor{ColorGeneral!40}
InternVL3-8B      & {57.9} & {13.1} & {48.5} & {31.2} & 75.4 & {57.7} & {62.1} & {77.5} \\
\addlinespace
\multicolumn{9}{l}{\textit{--- Models > 10B ---}} \\
\rowcolor{ColorGeneral!40}
Qwen2.5VL-32B      &66.5 & {15.6} & 54.2 & 37.6 & 68.4 & 63.0 & 71.6 & 83.2 \\
\rowcolor{ColorGeneral!40}
InternVL3-14B      & 65.4  & 14.1 & 49.5 & 37.9 & {77.2} & 62.0 & 70.1 & 81.7 \\
\rowcolor{ColorGeneral!40}
InternVL2.5-38B    & 71.5 & 14.7 & {55.0} & {39.9} & 74.2 & {65.9} & {74.4} & {84.6} \\
\rowcolor{ColorGeneral!40}
InternVL3-38B      & 72.1 & {16.0} & {54.6} & {42.5} & 73.2 & {64.9} & {73.5} & {83.8} \\
\addlinespace
\midrule

\multicolumn{9}{c}{\textit{Medical Multimodal VLMs}} \\
\midrule
\addlinespace
\multicolumn{9}{l}{\textit{--- Models \textless\ 10B ---}} \\
\rowcolor{ColorMedical!50}
MedVLM-R1-2B      & 24.9 & 11.8 & 33.8 & 19.1 & 66.4 & 39.7 & 42.3 & 51.8 \\
\rowcolor{ColorMedical!50}
BioMediX2-8B      & 40.8 & 13.4 & 45.9 & 25.2 & 75.2 & 52.9 & 58.9 & 68.6 \\
\rowcolor{ColorMedical!50}
MedGemma-4B-IT    & 38.6 & 12.8 & 45.6 & 21.6 & 72.2 & 52.2 & 56.2 & 66.7 \\
\rowcolor{ColorMedical!50}
HealthGPT-M3      & 38.3 & 11.5 & 41.4 & 18.9 & 57.8 & 54.2 & 55.0 & 72.5 \\
\rowcolor{ColorMedical!50}
LLaVA-Med-7B      & 16.6 & 9.9  & 34.4 & 16.1 & 26.4 & 39.4 & 42.0 & 50.6 \\
\rowcolor{ColorMedical!50}
HuatuoGPT-V-7B    & 44.6 & 10.1 & 40.9 & 21.9 & 72.8 & 51.2 & 52.9 & 69.3 \\
\rowcolor{ColorMedical!50}
Lingshu-7B        & \underline{50.4} & \underline{16.5} & \underline{56.2} & \underline{26.3} & \underline{76.6} & \underline{55.9} & \underline{63.3} & \underline{74.5} \\
% \rowcolor{ColorMedical!50}
% \textbf{Hulu-Med-4B}  & {58.7} & {16.9} & {59.4} & {29.4} & 77.6 & {64.5} & {71.8} & {78.5} \\
\rowcolor{ColorMedical!50}
\textbf{Hulu-Med-7B}  & \textbf{60.6} & \textbf{19.6} & \textbf{61.5} & \textbf{31.1} & \textbf{77.4} & \textbf{67.6} & \textbf{73.5} & \textbf{79.5} \\
% \rowcolor{ColorMedical!50}
% \textbf{Hulu-Med-8B}  & \textbf{62.2} & \textbf{20.0} & \textbf{65.9} & \textbf{33.7} & 77.6 & \textbf{67.5} & \textbf{73.8} & \textbf{81.6} \\
\addlinespace
\multicolumn{9}{l}{\textit{--- Models \textgreater\ 10B ---}} \\
\rowcolor{ColorMedical!50}
HealthGPT-14B     & 63.4 & 11.3 & 39.8 & 25.7 & 68.0 & 63.4 & 66.2 & 80.2 \\
\rowcolor{ColorMedical!50}
Lingshu-32B        & \underline{70.2}  & 22.7 & 65.4 & \underline{41.1} & 77.8 & 66.1 & 74.7 & \underline{84.7} \\
\rowcolor{ColorMedical!50}
HuatuoGPT-V-34B   & 51.8 & 11.4 & 42.7 & 26.5 & 72.2 & 54.7 & 58.8 & 74.7 \\
\rowcolor{ColorMedical!50}
MedDr-40B         & 55.6 & 12.0 & 44.3 & 24.0 & 77.4 & 38.4 & 59.2 & 65.2 \\
\rowcolor{ColorMedical!50}
\textbf{Hulu-Med-14B} & 68.0 & \underline{23.2} & \underline{68.5} & 37.7 & \underline{79.8} & \underline{70.4} & \underline{78.1} & 83.3 \\
\rowcolor{ColorMedical!50}
\textbf{Hulu-Med-32B} & \textbf{72.9} & \textbf{24.2} & \textbf{68.8} & \textbf{41.8} & \textbf{80.8} & \textbf{72.8} & \textbf{80.4} & \textbf{85.6} \\
\bottomrule
\end{tabular}}
\end{table}

% \begin{figure}[!ht]
% \centering
% \includegraphics[width=1\linewidth]{Multimodal_Bench.pdf}
% \caption{Performance comparison among three categories of VLMs (Proprietary, General-purpose, and Medical) on medical multimodal benchmarks. For the `Medical VLM < 10B` subgroup, \textbf{bold} and \underline{underline} scores indicate the best and second-best methods, respectively. $^\diamondsuit$Med-R1 is trained on part of the OmniMedVQA test set. $^\heartsuit$These models do not support multi-image inputs required by MedXpertQA. $^\dagger$Lingshu models use the MMMU-Med validation set for training, thus their results on its test set are excluded.}
% \label{fig:mm_bench}
% \end{figure}

% \begin{figure*}[!t]
% \centering
% \includegraphics[width=1\linewidth]{text_bench_plot.pdf}
% \caption{Performance comparison among three categories of VLMs (Proprietary, General-purpose, and Medical) on medical text benchmarks. Within each open-source subgroup, \textbf{bold} and \underline{underline} scores indicate the best and second-best methods, respectively. Note that MedQA, MedXQA, and SGPQA denote MedQA-USMLE, MedXpertQA-Text, and SuperGPQA-Medical benchmarks, respectively.}
% \label{tab}
% \end{figure*}

\begin{figure*}[!p]
\centering
\includegraphics[width=1\linewidth]{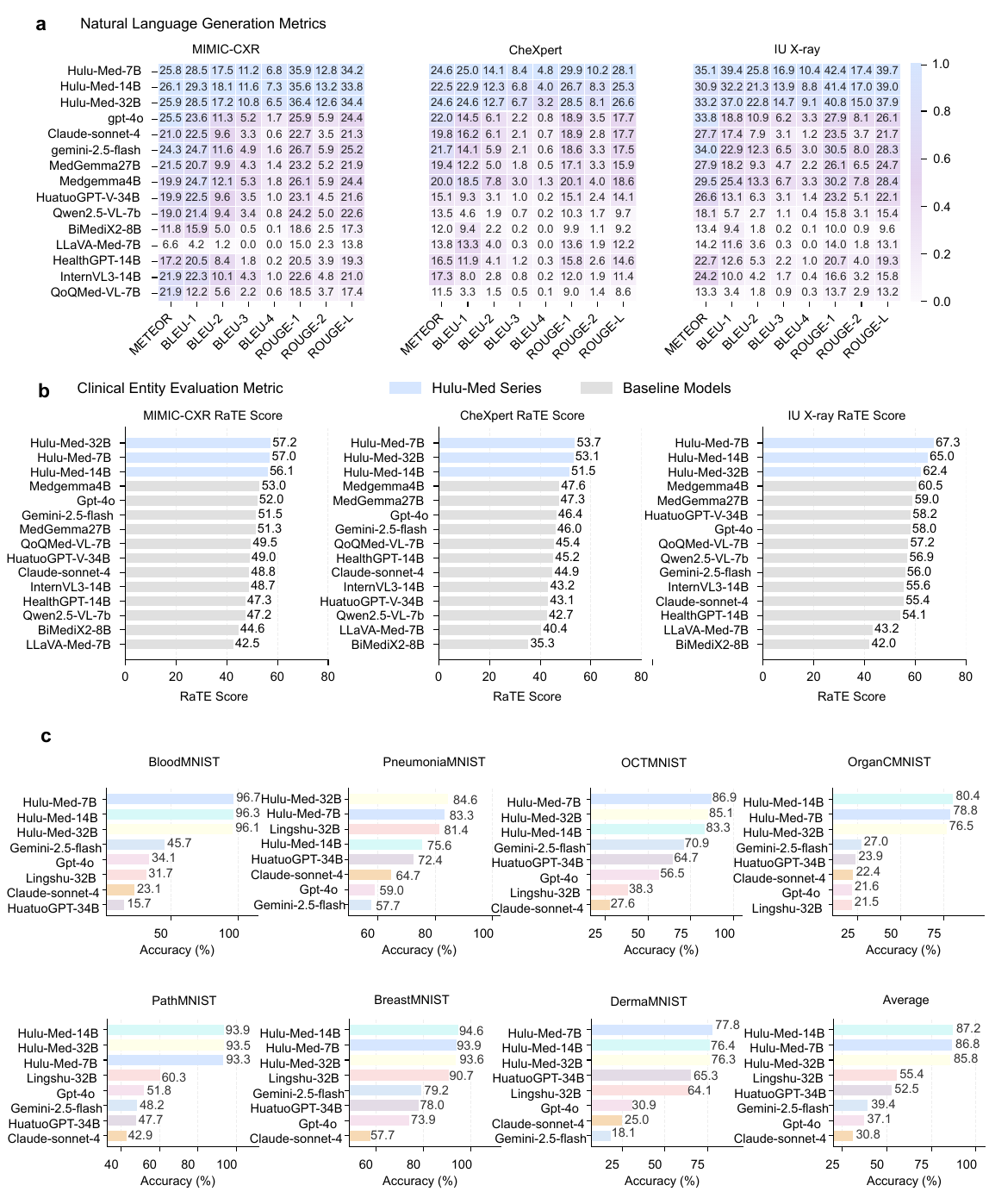}
\end{figure*}
\clearpage
\begin{center}
    \parbox{1\textwidth}{
        \captionof{figure}{\textbf{Performance evaluation of Hulu-Med on 2D medical image understanding tasks.} 
        \textbf{a}, Quantitative results for MRG on the MIMIC-CXR, CheXpert, and IU X-ray datasets are presented using standard NLG metrics.
        \textbf{b}, A head-to-head comparison of clinical fidelity in generated reports is shown using RaTEScore, a metric that more accurately reflects the semantic correctness of clinical entities than traditional language metrics.
        \textbf{c}, Comparative analysis of classification accuracy on seven sub-tasks of the MedMNIST benchmark demonstrates Hulu-Med's proficiency across a diverse range of 2D medical images. 
        }
        \label{fig:2d_results}
    }
\end{center}
\clearpage

% \begin{figure*}[!ht]
% \centering
% \includegraphics[width=1\linewidth]{med_omni_mrg.pdf}
% \caption{Results on medical report generation (MRG).}
% \label{tab}
% \end{figure*}

\begin{figure*}[!p]
\centering
\includegraphics[width=0.99\linewidth]{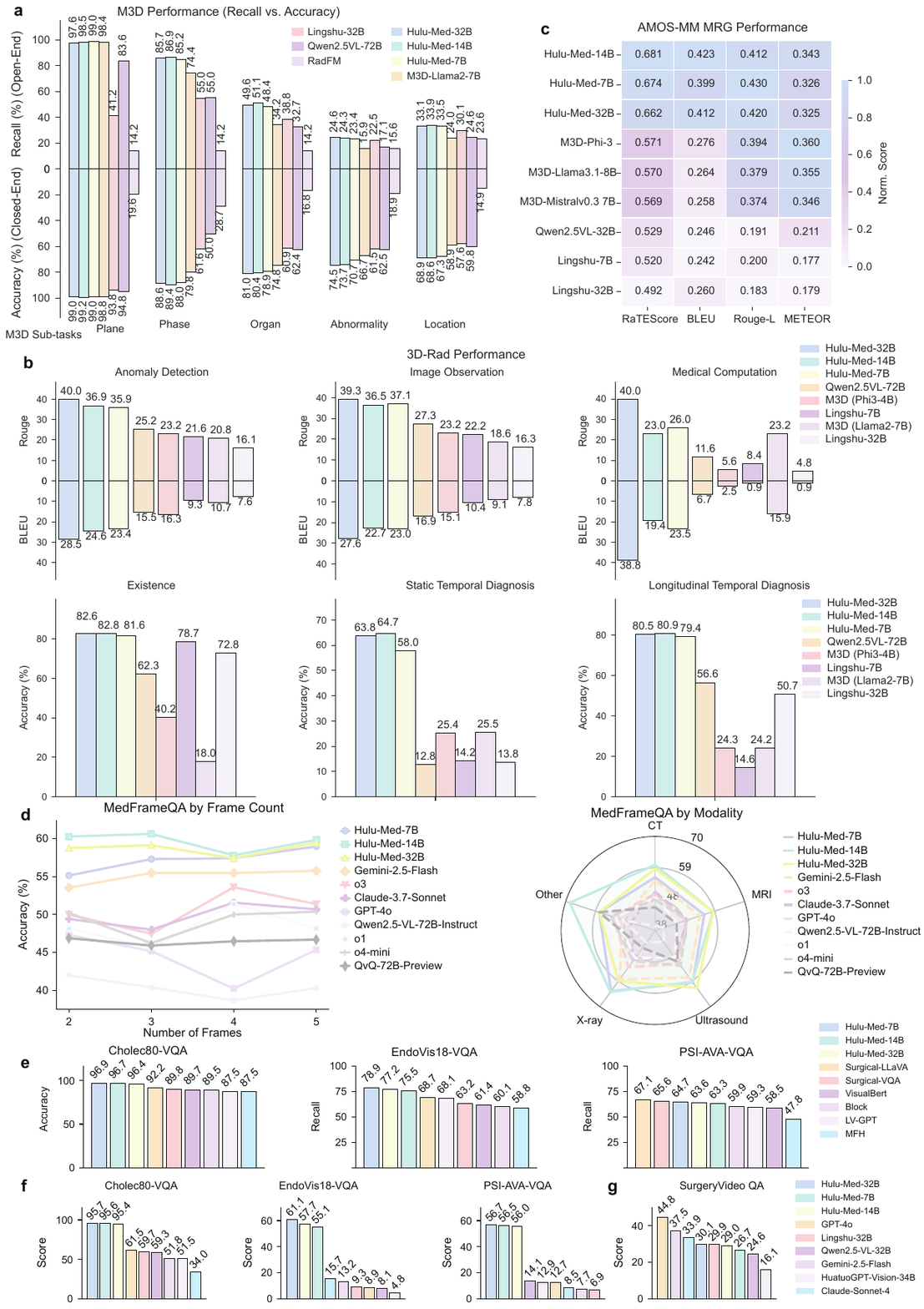}
\end{figure*}
\clearpage
\begin{center}
    \parbox{1\textwidth}{
        \captionof{figure}{\textbf{Performance evaluation of Hulu-Med on 3D and video understanding tasks.}
        \textbf{a}, Performance on M3D benchmark demonstrates high accuracy in discriminative (primarily closed yes/no questions) tasks and strong recall for descriptive (predominantly open-ended questions) tasks across anatomical categories.
        \textbf{b}, Results on 3D-RAD benchmark show proficiency in temporal reasoning for 3D volumetric data, including static and longitudinal diagnosis tasks critical for tracking disease dynamics.
        \textbf{c}, 3D MRG quality evaluated on AMOS-MM benchmark, where Hulu-Med achieves superior RaTEScore indicating high clinical fidelity, alongside strong performance on standard NLG metrics. 
        \textbf{d}, Multi-frame temporal reasoning on MedFrameQA benchmark shows leading performance than general VLMs and proprietary models.
        \textbf{e}, Surgical video comprehension evaluation comparing against surgery-specific models trained on Cholec80, EndoVis18, and PSI-AVA. Accuracy is reported for Cholec80 (closed-ended), while recall is used for EndoVis18 and PSI-AVA (open-ended).
        \textbf{f}, Comparison with general and medical VLMs using ChatGPT-4o-latest as judge to mitigate potential misjudgments from NLG metrics when answers are semantically similar but syntactically divergent.
        \textbf{g}, Results on the newly introduced SurgeryVideoQA benchmark containing both surgery-related and other medical educational videos for OOD evaluation. ChatGPT-4o-latest judges answer correctness to ensure fair comparison across different VLM output formats.
        }
        \label{fig:3d_video_bench}
    }
\end{center}

\begin{figure*}[!p]
    \centering
    \includegraphics[width=1\linewidth]{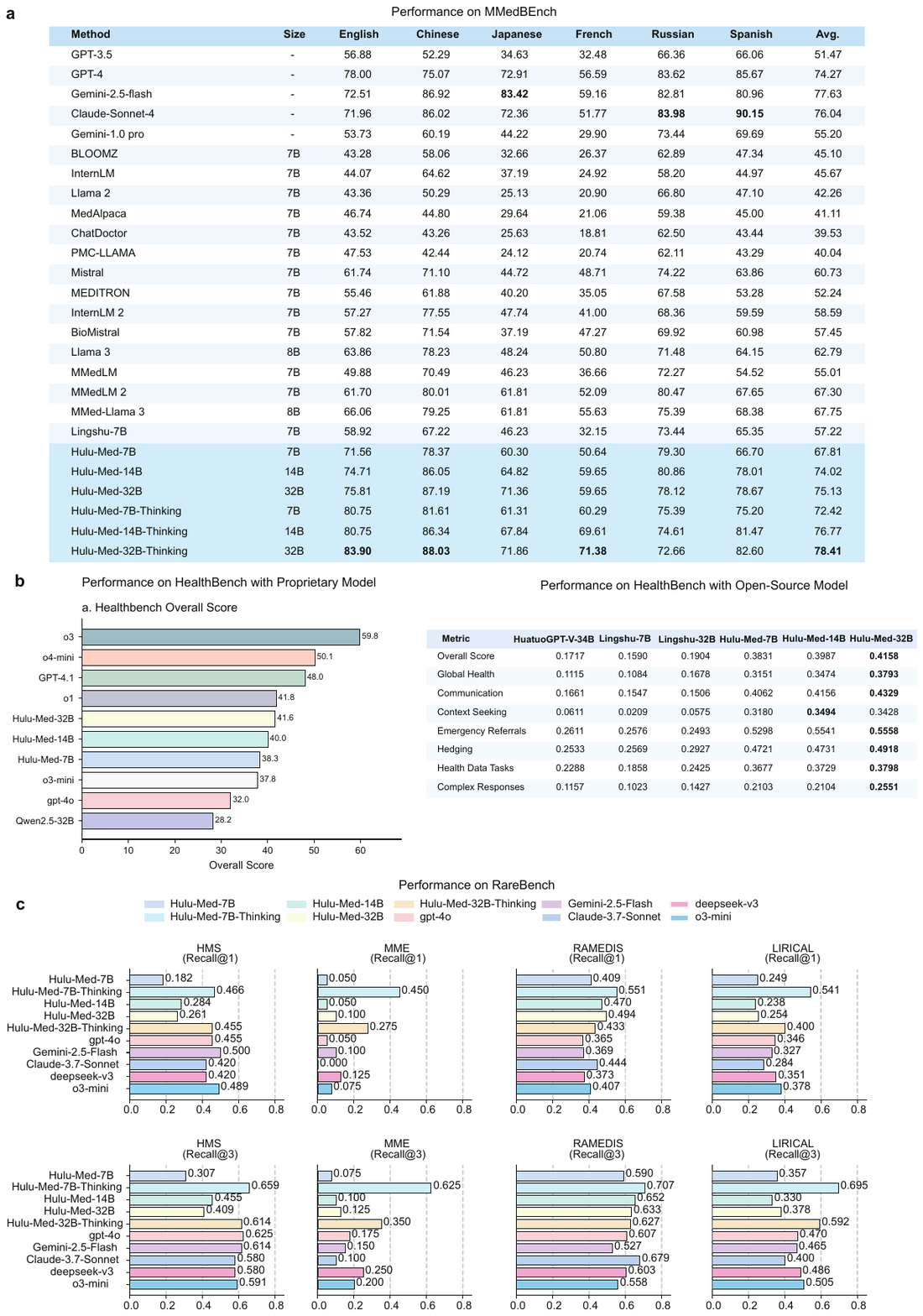}
\end{figure*}

\clearpage

\begin{center}
    \parbox{0.9\textwidth}{
         \captionof{figure}{\textbf{Evaluation of Hulu-Med's generalization capabilities in clinically critical, real-world scenarios.} 
        \textbf{a}, Multilingual medical reasoning proficiency is demonstrated on the MMedBench benchmark across six languages. \textbf{Bold text} indicates the best performance, while Hulu-Med establishes a new state-of-the-art average performance for open-source models and the proprietary GPT-4.
        \textbf{b}, Evaluation of conversational safety and clinical dialogue on HealthBench indicates that Hulu-Med outperforms general-purpose leaders such as GPT-4o and o3-mini, closes the gap with top-performing models including o3, o4-mini, and GPT-4.1, and significantly exceeds other specialized open-source models in multi-turn interactions, as assessed by physician-authored rubrics.
        \textbf{c}, Diagnostic reasoning on the long-tail rare diseases is evaluated on the RareBench benchmark, highlighting Hulu-Med's strong performance in data-constrained scenarios and its utility as a diagnostic aid.}
        \label{fig:generalization_bench} 
    }
\end{center}

\begin{figure*}[!p]
    \centering
    \includegraphics[width=1\linewidth]{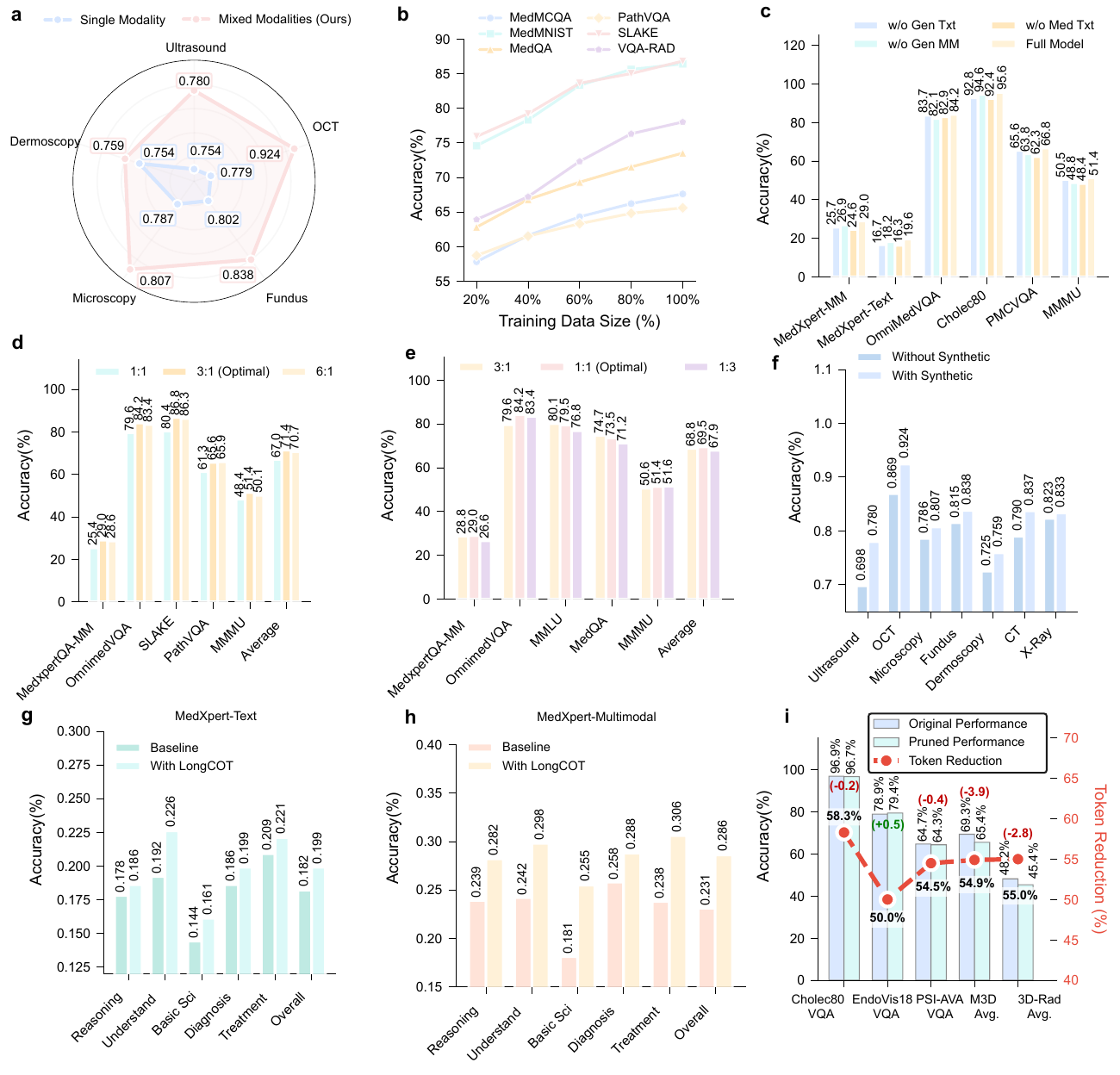}
\end{figure*}
\clearpage
\begin{center}
    \parbox{0.9\textwidth}{ 
        \captionof{figure}{\textbf{Data curation and architectural analysis of Hulu-Med.}
         \textbf{a}, The unified generalist architecture outperforms five individually trained specialist models on their respective underrepresented modalities.
        \textbf{b}, Performance exhibits a positive scaling law, monotonically increasing across text and multimodal benchmarks as training data grows from 20\% to 100\%.
         \textbf{c}, Ablation on data composition shows that removing any component—general text, general multimodal data, or medical text—degrades performance, confirming each is critical for robust reasoning.
         \textbf{d,e}, Analysis of data mixing ratios reveals optimal balances: 3:1 medical-to-general data mix and 1:1 text-to-multimodal mix yield best performance.
         \textbf{f}, Data enrichment through synthetic long captions improves accuracy across multiple imaging modalities on OmniMedVQA benchmark.
         \textbf{g,h}, Synthetically generated long CoTs provide valuable supervision, significantly improving performance on both text-only (MedXpert-Text) and multimodal (MedXpert-Multimodal) reasoning tasks.
          \textbf{i}, Medical-Aware Token Reduction achieves 55\% average token pruning during inference while maintaining performance comparable to the non-pruned model.
         }
        \label{fig:efficiency}
    }
\end{center}

% \begin{figure*}[!t]
% \centering
% \includegraphics[width=1\linewidth]{Figure5.pdf}
% \caption{Performance on Generalization Benchmarks. \textbf{a}, Multilingual reasoning across six languages on MMedBench. \textbf{b}, \textbf{d}, Diagnostic accuracy for rare diseases, showing overall scores and detailed recall metrics on RareBench. \textbf{c}, Conversational and safety performance in multi-turn clinical dialogues evaluated on HealthBench.
% }
% \label{fig:robust_bench}
% \end{figure*}

\clearpage
% GPT-4.1  (released between January and April 2025)

\section*{Methods}
\label{2d_3d_video_methods}
\subsection*{Model Architecture}
Hulu-Med is a unified VLM that enables processing a wide spectrum of medical input, including text, 2D images, 3D volumes, and videos, and generates coherent textual responses with a single, end-to-end framework. The model consists of four core components:
(1) a rotary position-adaptive visual encoder, (2) a text tokenizer, (3) a multimodal projector, and (4) an LLM decoder. 
%The processing pipeline for multimodal inputs is described sequentially through these components.

\paragraph{Rotary Position-Adaptive Visual Encoder}
The visual processing pipeline begins with the rotary position-adaptive visual encoder, designed to handle heterogeneous medical data by treating all visual inputs as a unified sequence of 2D image planes and adopting image patches as a universal processing unit. The encoder is a 27-layer Vision Transformer (ViT) with a hidden size of 1152, an intermediate MLP size of 4304, and 16 attention heads. Specifically, 3D medical volumes ({\em e.g.}, CT, MRI) are decomposed into their constituent slices, and videos are sampled into frames. Each plane is then partitioned into a grid of non-overlapping $16 \times 16$ pixel patches, which are linearly embedded. 

A key innovation is the replacement of standard, fixed-size absolute positional embeddings with 2D RoPE. To encode the relative position of a patch at grid coordinates $(m, n)$, we apply 1D RoPE  along the height and width dimensions independently. Let $\mathbf{q} \in \mathbb{R}^d$ denote a query (or key) vector ($d$ is even). We partition these $d$ features into $d/2$ pairs, where dimensions $2i-1$ and $2i$ form the $i$-th pair for $i \in \{1, \dots, d/2\}$. For each spatial dimension $p \in \{m, n\}$, we apply a rotation transformation to each pair:
\begin{equation}
\begin{pmatrix} q'_{2i-1} \\ q'_{2i} \end{pmatrix} = 
\begin{pmatrix} \cos(p\theta_i) & -\sin(p\theta_i) \\ \sin(p\theta_i) & \cos(p\theta_i) \end{pmatrix}
\begin{pmatrix} q_{2i-1} \\ q_{2i} \end{pmatrix}
\end{equation}
where the frequencies are $\theta_i = 10000^{-2i/d}$. The first $d/2$ dimensions encode the height position $m$, while the remaining $d/2$ dimensions encode the width position $n$. This design embeds relative spatial information directly into the self-attention mechanism without requiring learned positional embeddings, enabling the encoder to process images of arbitrary resolutions and aspect ratios within a single unified framework.
%\weidi{I don't understand what is `consecutive pairs'.(done)}

To manage the heavy computational load from 3D and video modalities, we employ a two-stage token reduction strategy. 
First, at the \textbf{intra-plane level}, we apply local spatial pooling to 3D and video inputs by setting a merge factor of 2. This step combines each $2 \times 2$ block of adjacent patch tokens into a single token via bilinear interpolation, reducing the number of visual tokens for each plane by a factor of 4. This pooling is omitted for single 2D images, which pose less computational burden. 
Second, at the \textbf{inter-plane level}, we implement a medical-aware token reduction strategy. This mechanism prunes redundant visual tokens by comparing corresponding patches across adjacent slices or frames. Specifically, for video inputs, we compute the $L_1$ distance between the embeddings of spatially corresponding patches from consecutive frames: $\text{diff}_t^{(i)} = \|f_v(\mathbf{v}_t)^{(i)} - f_v(\mathbf{v}_{t-1})^{(i)}\|_1$
where $f_v(\mathbf{v}_t) \in \mathbb{R}^{N_v \times d_v}$ represents the visual token embeddings at frame $t$ output by the vision encoder, $f_v(\mathbf{v}_t)^{(i)} \in \mathbb{R}^{d_v}$ denotes the embedding vector of the $i$-th patch at frame $t$, and $f_v(\mathbf{v}_{t-1})^{(i)} \in \mathbb{R}^{d_v}$ denotes the embedding of the spatially corresponding patch at frame $t-1$. Patches with $\text{diff}_t^{(i)} < \tau$ (where $\tau=0.1$) are considered temporally redundant and pruned from frame $t$. This patch-level pruning strategy is applied during the forward pass after visual encoding and is performed dynamically based on the current visual encoder parameters. The strategy reduces the final visual token count by up to 60\% for 3D and video inputs while maintaining comparable performance.

\paragraph{Text Tokenizer}
For textual input, we employ the tokenizer native to the LLM backbone, which is a Byte-Pair Encoding (BPE) tokenizer with a vocabulary size of 152,064 tokens~\cite{sennrich2016neural}. 

%This tokenizer is optimized for processing multilingual text and special characters common in medical literature.

\paragraph{Multimodal Projector}
To bridge the visual and text domains, a multimodal projector~($g(\cdot)$) aligns the output of the vision encoder with the LLM's embedding space with a two-layer Multilayer Perceptrons (MLPs). 
It takes the final sequence of visual patch embeddings from the vision encoder, $f_v(\mathbf{v}) \in \mathbb{R}^{N_v \times d_v}$ ($d_v = 1152$), and transforms it into a sequence of language-compatible embeddings, $g(f_v(\mathbf{v})) \in \mathbb{R}^{N_v \times d}$:
\begin{equation}
    g(f_v(\mathbf{v})) = W_2 \cdot \text{GELU}(W_1 \cdot f_v(\mathbf{v}) + b_1) + b_2,
\end{equation}
where $W_1, W_2, b_1, b_2$ refer to the learnable parameters in projector, which is crucial for enabling the LLM $\Phi(\cdot)$ to interpret the visual information as if it were part of its native language space.
%\weidi{you used $x$ as notation in previous section for medical-aware token reduction.(DONE\st{I changed medical aware token sec for f(v) same as problem formulation})}

\paragraph{LLM Decoder}
For our primary configuration, \texttt{Hulu-Med-7B} adopts Qwen2.5-7B-Instruct as language model backbone. The processed text embeddings and the projected visual embeddings are concatenated into a single, unified input sequence. The model then processes this sequence auto-regressively, predicting the next token based on all preceding visual and textual tokens. This  architecture allows Hulu-Med to perform a diverse array of generative tasks without requiring any task-specific modifications. To demonstrate framework scalability, we also developed \texttt{Hulu-Med-14B} and \texttt{Hulu-Med-32B}, which are built upon Qwen3-14B and Qwen2.5-32B backbone respectively, providing a range of model sizes to balance performance and computational efficiency.

\subsection*{Training Strategy}
\label{sec:train_strategy}

Hulu-Med is trained with a progressive three-stage curriculum: 
(1) vision–language alignment, (2) continual medical multimodal pretraining, and (3) mixed-modality instruction tuning. 
Specifically, stages 1 and 2 consolidate 2D single-image competence, stage 3 introduces interleaved multi-image contexts and spatiotemporal reasoning over 3D volumes and videos. 
Each stage uses a distinct large-scale data mixture that combines public datasets with targeted synthetic pipelines, addressing two common bottlenecks in medical VLMs: the limited volume and diversity of visual instruction data, and the insufficient integration of general and medical knowledge.

\paragraph{Stage 1: Vision-Language Alignment}
The initial stage focuses on establishing a foundational alignment between the vision encoder and the LLM backbone. The primary task is short caption generation, where the model learns to produce text for a given image, and the training loss is calculated against the ground-truth caption. To this end, we utilized a corpus of 1.4 million image-text pairs sourced entirely from a collection of public medical datasets (\textcolor{blue}{Extended Tab.~\ref{tab:stage0_data}}) including Quilt, MedICaT, and ROCO. This data spans a wide range of modalities and resolutions, enabling the rotary position-adaptive visual encoder to learn to handle diverse visual inputs. During this stage, the LLM backbone remains frozen; we only fine-tune the multimodal projector and the vision encoder with learning rates of $1 \times 10^{-3}$ and $1 \times 10^{-5}$, respectively.

\paragraph{Stage 2: Medical Multimodal Pre-training}
The second stage aims to inject extensive medical knowledge while enhancing the model's visual understanding capability, training on a corpus of 4.9 million samples. The training objective is elevated to more complex generative tasks, primarily long-form caption generation and open-ended question answering. For this, we first compiled a 2.6 million sample corpus from public datasets~(\textcolor{blue}{Extended Tab.~\ref{tab:stage2_data_corrected}}). This included long-form medical captions ({\em e.g.}, PubMedVision) and a variety of general-domain data such as documents and charts, along with approximately 10\% general-domain text to preserve core language capabilities.

However, public datasets exhibit a significant long-tail distribution, where modalities like ultrasound and dermatology with detailed text annotations are underrepresented. To mitigate this, we synthesized an additional 2.3 million high-quality long captions. For images with only short captions, a multi-agent pipeline employed a large VLM (Gemini-2.5-Pro) to rewrite them into rich, detailed descriptions, yielding 1.4 million enhanced captions. For images that lacked any text annotations, we implemented a distinct multi-agent generation process where a core VLM generated candidate captions that were then evaluated and ranked by specialized `judge' models. 
At this pre-training stage, all model components are trainable, with learning rates of $2 \times 10^{-6}$ (vision encoder), $1 \times 10^{-5}$ (projector), and $2.5 \times 10^{-5}$ (LLM), managed by a cosine scheduler.

\paragraph{Stage 3: Mix-Modality Instruction Tuning}
The final stage trains on a wide spectrum of downstream tasks to encourage the instruction-following ability. 
We train on a dataset with 10.5 million instances, including discriminative tasks like, VQA and classification, as well as complex generative tasks such as MRG and CoT reasoning. 
The dataset was gathered from public instruction-tuning data (\textcolor{blue}{Extended Tab.~\ref{tab:final_definitive_verified}}), including 5.9 million text-based and 4.5 million multimodal instructions, which include diverse formats such as multi-image, interleaved, 3D, and video data.

To address critical data limitation in public resources, we developed three novel synthesis pipelines. First, to enhance multilingual reasoning, we synthesized a 45K sample CoT dataset. Our methodology employed a role-play prompting strategy combined with rejection sampling, where we retained only the reasoning paths that ends up with correct final answers. Second, we generated 600K high-quality VQA pairs by prompting Gemini-2.5-Pro to create questions directly answerable from our generated long captions. Finally, to overcome the scarcity of annotated medical videos, we developed a ``divide-and-conquer'' captioning method, yielding ~20K video captions. During this stage, all model parameters remained trainable, with the LLM learning rate increased to $5 \times 10^{-5}$.

\subsection*{Evaluation Framework and Metrics}
\label{sec:eval_metrics}
To comprehensively assess the capabilities of \textbf{Hulu-Med}, we established a comprehensive and rigorous evaluation framework, examining the model's performance on various data modalities and clinical tasks, ensuring a holistic understanding of its strengths and limitations. The benchmarks are organized by modality—text, 2D images, 3D volumes, and video—with appropriate metrics tailored to each task type.

\paragraph{Text-Based Medical Reasoning and Generalization} 
To ensure that multimodal training maintains core textual knowledge and reasoning ability, we evaluated the model on both standard medical benchmarks and specialized generalization tasks. 
For text-only question-answering, we assess medical knowledge without visual input across eight challenging benchmarks simulating professional medical board examinations (MedQA-USMLE, MedMCQA, MMLU-Med, MedBullets), evaluating factual recall from biomedical literature (PubMedQA), and probing advanced expert-level reasoning skills (MMLU-Pro-Med, MedXpertQA-Text, SuperGPQA-Medical). For these multiple-choice benchmarks, we report accuracy as the primary performance measure, quantifying the percentage of correct predictions against ground-truth labels.

In addition, we also assess the model for real-world deployment on three benchmarks. MMedBench evaluates multilingual medical understanding across six languages (English, Chinese, Spanish, French, Russian, and Japanese), with performance measured using accuracy on multiple-choice questions. HealthBench assesses conversational safety and clinical performance in realistic multi-turn dialogues against fine-grained physician-authored rubrics, covering seven core clinical competency themes: Global Health, Communication, Context Seeking, Emergency Referrals, Hedging, Health Data Tasks, and Complex Responses. RareBench measures diagnostic reasoning on rare diseases, testing performance in data-scarce scenarios where the model must handle uncommon clinical presentations with limited training examples.

For complex open-ended response tasks requiring nuanced evaluation, we employ advanced LLMs as judges. In HealthBench, given the complexity of physician-designed rubrics and the need to assess long-form conversational responses, we use Gemini-2.5-Pro as judge. For diagnostic reasoning tasks in RareBench, evaluation is conducted using ChatGPT-4o-latest, which assesses the correctness and clinical appropriateness of differential diagnoses.

\paragraph{2D Medical Image Understanding}
We assess performance on two primary tasks: VQA and MRG. For VQA, we adopt seven benchmarks to test visual-language alignment across multiple dimensions: broad multi-modal understanding across various imaging types (OmniMedVQA, PMC-VQA), domain-specific knowledge in radiology (VQA-RAD, SLAKE) and pathology (PathVQA), and higher-order cognitive skills integrating external knowledge with visual reasoning (MedXQA, MMMU-Med). For classification tasks on MedMNIST and the majority of closed-ended VQA benchmarks, we report accuracy as the primary metric.

For MRG, we evaluate the model's ability to produce clinically accurate reports from chest radiographs on the MIMIC-CXR, CheXpert, and IU X-ray datasets. We employ a multi-faceted approach for these generative tasks: linguistic fluency is assessed using standard NLG metrics, including BLEU (1-4), ROUGE-L, and METEOR; to measure the inclusion of key clinical concepts, we compute recall; and to assess clinical utility beyond lexical similarity, we incorporate RaTEScore, a domain-specific metric that evaluates the semantic correctness of medical entities, their attributes, and negations.

\paragraph{3D Volumetric and Spatiotemporal Analysis} 
\label{3D Volumetric and Spatiotemporal Analysis}
To evaluate Hulu-Med's ability to process 3D volumetric data, we select benchmarks that test both anatomical understanding and temporal reasoning within image series. 
Specifically, for 3D spatial reasoning, including tasks like plane detection and organ identification, we primarily evaluate on the M3D benchmark. To assess a broader spectrum of clinical reasoning skills, we employ the comprehensive 3D-RAD benchmark, which is composed of multiple distinct sub-tasks, from descriptive generation ({\em e.g.}, anomaly detection and image observation) and closed-ended classification ({\em e.g.}, existence) to both static and longitudinal temporal diagnosis.

Our evaluation strategy for 3D tasks mirrors that of the 2D domain. For reasoning tasks within the M3D and 3D-RAD benchmarks, performance on closed-ended questions is measured by accuracy, while descriptive, open-ended sub-tasks are evaluated using recall to assess the coverage of key clinical information. To test its generative capabilities, we use the AMOS-MM benchmark to assess the quality and clinical fidelity of 3D medical report generation, employing the same combination of NLG metrics (BLEU, ROUGE-L, METEOR) and the clinically-aligned RaTEScore.

\paragraph{Surgical and Medical Video Comprehension} 
The model's ability to interpret dynamic visual data is tested on a set of video-based benchmarks. Surgical video datasets, including Cholec80-VQA, EndoVis18-VQA, PSI-AVA-VQA, and the general SurgeryVideoQA, are used to evaluate the understanding of surgical phases, instruments, and actions. Additionally, the MedFrameQA benchmark is used to specifically assess multi-frame temporal reasoning across various medical imaging sequences, testing the model's ability to comprehend dynamic processes.

Our evaluation strategy is tailored to the specific characteristics of each dataset. For Cholec80-VQA, where most questions are closed-ended, we primarily use accuracy as the evaluation metric. For EndoVis18-VQA and PSI-AVA-VQA, where answers consist of short descriptive phrases, we employ recall to evaluate whether the model captures the essential clinical concepts. For MedFrameQA, we similarly adopt accuracy as the metric since it comprises multiple-choice questions. For SurgeryVideoQA, which features open-ended questions requiring free-form responses, traditional metrics are insufficient; therefore, we utilize ChatGPT-4o-latest as an automated judge to assess answer quality through semantic evaluation.
Furthermore, to ensure a more comprehensive evaluation, we extend the use of ChatGPT-4o-latest beyond SurgeryVideoQA to Cholec80-VQA, EndoVis18-VQA, and PSI-AVA-VQA, where answers range from single words to short phrases, as a supplementary metric. This provides a semantic assessment that captures clinical correctness beyond simple lexical matching.

\section*{Code and Data Availability}
The detailed implementation, including fine-tuned models and code, as well as all datasets used in this work, are publicly available at \textcolor{blue}{{https://github.com/ZJUI-AI4H/Hulu-Med}}. Detailed licensing information and data download links can be found in \textcolor{blue}{Extended Table~\ref{tab:dataset_license_full}}.

%\section*{Acknowledgements}
%This work is supported by the National Key R\&D Program of China (Grant No. 2024YFC3308304), the "Pioneer" and "Leading Goose" R\&D Program of Zhejiang (Grant no. 2025C01128), the National Natural Science Foundation of China (Grant No. 62476241), the Natural Science Foundation of Zhejiang Province, China (Grant No. LZ23F020008).

\section*{Author Contributions Statement}
W.X., J.S., J.W., and Z.L. conceived the project. S.J. designed the algorithm and performed model training. Y.W., C.Z., Y.Z., B.P., and S.S. carried out data collection. S.J., Y.W., and T.H. designed the experiments. Data analysis was conducted by S.J., J.T.Z., J.H., Z.C., R.W., J.L., H.X., T.T., K.L., J.X., B.F., and F.Z. The figures were generated and revised by C.Z., S.J., T.H., Z.Y., and Y.F. The results were interpreted by S.J., W.X., J.S., J.W., and Z.L. The manuscript was written by S.J., T.H., Z.L., J.S., and W.X. All authors contributed to the final revision of the manuscript.

\section*{Competing Interests Statement}
S.S. and Z.Y. are employees of Alibaba Inc. Y.F. is an employee of Angelalign Technology Inc. T.T. is an employee of China Mobile Group Zhejiang Company Limited. The remaining authors declare no competing interests.

\end{bibunit}

% %Supplementary
% \clearpage
\renewcommand{\tablename}{Extended Table}
\renewcommand{\thetable}{\arabic{table}}
\renewcommand{\figurename}{Extended Figure}
\renewcommand{\thefigure}{\arabic{figure}}
\setcounter{table}{0}
\setcounter{figure}{0}

\clearpage%Supp Figure

\begin{table*}[t]
\centering
\caption{Comparison of medical vision-language models}
\label{tab:data_composition}
\resizebox{\textwidth}{!}{
\begin{tabular}{l|c|c|c|c|ccc|c|cccc}
\toprule
\textbf{Model} & \textbf{Model Sizes} & \textbf{Open Model} & \textbf{Open Data} & \textbf{Data Size} & \multicolumn{3}{c|}{\textbf{Data Source}} & \textbf{Modalities} & \multicolumn{4}{c}{\textbf{Downstream Tasks}} \\
& & & & & \textbf{General} & \multicolumn{2}{c|}{\textbf{Medical}} & & \textbf{Text} & \textbf{2D} & \textbf{3D} & \textbf{Video} \\
& & & & & & \textbf{From Papers} & \textbf{Real-world} & & & & & \\
\midrule
Lingshu & 7B, 32B & \checkmark & \texttimes & 12.2M & 7.15M & 2.6M & 2.45M & 12 & \checkmark & \checkmark & \texttimes & \texttimes \\
HuatuoGPT-Vision & 7B, 34B & \checkmark & \checkmark & 1.3M & - & 1.3M & - & 9 & \texttimes & \checkmark & \texttimes & \texttimes \\
LLaVA-Med & 7B & \checkmark & \checkmark & 560K & - & 560K & - & 4 & \texttimes & \checkmark & \texttimes & \texttimes \\
RadFM & 16M & \checkmark & \checkmark\textsuperscript{*} & - & - & 14.16M & 1.84M & 6 & \texttimes & \checkmark & \checkmark & \texttimes \\
HealthGPT & 4B, 14B & \checkmark & \checkmark & 1.82M & 558K & 1.21M & 56K & 7 & \texttimes & \checkmark & \texttimes & \texttimes \\
\midrule
\textbf{Hulu-Med (Ours)} & \textbf{4B, 7B, 14B, 32B} & \checkmark & \checkmark & \textbf{16.6M} & \textbf{4.5M} & \textbf{1.8M} & \textbf{10.3M} & \textbf{14} & \checkmark & \checkmark & \checkmark & \checkmark \\
\bottomrule
\end{tabular}
}
\vspace{-0.2cm}
\footnotesize{\textsuperscript{*}Partially open-sourced, requires application for some datasets. "From Papers" refers to data from PubMed/PMC.}
\end{table*}

\setcounter{table}{1}
\begin{longtable}{p{2.4cm} p{1cm} p{1cm} p{1.5cm} p{1cm} p{1.7cm} p{6cm}}
\caption{Overview of medical benchmarks} 
\label{tab:benchmark}\\
\toprule
\textbf{Benchmark} & \textbf{Type} & \textbf{Mod.} & \textbf{w/ Clin.} & \textbf{Num.} & \textbf{Dist.} & \textbf{Data Source Description} \\
\midrule
\endfirsthead

\toprule
\textbf{Benchmark} & \textbf{Type} & \textbf{Mod.} & \textbf{w/ Clin.} & \textbf{Num.} & \textbf{Dist.} & \textbf{Data Source Description} \\
\midrule
\endhead

\midrule
\multicolumn{7}{r}{\textit{Continued on next page}} \\
\midrule
\endfoot

\bottomrule
\endlastfoot

MMLU-Med       & QA             & text   & No  & 633   & in-domain   & USMLE practice exams, textbooks, prep materials \\
PubMedQA       & QA             & text   & Yes & 1000  & in-domain & PubMed biomedical abstracts and conclusions \\
MedMCQA        & QA             & text   & No  & 6150  & in-domain      & AIIMS PG \& NEET-PG official exam banks (1991--present) \\
MedQA          & QA             & text   & No  & 1273  & in-domain     & USMLE, Chinese \& Taiwanese medical license exam questions \\
MedBullets     & QA             & text   & No  & 124   & ood       & USMLE Step 2 \& 3 style questions from MedBullets platform \\
SGPQA          & QA             & text   & No  & 2755  & ood       & Graduate-level multiple-choice expert-authored questions \\
MMLU-Pro-Med   & QA             & text   & No  & 818   & ood       & Academic exams \& textbooks (medical portion) \\
MedXpertQA-Text     & QA             & text   & Yes & 2000  & ood       & Expert-level exam questions + clinical images \& patient records \\
MedXpertQA-MM     & QA             & 2D   & Yes & 2000  & ood       & Expert-level exam questions + clinical images \& patient records \\
OmniMedVQA     & Mixed          & Mixed  & Yes & 87944 & ood       & Images and QAs from 73 medical datasets (12 modalities) \\
PMC-VQA        & VQA            & 2D     & Yes & 33430 & in-domain & Figures and captions from PubMed Central OA articles \\
MMMU-Med       & VQA            & 2D     & No  & 1751  & ood       & College-level exams, quizzes, and textbooks (Health \& Medicine) \\
VQA-RAD        & VQA            & 2D     & Yes & 451   & in-domain & Radiology images with clinician-authored QAs \\
SLAKE          & VQA            & 2D     & Yes & 1061  & in-domain & Radiology images + knowledge graph generated QAs \\
PathVQA        & VQA            & 2D     & Yes & 6761  & in-domain & Pathology images from textbooks \& digital libraries \\
MedMNIST       & Class. & 2D     & No  & 22602 & in-domain & Biomedical images (public datasets, downsampled, CC licensed) \\
MIMIC-CXR      & MRG            & 2D     & Yes & 2343  & in-domain & 377,110 chest X-rays + reports from BIDMC hospital (2011--2016) \\
CheXpert       & MRG            & 2D     & Yes & 234   & in-domain & 224,316 chest radiographs with uncertainty labels \\
IU-Xray        & MRG            & 2D     & Yes & 590   & in-domain & 3,996 reports, 8,121 X-rays from Indiana Network for Patient Care \\
M3D            & Mixed          & 3D     & Yes & 27582 & in-domain & 120K 3D CT image-report pairs, plus 25 public segmentation datasets \\
3D-Rad        & Mixed          & 3D     & Yes & 33910 & in-domain & 25,692 chest CT scans + reports, 21,304 patients \\
AMOS           & MRG            & 3D     & Yes & 400   & ood       & 500 abdominal CT + 100 MRI with 15 organ annotations \\
MedFrameQA     & VQA            & Video     & Yes & 2850  & ood       & Multi-image QA from clinical/educational surgical videos (YouTube etc.) \\
Cholec80-VQA   & VQA            & Video  & Yes & 6606  & in-domain & QA based on Cholec80 dataset (80 laparoscopic cholecystectomy videos) \\
EndoVis18-VQA  & VQA            & Video  & Yes & 643   & in-domain & QA derived from EndoVis 2018 surgical scene segmentation dataset \\
PSI-AVA-VQA    & VQA            & Video  & Yes & 4402  & in-domain & Holistic surgical scene dataset with $\sim$4402 QA pairs \\
SurgeryVideoQA & VQA            & Video  & Yes & 2690  & in-domain & QA derived from Cholec80 surgical workflow dataset \\
HealthBench    & Case           & text   & No  & 5000  & ood       & 5,000+ simulated medical conversations with evaluation rubrics designed by 262 physicians \\
RareBench      & Case           & text   & Yes & 1122    & ood       & 1,197 rare disease patient cases (Electronic Health Records) \\
MMedBench      & VQA            & text   & Yes & 8518  & in-domain      & 21 medical fields, including Internal Medicine, Biochemistry, Pharmacology, and Psychiatry \\
\label{tab:bench_all}
\end{longtable}

%%%%Benchmark on table format
\FloatBarrier

\begin{table*}[t]
\centering
\caption{Comprehensive modality coverage in the Hulu-Med dataset, detailing its 14 main modalities and 65 listed sub-modality examples.}
\label{tab:modality_coverage}
\resizebox{\textwidth}{!}{
\begin{tabular}{l|p{11.5cm}}
\toprule
\textbf{Main Modality} & \textbf{Sub-modalities and Examples} \\
\midrule
CT & CTA, CECT, DECT, HRCT, CBCT, Cardiac CT, etc. \\
\midrule
MRI & fMRI, DTI, DWI, SWI, MRA, MRCP, MRV, Cardiac MRI/CMR, etc. \\
\midrule
Radiography (X-ray) & Chest X-ray (CXR), Mammography/DBT, DXA/DEXA, etc. \\
\midrule
Ultrasound & Echocardiography, Doppler, CEUS, IVUS, etc. \\
\midrule
Nuclear Medicine & PET, FDG-PET, PET/CT, PET/MRI, SPECT, Scintigraphy, Gamma Camera, etc. \\
\midrule
Fluoroscopy & C-arm Fluoroscopy, Cinefluoroscopy, Voiding Cystourethrography (VCUG), etc. \\
\midrule
Angiography & Catheter Angiography, Coronary Angiography, Venography, DSA, etc. \\
\midrule
Endoscopy & Gastroscopy, Colonoscopy, Bronchoscopy, Arthroscopy, Laparoscopy, etc. \\
\midrule
OCT & SD-OCT, SS-OCT, OCTA, OFDI, LC-OCT, HF-OCT, etc. \\
\midrule
Ophthalmic Imaging & Fundus Photography, Fluorescein Angiography (FA), ICG Angiography (ICGA), SLO/SLO-AF, RetCam, Ophthalmoscopy, etc. \\
\midrule
Dermatology Imaging & Dermoscopy, Trichoscopy, Reflectance Confocal Microscopy (RCM), etc. \\
\midrule
Pathology/Microscopy & Histopathology, Cytology/Cytopathology, Immunohistochemistry (IHC), Electron Microscopy (SEM/TEM), Gross Pathology, etc. \\
\midrule
Clinical Photography & Digital Photography, Clinical Photograph/Image/View, etc. \\
\midrule
Physiological Signals & Medical Graph/Chart/Diagram, ECG/EKG/EEG, etc. \\
\bottomrule
\end{tabular}
}
\end{table*}

\FloatBarrier
\begin{table*}[!htbp]
\centering
\caption{Stage 1 training data composition (1.42M entries)}
\label{tab:stage0_data}
\sisetup{group-separator={,}}
\setlength{\tabcolsep}{8pt} % Adjust column spacing for this table
\renewcommand{\arraystretch}{1.2} % Adjust row spacing for this table

\begin{tabular}{l l l S[table-format=7.0]}
\toprule
% --- TABLE HEADER ---
\rowcolor{HeaderColor}
\multicolumn{1}{c}{\textcolor{white}{\textbf{Category}}} & 
\multicolumn{1}{c}{\textcolor{white}{\textbf{Modality}}} & 
\multicolumn{1}{c}{\textcolor{white}{\textbf{Dataset Name}}} & 
\multicolumn{1}{c}{\textcolor{white}{\textbf{Entry Count}}} \\
\midrule

%===========================================================%
%                  MEDICAL SHORT CAPTION DATA               %
%===========================================================%
% Using multirow for the main category to maintain visual consistency with other tables.
\multirow{5}{*}{\rotatebox{90}{\bfseries\textcolor{Stage0Color}{Short Caption}}}
 & \textit{Histopathology} & Quilt-LLaVA-Pretrain & 723328 \\
 \cmidrule(l){2-4}
 & \textit{Clinical} & biomedica-clinical & 395616 \\
 \cmidrule(l){2-4}
 & \textit{Multimodal} & Medicat & 217060 \\
 \cmidrule(l){2-4}
 & \textit{Radiology} & ROCOv2-radiology & 79793 \\
 & & Medpix2.0 & 2050 \\
\midrule
\rowcolor{SubtotalColor}
\multicolumn{3}{r}{\textbf{GRAND TOTAL}} & \textbf{1,417,847} \\
\bottomrule
\end{tabular}
\end{table*}

\begin{table*}[!htbp]
\centering
\caption{Stage 2 training data composition (4.85M entries)}
\label{tab:stage2_data_corrected}
\sisetup{group-separator={,}}
\setlength{\tabcolsep}{6pt}
\renewcommand{\arraystretch}{1.15}

\begin{tabular}{l l l S[table-format=7.0]}
\toprule
% --- TABLE HEADER ---
\rowcolor{HeaderColor}
\multicolumn{1}{c}{\textcolor{white}{\textbf{Source}}} & 
\multicolumn{1}{c}{\textcolor{white}{\textbf{Modality / Domain}}} & 
\multicolumn{1}{c}{\textcolor{white}{\textbf{Dataset Name}}} & 
\multicolumn{1}{c}{\textcolor{white}{\textbf{Entry Count}}} \\
\midrule

%===========================================================%
%                       SYNTHETIC DATA (Corrected)          %
%===========================================================%
\multirow{16}{*}{\rotatebox{90}{\bfseries\textcolor{SyntheticColor}{Synthetic Data}}}
 & \textit{Medical Clinical Caption} & biomedica\_clinical\_synthetic & 350768 \\
 & \textit{Medical Dermatology Caption} & biomedica\_dermatology\_synthetic & 111901 \\
 & & dermoscopy\_synthetic & 196537 \\
 \cmidrule(l){2-4}
 & \textit{Medical Histopathology} & biomedica\_histopathology\_synthetic & 194075 \\
 \cmidrule(l){2-4}
 & \textit{Medical Microscopy Caption} & biomedica\_microscopy\_synthetic & 104830 \\
 & & Microscopy\_synthetic & 22417 \\
 \cmidrule(l){2-4}
 & \textit{Medical Surgery Caption} & biomedica\_surgery\_synthetic & 99024 \\
 \cmidrule(l){2-4}
 & \textit{Medical Radiology Caption} & ROCOv2\_radiology\_synthetic & 79788 \\
 & & mimic\_synthetic & 242009 \\
 & & iu\_xray\_synthetic & 2365 \\
 \cmidrule(l){2-4}
 & \textit{Medical Multimodal Caption} & medicat\_synthetic & 217052 \\
 & & medmnist\_synthetic & 149704 \\
 & & train\_all\_reformat2\_synthetic & 3363 \\
 \cmidrule(l){2-4}
 & \textit{Medical Fundus Caption} & Fundus\_OCT\_synthetic & 86139 \\
 \cmidrule(l){2-4}
 & \textit{Medical Ultrasound Caption} & Ultrasound\_synthetic & 28559 \\
 & & Radimagenet\_synthetic & 379030 \\
\midrule
\multicolumn{3}{r}{\textbf{Synthetic Data Subtotal}} & \textbf{2,267,561} \\
\midrule

%===========================================================%
%                    PUBLIC RELEASED DATA (Corrected)       %
%===========================================================%
\multirow{7}{*}{\rotatebox{90}{\bfseries\textcolor{PublicColor}{Public Released}}}
 & \textit{Medical Multimodal Caption} & PubMedVision\_Alignment\_VQA2 & 646759 \\
 \cmidrule(l){2-4}
 & \textit{Medical Grounded Caption} & MedTrinity161K & 161630 \\
 \cmidrule(l){2-4}
 & \textit{General Multimodal Caption} & LLaVA-ReCap-558K & 558128 \\
 & & pixmo-cap & 706830 \\
 \cmidrule(l){2-4}
 & \textit{General Chart Caption} & processed\_charts\_data & 4000 \\
 \cmidrule(l){2-4}
 & \textit{General Document Caption} & textOCR\_train & 25117 \\
 \cmidrule(l){2-4}
 & \textit{General Text} & Infinity-Instruct & 479997 \\
\midrule
\multicolumn{3}{r}{\textbf{Public Data Subtotal}} & \textbf{2,582,461} \\
\midrule
\rowcolor{SubtotalColor}
\multicolumn{3}{r}{\textbf{GRAND TOTAL}} & \textbf{4850022} \\
\bottomrule
\end{tabular}
\end{table*}

\begin{table*}[!htbp]
\caption{Stage 3 training data composition  (\textasciitilde10.4M entries)}
\label{tab:final_definitive_verified}
\sisetup{group-separator={,}}
\setlength{\tabcolsep}{3pt}
\renewcommand{\arraystretch}{1.15}

% --- LEFT PANEL: TEXT DATA ---
\begin{minipage}[t]{0.48\textwidth}
\centering
\large\textbf{Text Data (5.9M)}
\par\vspace{1ex}
\footnotesize
\begin{tabular}{l l l S[table-format=7.0]}
\toprule
% --- TABLE HEADER ---
\rowcolor{HeaderColor}
\multicolumn{1}{c}{\textcolor{white}{\textbf{ }}} & 
\multicolumn{1}{c}{\textcolor{white}{\textbf{Task}}} & 
\multicolumn{1}{c}{\textcolor{white}{\textbf{Dataset}}} & 
\multicolumn{1}{c}{\textcolor{white}{\textbf{Count}}} \\
\midrule

%--- Medical Text ---
\multirow{14}{*}{\rotatebox{90}{\bfseries\textcolor{MedText}{Medical}}} %<-- FIX: Row span corrected
 & \textit{Factoid QA} & Apollo-Pre & 1859880 \\
 & & MedQuAD & 16407 \\
 \cmidrule(l){2-4}
 & \textit{LongCoT Data} & II-Medical SFT & 700000 \\
 & & ReasonMed & 369983 \\
 \cmidrule(l){2-4}
 & \textit{Reasoning Data} & medical-o1 & 65531 \\
 & & MedReason & 32682 \\
 & & medical-r1 & 22000 \\
 & & MedThought & 7716 \\
 \cmidrule(l){2-4}
 & \textit{Clinical Dialogue} & Miriad (Sampled) & 1255356 \\
 & & HealthCareMagic & 112165 \\
 & & iCliniq & 7321 \\
 \cmidrule(l){2-4}
 & \textit{Medical Instruct} & AlpaCare & 52002 \\
 & & Apollo-SFT & 417241 \\
  \cmidrule(l){2-4}
 & \textit{Multilingual QA} & MMedC & 45048 \\
\midrule
\multicolumn{3}{r}{\textbf{Subtotal}} & \textbf{4963332} \\
\midrule

%--- General Text ---
\multirow{16}{*}{\rotatebox{90}{\bfseries\textcolor{GenText}{General}}} %<-- FIX: Row span corrected
 & \textit{Instruction} & Openhermes & 496743 \\
 & & Glaive-code-assist & 182240 \\
 & & CamelAI & 78390 \\
 & & Metamath & 56448 \\
 & & EvolInstruct\_70k & 51948 \\
 & & Cot\_alpaca\_gpt4 & 42026 \\
 & & Airoboros2.2 & 35380 \\
 & & Platypus & 22280 \\
 & & GPT-4 Comparison & 14928 \\
 & & UnnaturalInstruct & 8610 \\
 & & CogStackMed & 4443 \\
 & & LMSys Chatbot Arena & 3136 \\
 & & Caseus\_custom & 2688 \\
 & & Lmsys1m & 1631 \\
 & & Econ\_domain\_expert & 660 \\
\midrule
\multicolumn{3}{r}{\textbf{Subtotal}} & \textbf{1001551} \\
\bottomrule
\end{tabular}
\end{minipage}
\hfill % Flexible space between panels
%
% --- RIGHT PANEL: MULTIMODAL DATA ---
\begin{minipage}[t]{0.48\textwidth}
\centering
\large\textbf{Multimodal Data (4.5M)}
\par\vspace{1ex}
\footnotesize
\begin{tabular}{l l l S[table-format=7.0]}
\toprule
% --- TABLE HEADER ---
\rowcolor{HeaderColor}
\multicolumn{1}{c}{\textcolor{white}{\textbf{ }}} & 
\multicolumn{1}{c}{\textcolor{white}{\textbf{Task}}} & 
\multicolumn{1}{c}{\textcolor{white}{\textbf{Dataset}}} & 
\multicolumn{1}{c}{\textcolor{white}{\textbf{Count}}} \\
\midrule

%--- Medical Multimodal ---
\multirow{30}{*}{\rotatebox{90}{\bfseries\textcolor{MedMulti}{Medical}}} %<-- FIX: Row span corrected
 & \textit{2D VQA} & PubMedVision & 646750 \\
 & & Generated QA & 594237 \\
 & & PMC-VQA & 152602 \\
 & & MIMIC-CXR-VQA & 77035 \\
 & & PathVQA & 39510 \\   %<-- FIX: Split as requested
 & & SLAKE & 9837 \\      %<-- FIX: Split as requested
 & & RADVQA & 6128 \\    %<-- FIX: Split as requested
 & & GMAI-Reasoning & 7004 \\
 \cmidrule(l){2-4}
 & \textit{Classification} & MedMNIST & 74689 \\
 \cmidrule(l){2-4}
 & \textit{Report Gen.} & MIMIC-CXR-MRG & 242310 \\
 & & CheXpert-MRG & 223228 \\
 & & IU-Xray-MRG & 2365 \\
 \cmidrule(l){2-4}
 & \textit{3D Caption} & M3D-Cap & 31928 \\
 & & CT-Rate-Cap & 47149 \\
 & & RadFM-Cap & 26891 \\
 & & AMOS-Cap & 1286 \\
 \cmidrule(l){2-4}
 & \textit{3D VQA} & M3D-VQA & 84144 \\
 & & RadFM-VQA & 83049 \\
 & & CT-Rate-VQA & 46033 \\
 & & AMOS-VQA & 13735 \\
 \cmidrule(l){2-4}
 & \textit{Video Caption} & Cholec80-Cap & 17010 \\
 & & PSI-AVA-Cap & 1195 \\
 & & EndoVis-Cap & 165 \\
 \cmidrule(l){2-4}
 & \textit{Video QA} & Cholec80-VQA & 24829 \\
 & & PSI-AVA-VQA & 5244 \\
 & & EndoVis-VQA & 4358 \\
 \cmidrule(l){2-4}
 & \textit{Ground QA} & CoPESD & 74561 \\
 \cmidrule(l){2-4}
 & \textit{Interleaved} & Quilt-Instruct & 105745 \\ %<-- FIX: RESTORED
 & & Llava-Med-Instruct & 56408 \\  %<-- FIX: RESTORED
\midrule
\multicolumn{3}{r}{\textbf{Subtotal}} & \textbf{2699370} \\
\midrule

%--- General Multimodal ---
\multirow{11}{*}{\rotatebox{90}{\bfseries\textcolor{GenMulti}{General}}} %<-- FIX: Row span corrected
 & \textit{Instruction} & LLaVA\_NeXT & 779287 \\
 \cmidrule(l){2-4}
 & \textit{VQA} & PixMo-QA & 268309 \\ %<-- FIX: Rebuilt this section
 \cmidrule(l){2-4}
 & \textit{Interleaved} & Llava-Interleaved & 36541 \\
 & & Mantis & 696781 \\
 \cmidrule(l){2-4}
 & \textit{Video QA} & NextQA & 3870 \\
 & & STAR & 3032 \\
 \cmidrule(l){2-4}
 & \textit{3D Imaging} & Embodied 3D & 4989 \\
\midrule
\multicolumn{3}{r}{\textbf{Subtotal}} & \textbf{1792809} \\
\bottomrule
\end{tabular}
\end{minipage}

\vspace{1em}

% --- GRAND TOTAL BAR ---
\begin{center}
\begin{tabular}{p{0.97\textwidth}}
\rowcolor{SubtotalColor}
\hfill \textbf{GRAND TOTAL} \hfill \textbf{10,457,117} \hfill \\
\end{tabular}
\end{center}

\end{table*}
\FloatBarrier

\begin{table}[h!]
\centering
\caption{Prompt templates for different task types during inference}
\label{tab:prompt_templates}
\begin{tabular}{@{}l p{5.5cm} p{5.5cm}@{}}
\toprule
\textbf{Task Type} & \textbf{Prompt for Direct Answering} & \textbf{Prompt for CoT Reasoning} \\
\midrule
\textbf{Multiple-Choice} & 
\texttt{Question: \{Question\}}\newline
\texttt{Options:}\newline
\texttt{\{Options\}}\newline
Answer with the option's letter from the given choices directly. & 
\texttt{Question: \{Question\}}\newline
\texttt{Options:}\newline
\texttt{\{Options\}}\newline
Please reason step by step, and put your final answer within \texttt{\textbackslash boxed\{\}}. \\
\addlinespace

\textbf{Judgement} & 
\texttt{\{Question\}}\newline
Please output 'yes' or 'no' (no extra output). & 
\texttt{\{Question\}}\newline
Please reason step by step, and put your final answer within \texttt{\textbackslash boxed\{\}}. \\
\addlinespace

\textbf{Close-Ended} & 
\texttt{\{Question\}}\newline
Answer the question using a single word or phrase. & 
\texttt{\{Question\}}\newline
Please reason step by step, and put your final answer within \texttt{\textbackslash boxed\{\}}. \\
\addlinespace

\textbf{Open-Ended} & 
\texttt{\{Question\}}\newline
Please answer the question concisely. & 
\texttt{\{Question\}}\newline
Please reason step by step, and put your final answer within \texttt{\textbackslash boxed\{\}}. \\
\addlinespace

\textbf{Report Generation} & 
\multicolumn{2}{p{11.5cm}}{You are a helpful assistant. Please generate a report for the given images, including both findings and impressions. Return the report in the following format: \texttt{Findings: \{\} Impression: \{\}}.} \\
\bottomrule
\end{tabular}
\end{table}

\begin{table*}[ht]
\centering
\caption{Performance comparison on MedFrameQA. For each metric (column), the best and second-best results are highlighted in \textbf{bold} and with an \underline{underline}, respectively. *SD* indicates the standard deviation of accuracy across frame counts (2–5), reflecting prediction stability.}
\label{tab:model_accuracy_updated}
\resizebox{\textwidth}{!}{% 
\begin{tabular}{@{}lcccccccccc@{}}
\toprule
\textbf{Model} & \multicolumn{5}{c}{\textbf{Accuracy (\%) by Frame Count}} & \multicolumn{5}{c}{\textbf{Accuracy (\%) by Modality}} \\
\cmidrule(lr){2-6} \cmidrule(lr){7-11}
 & \textbf{2} & \textbf{3} & \textbf{4} & \textbf{5} & \textbf{SD} & \textbf{CT} & \textbf{MRI} & \textbf{Ultrasound} & \textbf{X-ray} & \textbf{Other} \\
\midrule
o1 & 48.16 & 45.64 & 51.43 & 48.15 & 2.37 & 48.98 & 45.40 & 49.05 & 49.16 & 51.64 \\
o3 & 50.00 & 47.46 & 53.60 & 51.38 & 2.57 & 50.09 & 48.57 & 51.45 & 53.06 & 52.38 \\
o4-mini & 50.21 & 46.23 & 50.00 & 50.37 & 1.99 & 48.08 & 48.85 & 52.34 & 50.33 & 53.49 \\
Gemini-2.5-Flash & 53.54 & 55.48 & 55.47 & 55.76 & 1.02 & 54.57 & 53.60 & 57.36 & 58.14 & 49.24 \\
QvQ-72B-Preview & 46.88 & 45.91 & 46.48 & 46.69 & 0.42 & 45.45 & 45.24 & 50.65 & 44.85 & 57.58 \\
GPT-4-Turbo-V & 47.47 & 45.51 & 46.88 & 46.34 & 0.83 & 46.83 & 43.48 & 50.65 & 49.17 & 51.52 \\
GPT-4o & 47.30 & 45.18 & 40.23 & 45.35 & 3.01 & 45.52 & 43.27 & 48.58 & 47.51 & 51.52 \\
GPT-4o-mini & 35.16 & 36.21 & 32.42 & 33.09 & 1.77 & 35.26 & 34.31 & 34.88 & 34.55 & 29.55 \\
Claude-3.7-Sonnet & 49.41 & 48.01 & 51.56 & 50.68 & 1.55 & 50.75 & 49.11 & 49.10 & 49.83 & 46.21 \\
Qwen2.5-VL-72B-Instruct & 41.99 & 40.40 & 38.67 & 40.32 & 1.36 & 38.99 & 40.73 & 42.38 & 42.52 & 49.24 \\
\midrule
\rowcolor{ColorMedical!60}
\textbf{Hulu-Med-7B} & 55.14 & 57.31 & \underline{57.42} & 58.98 & 1.47 & 55.69 & 55.16 & \underline{59.43} & \underline{63.12} & \underline{57.58} \\
\rowcolor{ColorMedical!60}
\textbf{Hulu-Med-14B} & \textbf{60.29} & \textbf{60.63} & \textbf{57.81} & \textbf{59.85} & 1.26 & \textbf{59.89} & \underline{58.29} & 59.17 & \textbf{63.46} & \textbf{68.18} \\
\rowcolor{ColorMedical!60}
\textbf{Hulu-Med-32B} & \underline{58.77} & \underline{59.14} & \underline{57.42} & \underline{59.48} & 0.80 & \underline{58.58} & \textbf{58.39} & \textbf{61.76} & 58.80 & \underline{57.58} \\
\bottomrule
\end{tabular}%
}
\end{table*}

\begin{table}[ht]
\centering
\caption{Data availability and licenses for datasets used in our study. ``Access'' directly lists the dataset license. Synthetically generated datasets and those requiring specific permissions are marked as \textit{Credentialed Access}.}
\label{tab:dataset_license_full}
\definecolor{StageZero}{HTML}{d7f9f8}
\definecolor{StageOne}{HTML}{ffffea}
\definecolor{StageTwo}{HTML}{fff0d4}
\adjustbox{max width=1.0\textwidth}{
\begin{tabular}{l l l}
\toprule
\textbf{Dataset Name} & \textbf{Link} & \textbf{Access} \\
\midrule
\rowcolor{StageZero}\multicolumn{3}{l}{\textbf{Stage 1}} \\
\rowcolor{StageZero} BIOMEDICA Clinical Subset (medical multimodal) & https://minwoosun.github.io/biomedica-website/ & Under CC \\
\rowcolor{StageZero} Medicat (medical multimodal) & https://github.com/allenai/medicat & PhysioNet License \\
\rowcolor{StageZero} MedPix 2.0 (medical multimodal) & https://huggingface.co/datasets/CHILab1/MedPix-2.0 & CC BY-NC-SA 4.0 \\
\rowcolor{StageZero}  Quilt-Pretrain (medical multimodal) & https://huggingface.co/datasets/wisdomik/Quilt-LLaVA-Pretrain & CC BY 4.0 \\
\rowcolor{StageZero}  ROCOv2 (medical multimodal) & https://huggingface.co/datasets/eltorio/ROCOv2-radiology & CC BY 4.0 \\
\midrule
\rowcolor{StageOne}\multicolumn{3}{l}{\textbf{Stage 2}} \\
\rowcolor{StageOne} biomedica\_clinical\_recaption (medical multimodal) &Synthetic Data  & Credentialed Access \\
\rowcolor{StageOne} biomedica\_dermatology\_recaption (medical multimodal) &Synthetic Data  & Credentialed Access \\
\rowcolor{StageOne} biomedica\_histopathology\_recaption (medical multimodal) &Synthetic Data  & Credentialed Access \\
\rowcolor{StageOne} biomedica\_microscopy\_recaption (medical multimodal) &Synthetic Data  & Credentialed Access \\
\rowcolor{StageOne} biomedica\_surgery\_recaption (medical multimodal) &Synthetic Data  & Credentialed Access \\
\rowcolor{StageOne} Dermoscopy\_SyntheticCap (medical multimodal) &Synthetic Data  & Credentialed Access \\
\rowcolor{StageOne} Fundus\_OCT\_SyntheticCap (medical multimodal) &Synthetic Data  & Credentialed Access \\
\rowcolor{StageOne} LLaVA-ReCap-558K (general multimodal) & https://huggingface.co/datasets/lmms-lab/LLaVA-ReCap-558K & CC BY 4.0 \\
\rowcolor{StageOne} medicat\_recaption (medical multimodal) &Synthetic Data  & Credentialed Access \\
\rowcolor{StageOne} medmnist\_generated\_captions (medical multimodal) &Synthetic Data  & Credentialed Access \\
\rowcolor{StageOne} MedTrinity161K (medical multimodal) & https://proceedings.iclr.cc/paper\_files/paper/2025/hash/11c483499c285f30daf832c17dc752bd-Abstract-Conference.html & Unknown \\
\rowcolor{StageOne} Microscopy\_SyntheticCap (medical multimodal) &Synthetic Data  & Credentialed Access \\
\rowcolor{StageOne} mimic-pretrain-recaption (medical multimodal) &Synthetic Data  & Credentialed Access \\
\rowcolor{StageOne} pixmo-cap (general multimodal) & https://huggingface.co/datasets/allenai/pixmo-cap & odc-by \\
\rowcolor{StageOne} processed\_charts\_data (general multimodal-Chart) &https://huggingface.co/datasets/LeroyDyer/chart\_text\_to\_Base64  & MIT \\
\rowcolor{StageOne} PubMedVision\_Alignment (medical multimodal) & https://huggingface.co/datasets/FreedomIntelligence/PubMedVision & CC BY 4.0 \\
\rowcolor{StageOne} Rad-Slake-Pvqa-SyntheticCap (medical multimodal) &Synthetic Data  & Credentialed Access \\
\rowcolor{StageOne} Radimagenet\_SyntheticCap-Ultrasound (medical multimodal) &Synthetic Data  & Credentialed Access \\
\rowcolor{StageOne} ROCOv2-radiology-recap (medical multimodal) & https://huggingface.co/datasets/eltorio/ROCOv2-radiology & CC BY 4.0 \\
\rowcolor{StageOne} TextOCR (general multimodal-Scene Text Image) & https://www.kaggle.com/datasets/robikscube/textocr-text-extraction-from-images-dataset & MIT \\
\rowcolor{StageOne} Ultrasound\_SyntheticCap (medical multimodal) &Synthetic Data  & Credentialed Access \\
% --- Added: missing Stage 1 items from your list ---
\rowcolor{StageOne} Mimic-recaption (medical multimodal) & Synthetic Data & Credentialed Access \\
\rowcolor{StageOne} IUXray-recaption (medical multimodal) & Synthetic Data & Credentialed Access \\
\rowcolor{StageOne} InfInstruct (general text) & https://huggingface.co/datasets/BAAI/Infinity-Instruct & CC BY SA 4.0 \\
\midrule
\rowcolor{StageTwo}\multicolumn{3}{l}{\textbf{Stage 3}} \\
\rowcolor{StageTwo} AlpaCare-MedInstruct-52k (medical text) & https://huggingface.co/datasets/lavita/AlpaCare-MedInstruct-52k & CC BY 4.0 \\
\rowcolor{StageTwo} ChatDoctor-HealthCareMagic-100k (medical text) & https://huggingface.co/datasets/lavita/ChatDoctor-HealthCareMagic-100k & CC BY 4.0 \\
\rowcolor{StageTwo} GMAI-Reasoning10K (medical multimodal) & https://huggingface.co/datasets/General-Medical-AI/GMAI-Reasoning10K & CC BY 4.0 \\
\rowcolor{StageTwo} iCliniq-10K (medical text) & https://huggingface.co/datasets/zhengComing/iCliniq-10K & CC BY 4.0 \\
\rowcolor{StageTwo} LLaVA-Med (interleaved) (medical multimodal) & https://github.com/microsoft/LLaVA-Med & CC BY 4.0 \\
\rowcolor{StageTwo} LLaVA-NeXT-SFT (general multimodal) & https://huggingface.co/datasets/lmms-lab/LLaVA-NeXT-Data & Apache 2.0 \\
\rowcolor{StageTwo} Mantis-Instruct (interleaved) (general multimodal) & https://huggingface.co/datasets/TIGER-Lab/Mantis-Instruct & Apache 2.0 \\
\rowcolor{StageTwo} Medical-o1 (medical text) & https://huggingface.co/datasets/FreedomIntelligence/medical-o1-verifiable-problem & CC BY 4.0 \\
\rowcolor{StageTwo} Medical-R1-Distill (medical text) & https://huggingface.co/datasets/FreedomIntelligence/Medical-R1-Distill-Data & CC BY 4.0 \\
\rowcolor{StageTwo} MedQuAD (medical text) & https://huggingface.co/datasets/lavita/MedQuAD & CC BY 4.0 \\
\rowcolor{StageTwo} MedReason (medical text) & https://huggingface.co/datasets/UCSC-VLAA/MedReason & CC BY 4.0 \\
\rowcolor{StageTwo} MedThoughts-8K (medical text) & https://huggingface.co/datasets/hw-hwei/MedThoughts-8K & MIT \\
\rowcolor{StageTwo} Miriad (20\% sample) (medical text) & https://huggingface.co/miriad & Apache 2.0 \\
\rowcolor{StageTwo} OpenHermes-2.5 (general text) & https://huggingface.co/datasets/teknium/OpenHermes-2.5|huggingface.co/datasets/Replete-AI/OpenHermes-2.5-Filtered & Apache 2.0 \\
\rowcolor{StageTwo} PixMo-QA (general multimodal) & https://huggingface.co/datasets/allenai/pixmo-cap & ODC-BY v1.0 \\
\rowcolor{StageTwo} PubMedVision-SFT (medical multimodal) & https://huggingface.co/datasets/FreedomIntelligence/PubMedVision & CC BY 4.0 \\
\rowcolor{StageTwo} QUILT-Instruct (medical multimodal) & https://huggingface.co/datasets/wisdomik/QUILT-LLaVA-Instruct-107K & CC BY 4.0 \\
\rowcolor{StageTwo} ReasonMed (medical text) & https://huggingface.co/datasets/lingshu-medical-mllm/ReasonMed & Apache 2.0 \\
\rowcolor{StageTwo} Synthetic-QA (medical multimodal) &Synthetic Data  & Credentialed Access \\
% --- Added: missing Stage 2 items from your list ---
\rowcolor{StageTwo} Apollo (medical text) & https://huggingface.co/datasets/FreedomIntelligence/ApolloCorpus & Apache 2.0 \\
\rowcolor{StageTwo} II-Medical-Reasoning-SFT (medical text) & https://huggingface.co/datasets/Intelligent-Internet/II-Medical-Reasoning-SFT& Open Access \\
\rowcolor{StageTwo} Multilingual COT (medical text) & Synthetic Data & Credentialed Access \\
\rowcolor{StageTwo} LLaVA-Next-Interleaved (general multimodal) & https://huggingface.co/datasets/lmms-lab/LLaVA-NeXT-Interleave-Bench & CC BY 4.0 \\
% \midrule
% \rowcolor{StageTwo}\multicolumn{3}{l}{\textbf{Stage 3}} \\
\rowcolor{StageTwo} AMOS-MRG (medical multimodal) & https://huggingface.co/datasets/mrmrx/CADS-dataset/blob/0d144b4c8c487d1337e80cae1762a501451349a2/0038\_amos/README\_0038\_amos.md & CC BY-NC-SA \\
\rowcolor{StageTwo} AMOS-VQA (medical multimodal) & https://huggingface.co/datasets/mrmrx/CADS-dataset/blob/0d144b4c8c487d1337e80cae1762a501451349a2/0038\_amos/README\_0038\_amos.md & CC BY-NC-SA \\
\rowcolor{StageTwo} CheXpert (medical multimodal) & https://aimi.stanford.edu/datasets/chexpert-plus & PhysioNet License \\
\rowcolor{StageTwo} Cholec80-Cap (medical multimodal) & https://camma.unistra.fr/datasets & CC BY 4.0 \\
\rowcolor{StageTwo} Cholec80-VQA (medical multimodal) & https://camma.unistra.fr/datasets & CC BY 4.0 \\
\rowcolor{StageTwo} CT-RATE-MRG (medical multimodal) & https://huggingface.co/datasets/ibrahimhamamci/CT-RATE & CC BY-NC-SA 4.0 \\
\rowcolor{StageTwo} CT-RATE-VQA (medical multimodal) & https://huggingface.co/datasets/ibrahimhamamci/CT-RATE & CC BY-NC-SA 4.0 \\
\rowcolor{StageTwo} Endovis-18-Cap (medical multimodal) & https://github.com/lalithjets/Surgical\_VQA & CC BY-NC-SA \\
\rowcolor{StageTwo} Endovis-18-VQA (medical multimodal) & https://github.com/lalithjets/Surgical\_VQA & CC BY-NC-SA \\
% \rowcolor{StageTwo} Generated QA (medical multimodal) &  & Proprietary \\
\rowcolor{StageTwo} IU-Xray (medical multimodal) & https://openi.nlm.nih.gov & Open Access \\
\rowcolor{StageTwo} M3D-MRG (medical multimodal) & https://github.com/BAAI-DCAI/M3D & Apache 2.0 \\
\rowcolor{StageTwo} M3D-VQA (medical multimodal) & https://huggingface.co/datasets/GoodBaiBai88/M3D-VQA & Apache 2.0 \\
% \rowcolor{StageTwo} Med-VQA 2019 (medical multimodal) & https://www.kaggle.com/datasets/claudiopisa9884/medical-vqa-imageclef-2019 & Unknown \\
\rowcolor{StageTwo} MedMNIST (medical multimodal) & https://huggingface.co/datasets/albertvillanova/medmnist-v2 & CC BY 4.0 \\
\rowcolor{StageTwo} MIMIC-CXR (medical multimodal) & https://physionet.org/content/mimic-cxr & PhysioNet License \\
\rowcolor{StageTwo} MIMIC-CXR-VQA (medical multimodal) & https://github.com/baeseongsu/mimic-cxr-vqa & MIT license \\
\rowcolor{StageTwo} nextqa-star-scanframe &https://huggingface.co/datasets/ShareGPTVideo/train\_video\_and\_instruction  & MIT license \\
\rowcolor{StageTwo} PMC-VQA (medical multimodal) & https://huggingface.co/datasets/RadGenome/PMC-VQA & CC BY-NC-SA 4.0 \\
\rowcolor{StageTwo} PSI-AVA-Cap (medical multimodal) & https://github.com/BCV-Uniandes/TAPIR & MIT \\
\rowcolor{StageTwo} PSI-AVA-VQA (medical multimodal) & https://github.com/BCV-Uniandes/TAPIR & MIT \\
% --- Added: missing Stage 3 items from your list (kept in StageTwo color block) ---
\rowcolor{StageTwo} RadVQA Rewriting (medical multimodal) & Synthetic Data & Credentialed Access \\
\rowcolor{StageTwo} SLAKE Rewriting (medical multimodal) & Synthetic Data & Credentialed Access \\
\rowcolor{StageTwo} PathVQA Rewriting (medical multimodal) & Synthetic Data & Credentialed Access \\
\rowcolor{StageTwo} RP3D-VQA (medical multimodal) & https://github.com/chaoyi-wu/RadFM & Credentialed Access \\
\rowcolor{StageTwo} RP3D-MRG (medical multimodal) & https://github.com/chaoyi-wu/RadFM & Credentialed Access\\
\rowcolor{StageTwo} CoPESD (medical multimodal) & https://github.com/gkw0010/CoPESD & Apache 2.0 \\
\bottomrule
\end{tabular}}
\end{table}

\begin{figure}[!ht]
\centering
\includegraphics[width=1\linewidth]{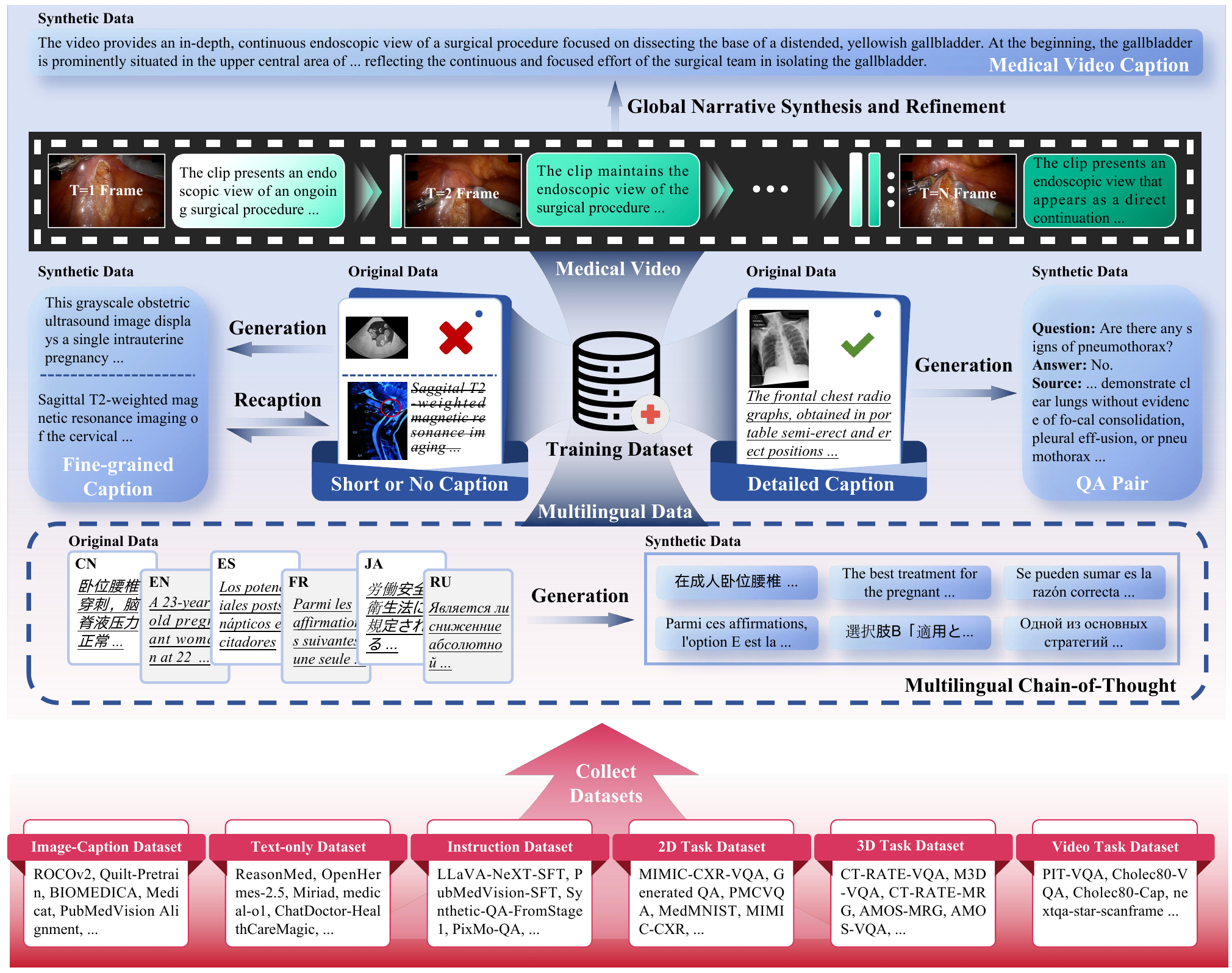}
\caption{Overivew of data synthetic strategy.}
\label{fig:data_syn}
\end{figure}

\begin{figure*}[!ht]
\centering
\includegraphics[width=1\linewidth]{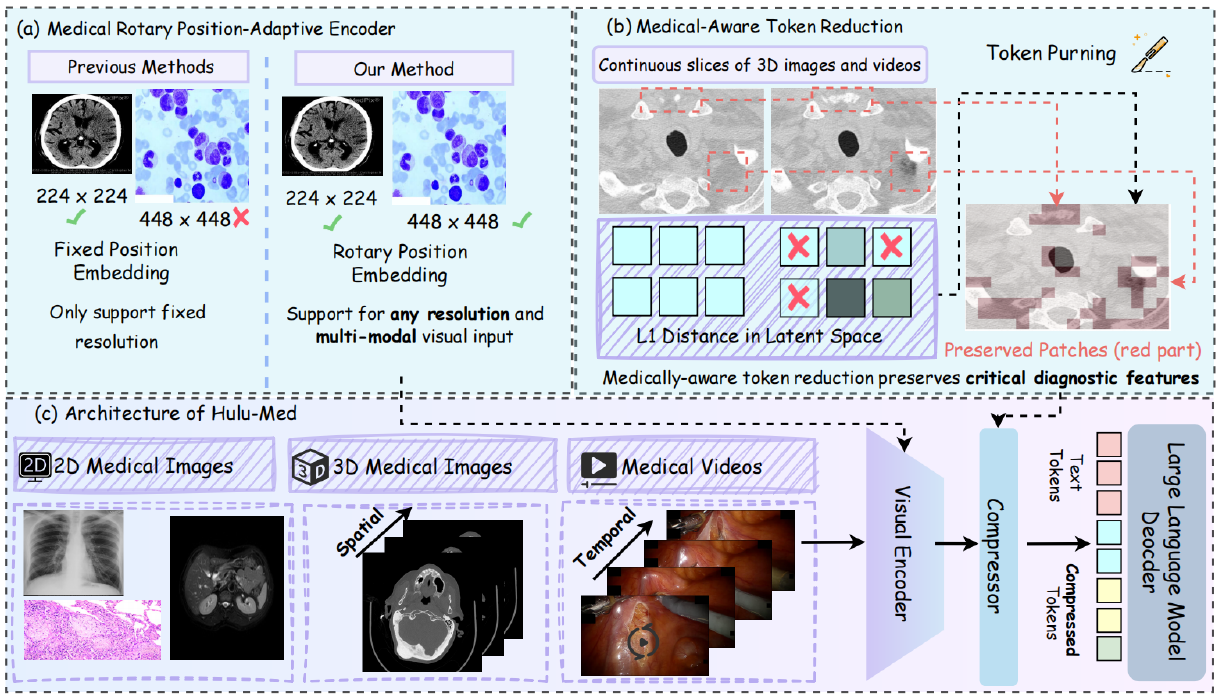}
\caption{An overview of Hulu-Med. The framework consists of three key components: (a). a medical visual encoder supporting arbitrary resolutions and modalities, (b). Medically-Guided Token Reduction to efficiently handle redundant frames and slices in videos and 3D images, and (c). the architecture of our Hulu-Med model.}
\label{fig:framework}
\end{figure*}

\begin{figure}[!ht]
\centering
\includegraphics[width=1\linewidth]{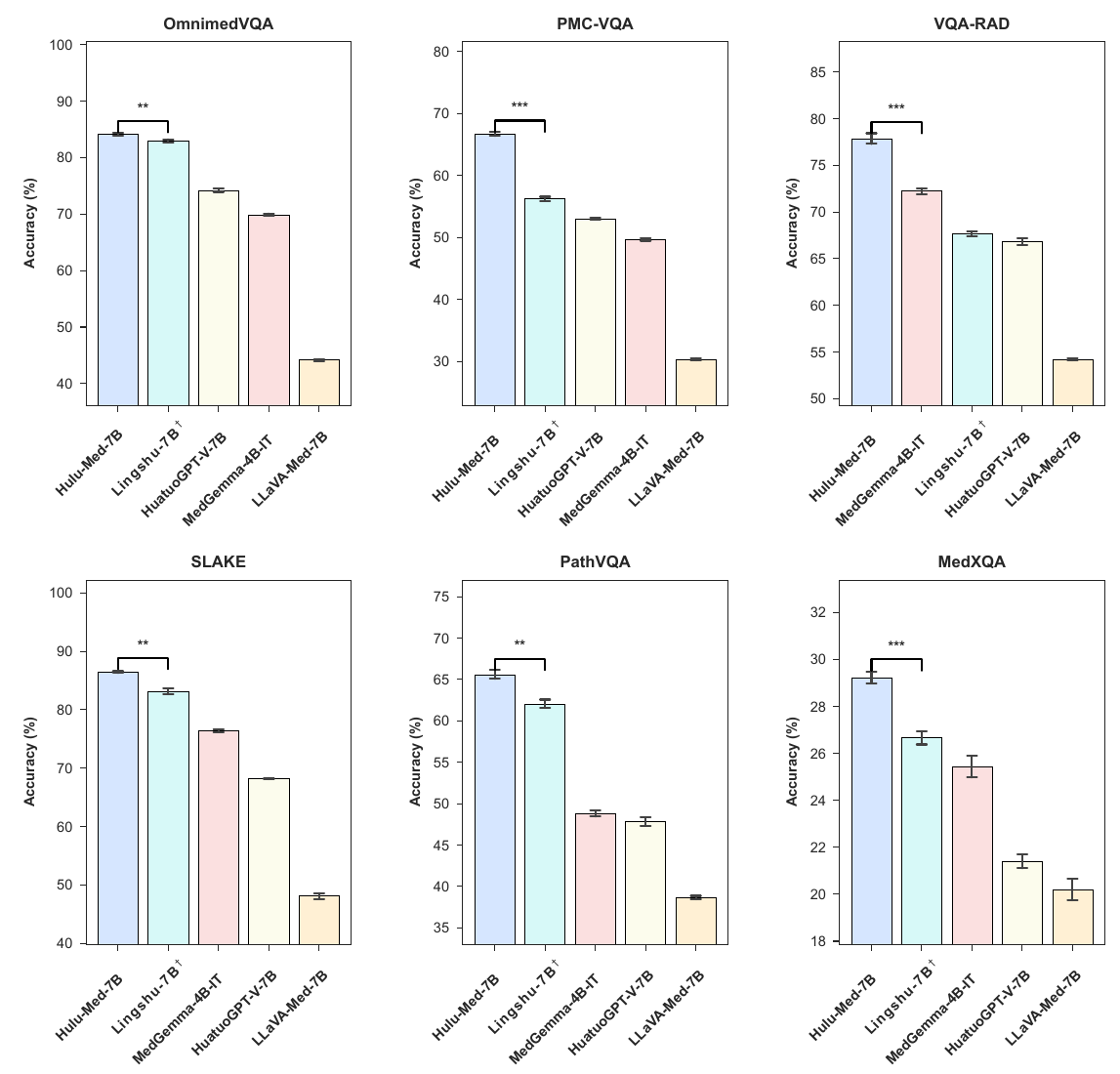}
\caption{Performance comparison of 7B-scale VLMs on medical multimodal benchmarks. All experiments were conducted over three random seeds with a temperature setting of 0.6. Evaluation on MMMU was not included due to submission limits imposed by the EvalAI platform (https://eval.ai/).}
\label{extendedfigure:mm_bench_pvalue}
\end{figure}

\begin{figure*}[!p]
    \centering
    \includegraphics[width=1\linewidth]{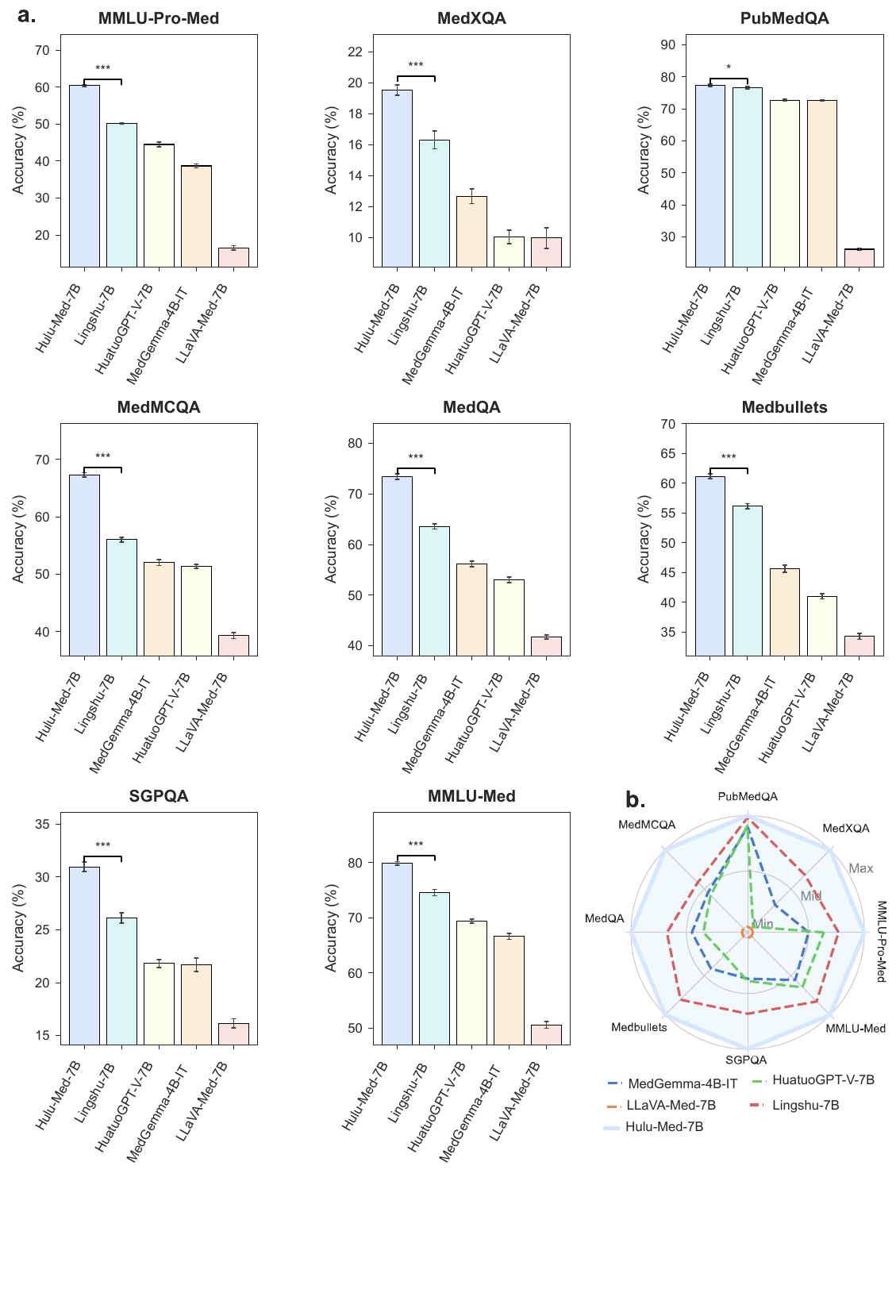}
\end{figure*}
\clearpage
\begin{center}
    \parbox{1\textwidth}{
         \captionof{figure}{\textbf{Evaluation of Hulu-Med's performance in text medical benchmarks.} 
        \textbf{a}, Performance comparison of 7B-scale VLMs on eight medical text benchmarks. Each result was averaged over three random runs with a decoding temperature of 0.6. MedQA, MedXQA, and SGPQA denote the MedQA-USMLE, MedXpertQA-Text, and SuperGPQA-Medical benchmarks, respectively.
    \textbf{b}, Overall comparison of model performance across the 8 medical text benchmarks.}
\label{extendedfigure:text_bench_pvalue}
    }
\end{center}

\begin{figure*}[!ht]
    \centering
    \includegraphics[width=1\linewidth]{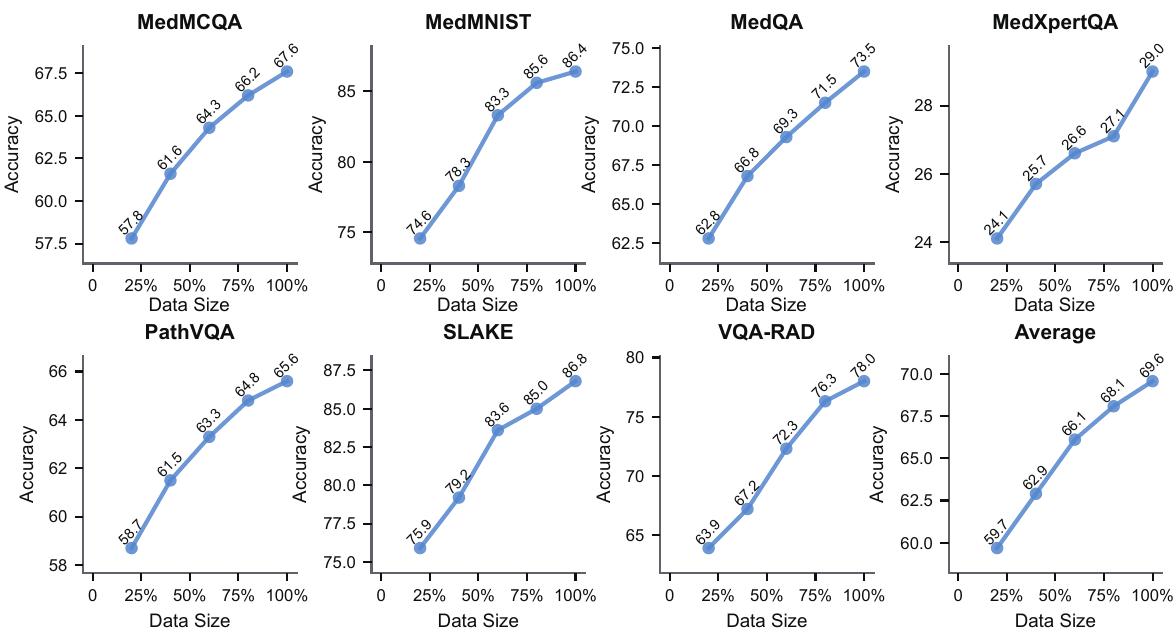}
    \caption{The model's performance consistently improves as the proportion of training data increases, demonstrating the effectiveness of scaling data for enhancing model capabilities}
    \label{fig:scaling_ana_detailed}
\includegraphics[width=1\linewidth]{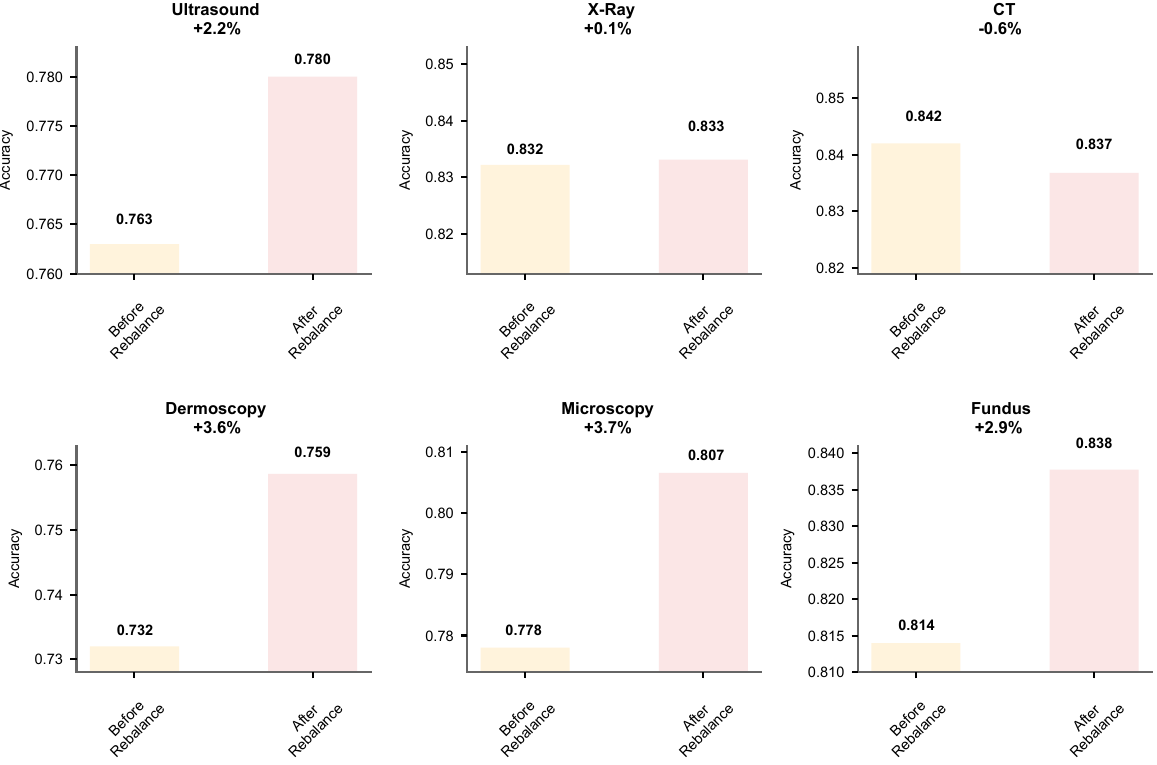}
\caption{Implementing a re-balancing strategy enhances model performance on rare modalities while still maintaining strong capabilities on common modalities.
}
\label{fig:rebalancing}
\end{figure*}

% \begin{figure*}[!t]
% \centering
% \includegraphics[width=1\linewidth]{Longcot_SyntheticCap.pdf}
% \caption{}
% \label{fig:longcot_ana}
% \end{figure*}

% \begin{figure*}[!t]
% \centering
% \includegraphics[width=1\linewidth]{Data_Ratio.pdf}
% \caption{}
% \label{fig:data_ratio}
% \end{figure*}

% \begin{figure*}[!t]
% \centering
% \includegraphics[width=1\linewidth]{Training_Strategy.pdf}
% \caption{}
% \label{fig:training_strategy}
% \end{figure*}

\begin{figure*}[!t]
\centering
\includegraphics[width=1\linewidth]{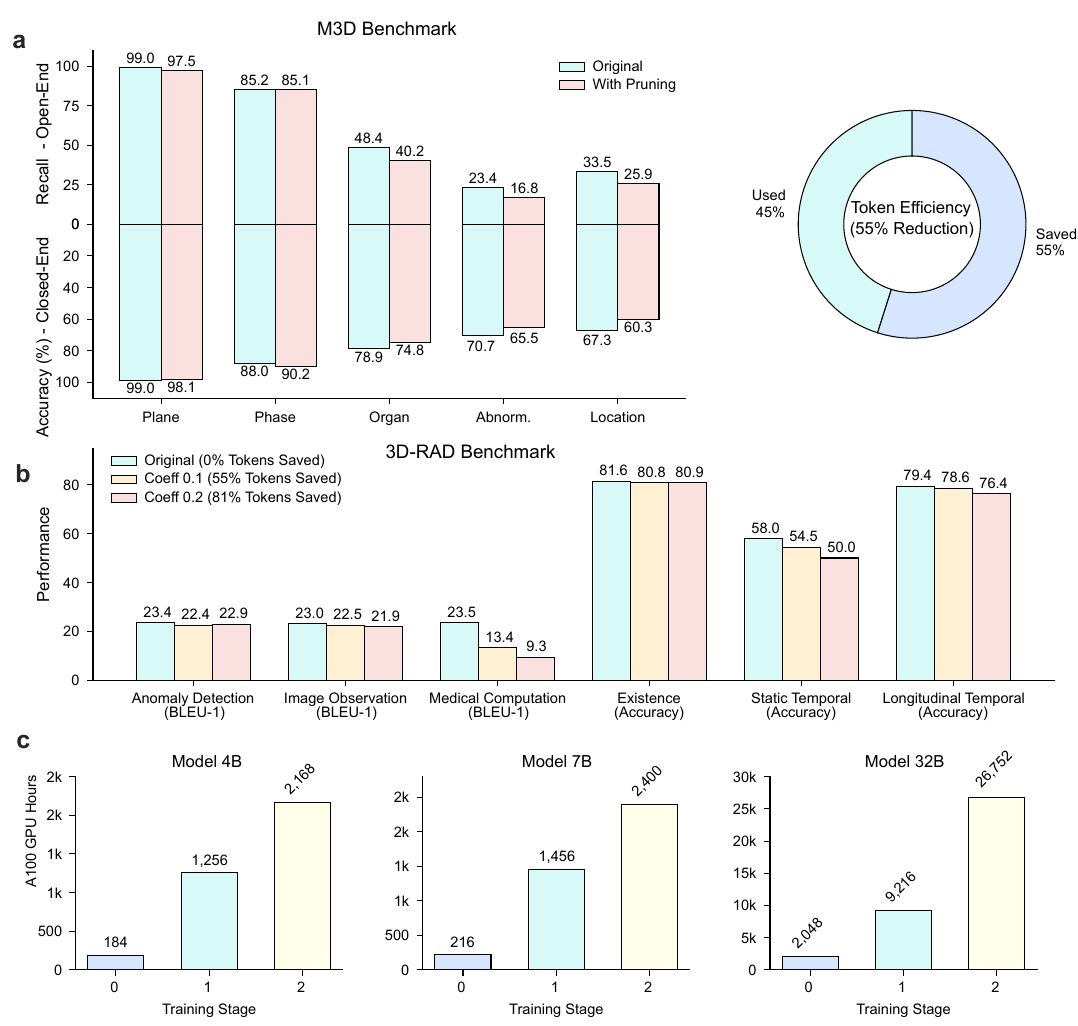}
\caption{
\textbf{a}. Model performance on the M3D benchmark after token pruning, together with the corresponding proportion of pruned tokens. 
\textbf{b}. Sub-task performance on 3D-RAD across different pruning coefficients ($\tau$). 
\textbf{c}. Computational training cost for models at 4B, 7B, and 32B parameter scales.
}
\label{fig:enxtended3_aba_efficiency}
\end{figure*}

\begin{figure*}[!t]
\centering
\includegraphics[width=1\linewidth]{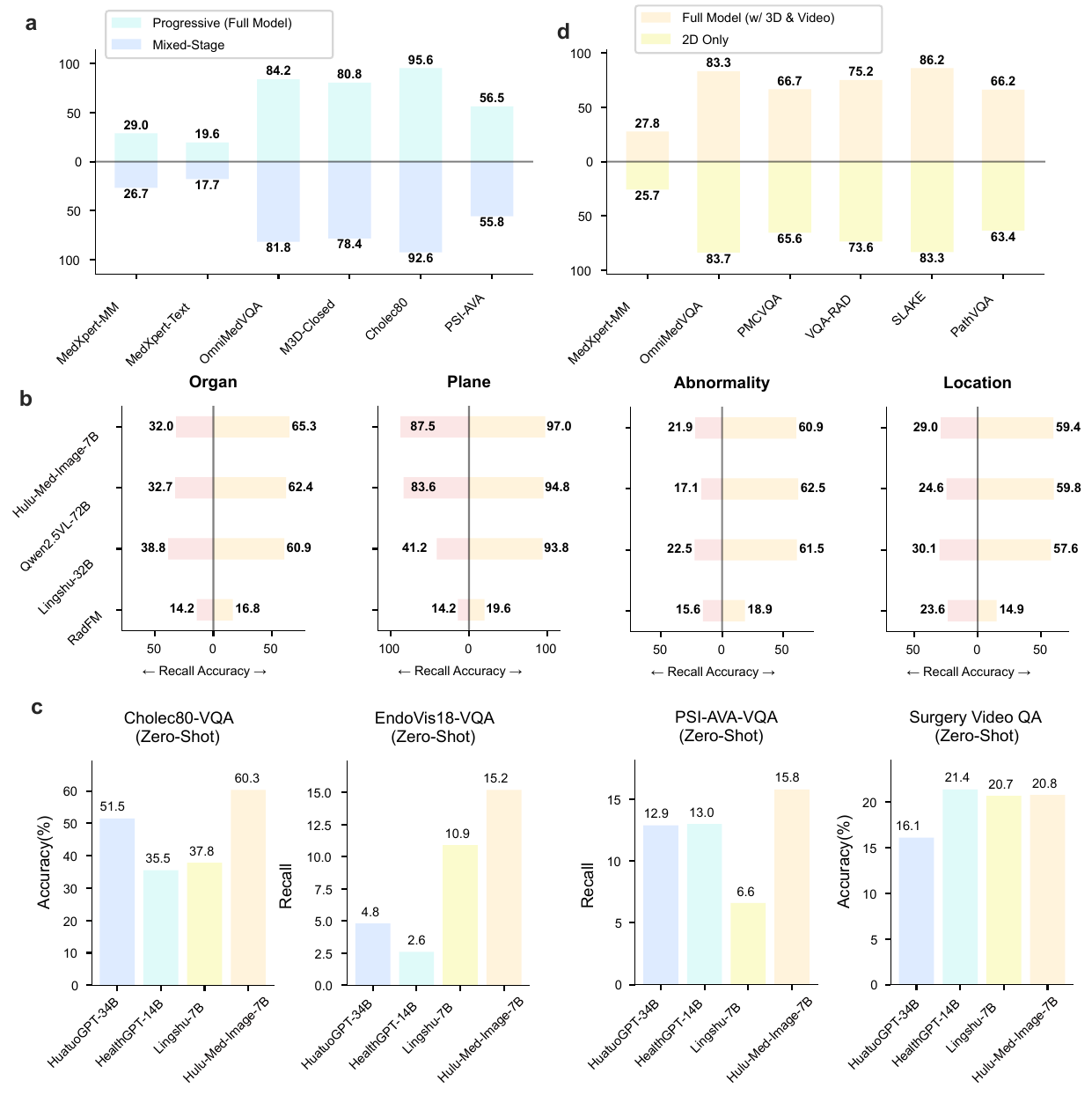}
\caption{
          \textbf{a}, The superiority of our progressive curriculum is confirmed by showing that it consistently outperforms a mixed-stage training approach, which is subject to significant performance drops, thereby validating the hierarchical learning strategy.
          \textbf{b,c}, The model demonstrates powerful emergent cross-modal capabilities, where a version trained exclusively on 2D data achieves competitive results on both 3D volumetric (b) and dynamic video (c) benchmarks, rivaling much larger, specialized models and highlighting the synergistic benefits of diverse multimodal training.
          \textbf{d}, Comparison of Stage 3 training with and without 3D and video data demonstrates that incorporating 3D and video modalities does not compromise 2D performance; on the contrary, it further enhances 2D learning.
}
\label{fig:enxtended3_aba_training_strategy}
\end{figure*}

% \begin{table*}[htbp]
% \centering
% \caption{Comprehensive Healthbench Performance Comparison of Hulu-Med Family and Baselines}
% \label{tab:healthbench_final_comparison}
% \begin{adjustbox}{max width=\textwidth}
% \begin{tabular}{lrrrrrr} % l for metric, and 6 'r' for the 6 model scores
% \toprule
% \textbf{Metric} & \textbf{HuatuoGPT-34B} & \textbf{Lingshu-7B} & \textbf{Lingshu-32B} & \textbf{Hulu-Med-7B} & \textbf{Hulu-Med-14B} & \textbf{Hulu-Med-32B} \\
% \midrule
% Overall Score                  & 0.1717 & 0.1590 & 0.1904 & 0.3831 & 0.3987 & \textbf{0.4158} \\
% Global Health                  & 0.1115 & 0.1084 & 0.1678 & 0.3151 & 0.3474 & \textbf{0.3793} \\
% Communication                  & 0.1661 & 0.1547 & 0.1506 & 0.4062 & 0.4156 & \textbf{0.4329} \\
% Context Seeking                & 0.0611 & 0.0209 & 0.0575 & 0.3180 & \textbf{0.3494} & 0.3428 \\
% Emergency Referrals            & 0.2611 & 0.2576 & 0.2493 & 0.5298 & 0.5541 & \textbf{0.5558} \\
% Hedging                        & 0.2533 & 0.2569 & 0.2927 & 0.4721 & 0.4731 & \textbf{0.4918} \\
% Health Data Tasks              & 0.2288 & 0.1858 & 0.2425 & 0.3677 & 0.3729 & \textbf{0.3798} \\
% Complex Responses              & 0.1157 & 0.1023 & 0.1427 & 0.2103 & 0.2104 & \textbf{0.2551} \\
% \bottomrule
% \end{tabular}
% \end{adjustbox}
% \end{table*}

\begin{figure*}[!t]
\centering
\includegraphics[width=1\linewidth]{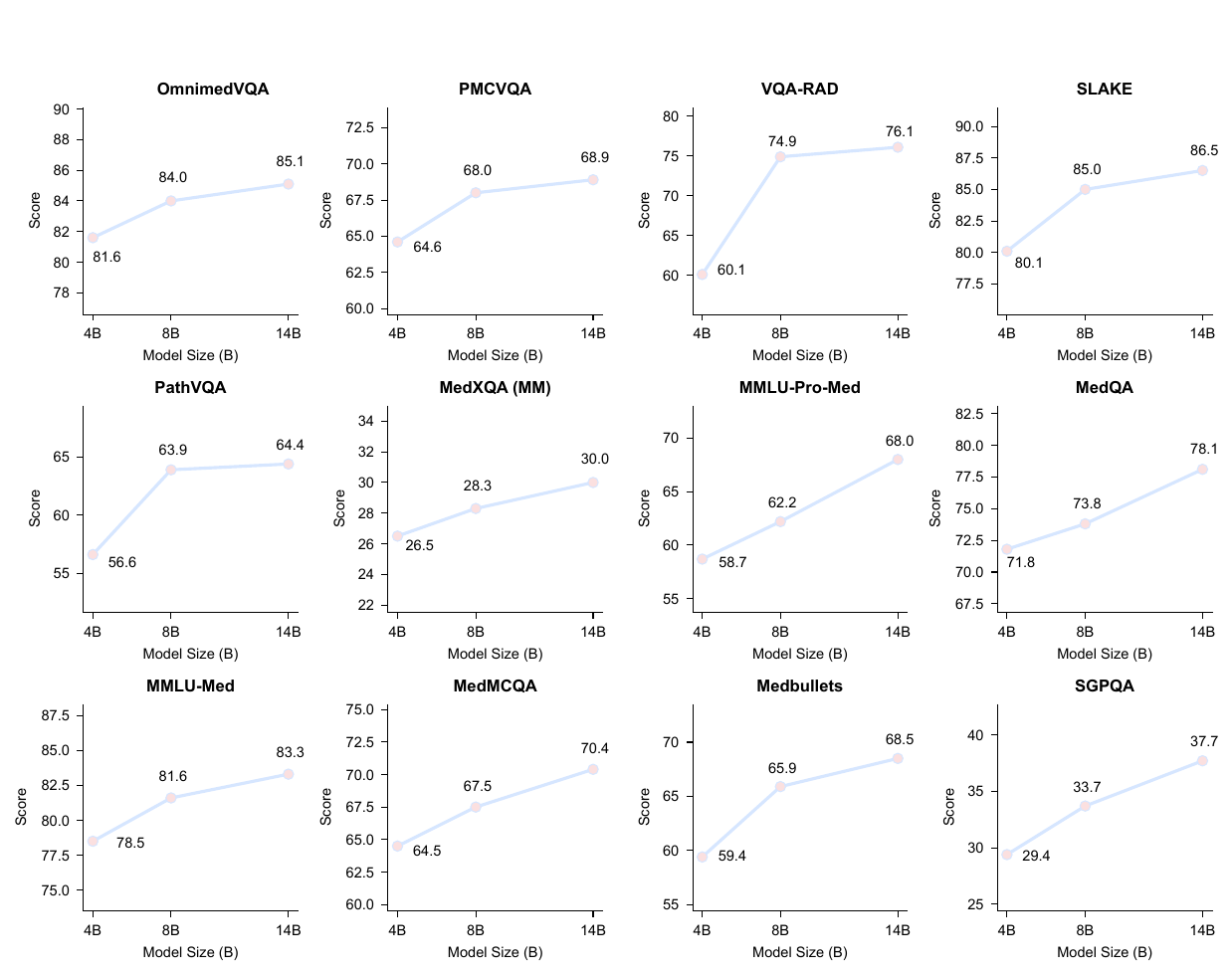}
\caption{Performance scaling with increasing model parameters on the Qwen3 series LLM backbone.}
\label{fig:abalation_llmsize}
\end{figure*}

\begin{figure*}[!t]
\centering
\includegraphics[width=1\linewidth]{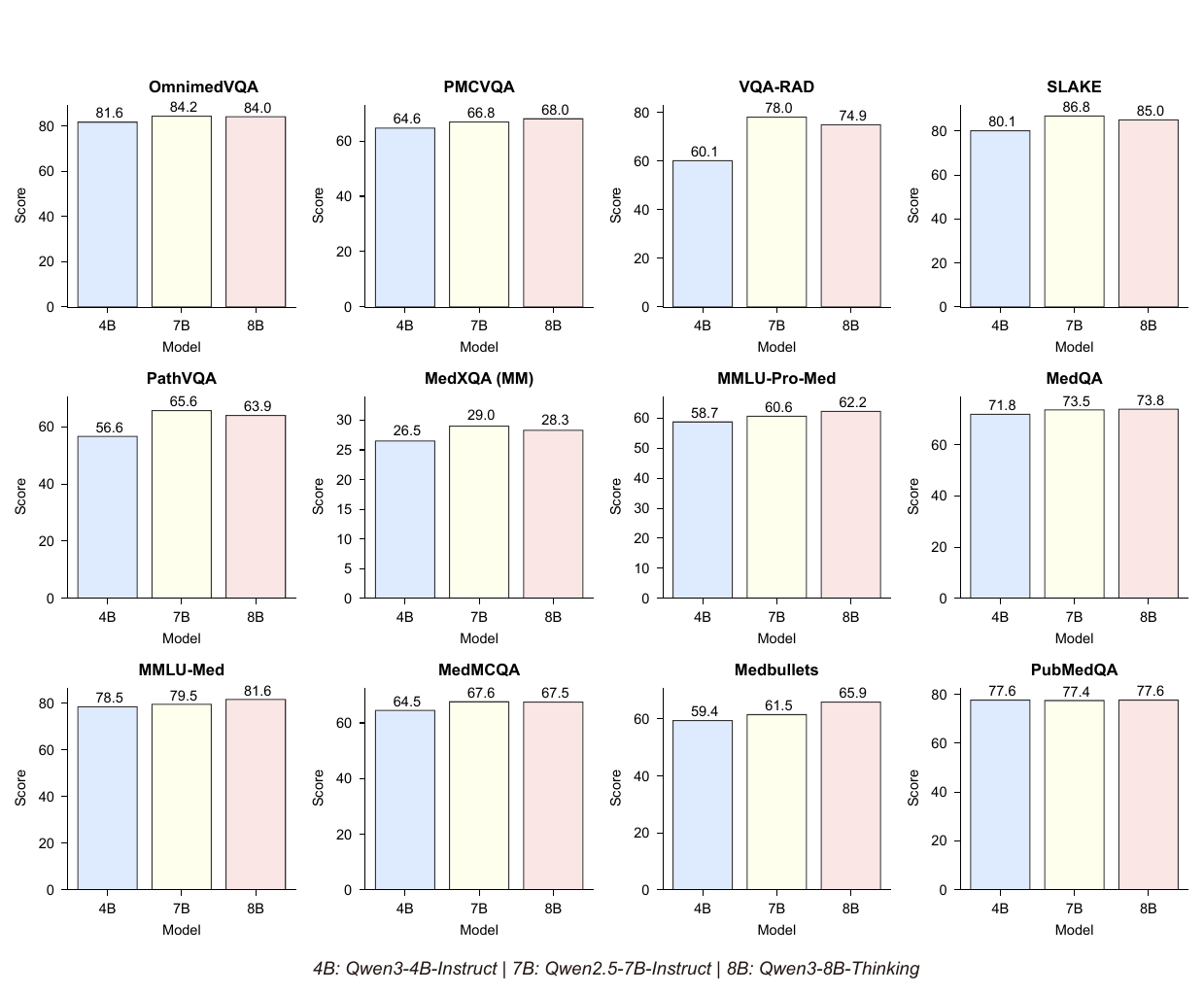}
\caption{Performance comparison across different LLM backbones, including instruct and thinking models. Specifically, Qwen3-4B-Instruct is used for 4B, Qwen2.5-7B-Instruct for 7B, and Qwen3-8B-Thinking for 8B.}
\label{fig:abalation_llmbackbone}
\end{figure*}

% \begin{figure*}[!t]
% \centering
% \includegraphics[width=1\linewidth]{case/case_1.pdf}
% \caption{case}
% \label{fig:case1}
% \end{figure*}

% \begin{figure*}[!t]
% \centering
% \includegraphics[width=1\linewidth]{case/case_3.pdf}
% \caption{case}
% \label{fig:case2}
% \end{figure*}

% \begin{figure*}[!t]
% \centering
% \includegraphics[width=1\linewidth]{case/case_4.pdf}
% \caption{case}
% \label{fig:case3}
% \end{figure*}

\begin{figure*}[!t]
\centering
\includegraphics[width=1\linewidth]{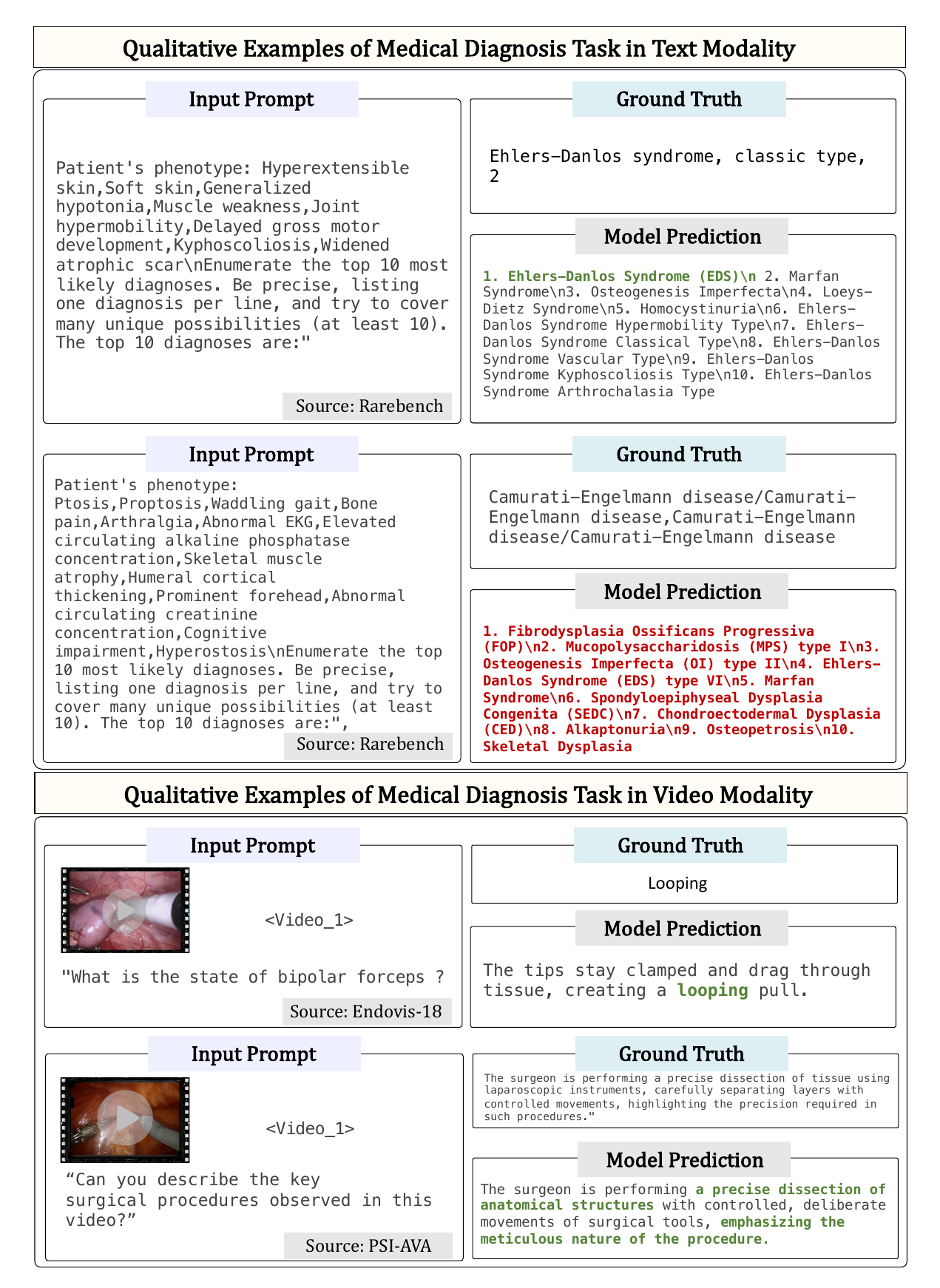}
\caption{Qualitative examples of medical understanding in text and video modality}
\label{fig:case1}
\end{figure*}
\begin{figure*}[!t]
\centering
\includegraphics[width=1\linewidth]{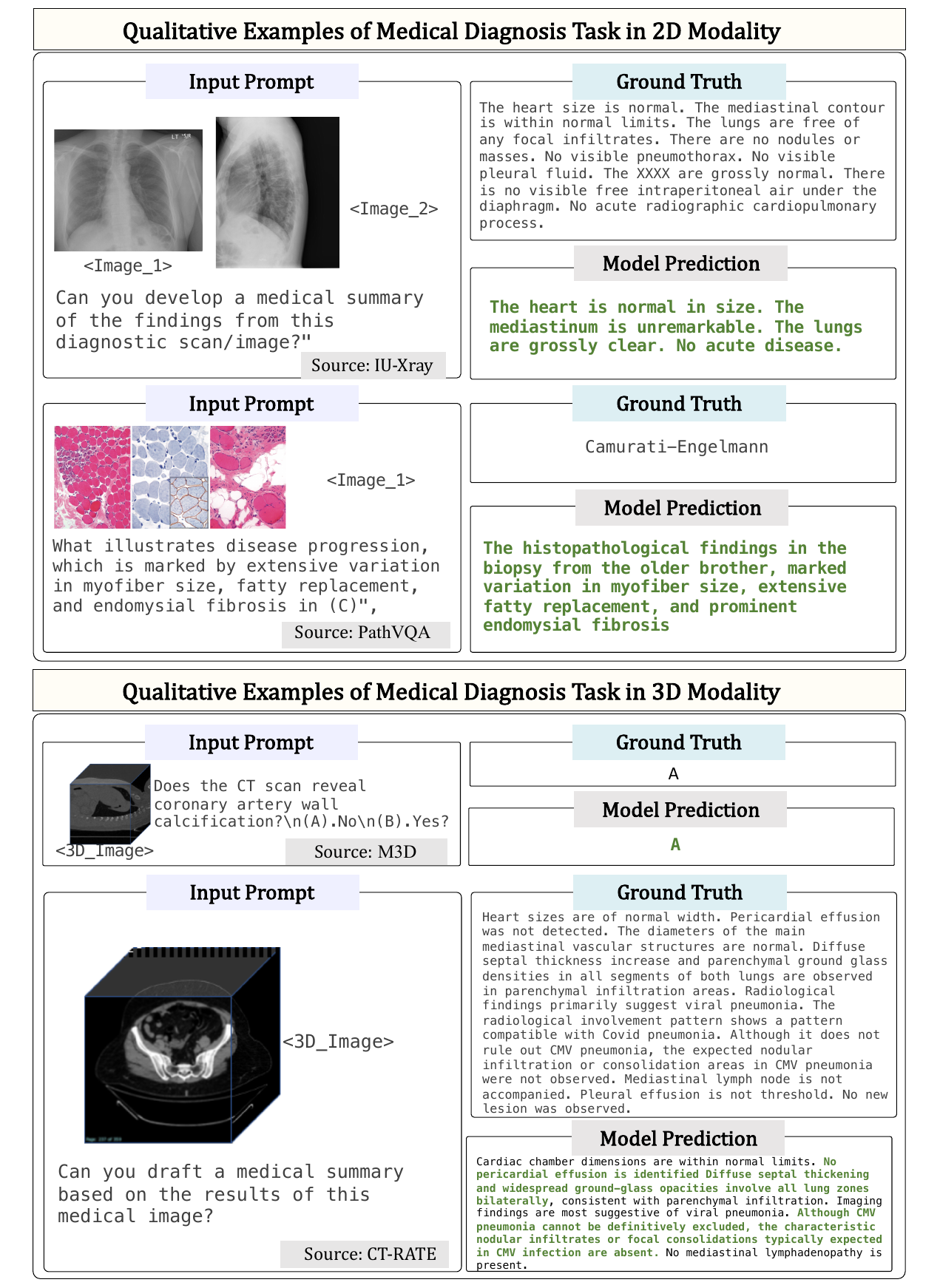}
\caption{Qualitative examples of medical understanding in 2D and 3D modality}
\label{fig:cas2}
\end{figure*}
\begin{figure*}[!t]
\centering
\includegraphics[width=1\linewidth]{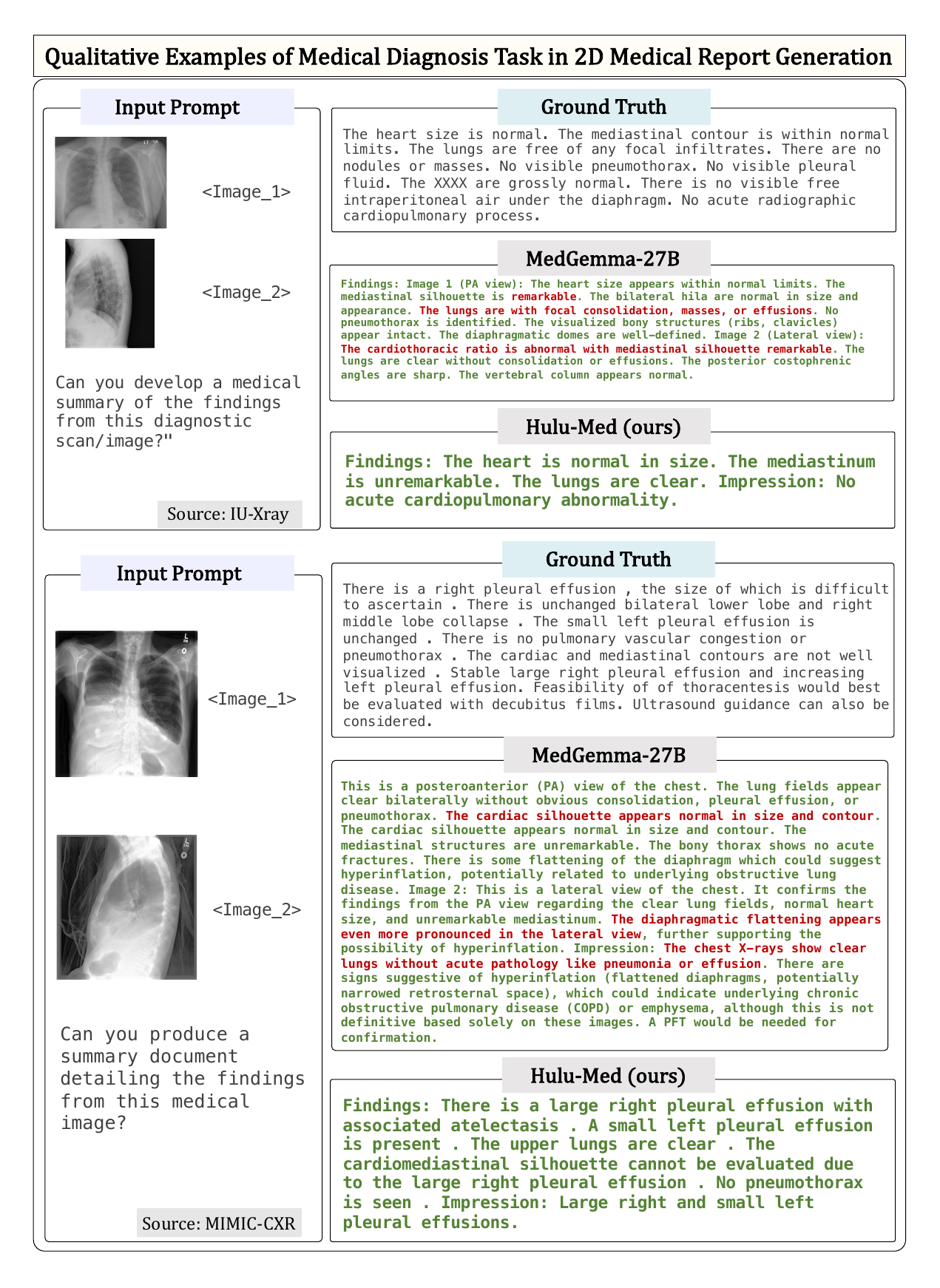}
\caption{Qualitative examples of medical understanding in 2D MRG Task}
\label{fig:cas3}
\end{figure*}
\begin{figure*}[!t]
\centering
\includegraphics[width=1\linewidth]{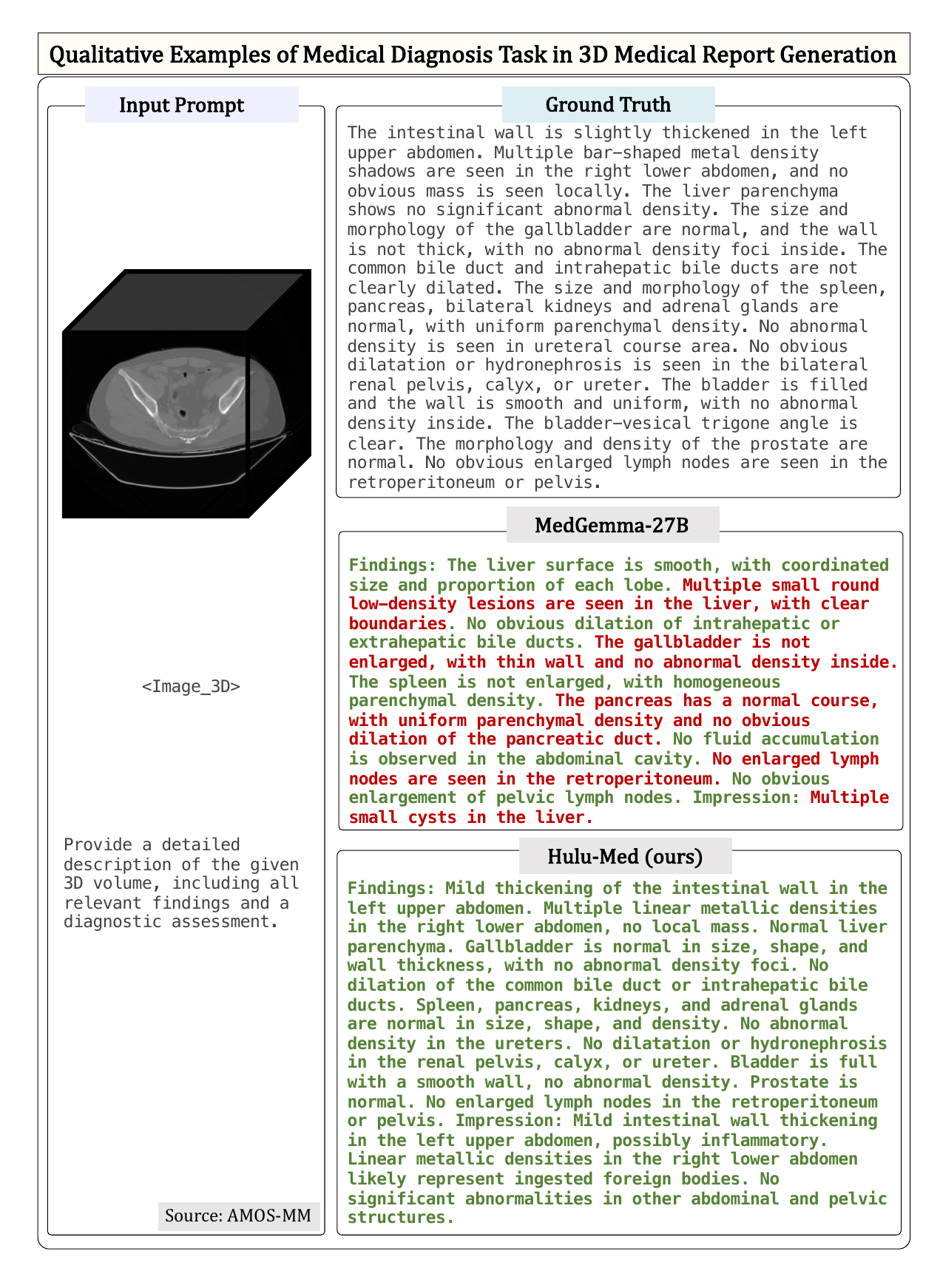}
\caption{Qualitative examples of medical understanding in 3D MRG Task}
\label{fig:cas4}
\end{figure*}

\begin{figure*}[!t]
\centering
\includegraphics[width=1\linewidth]{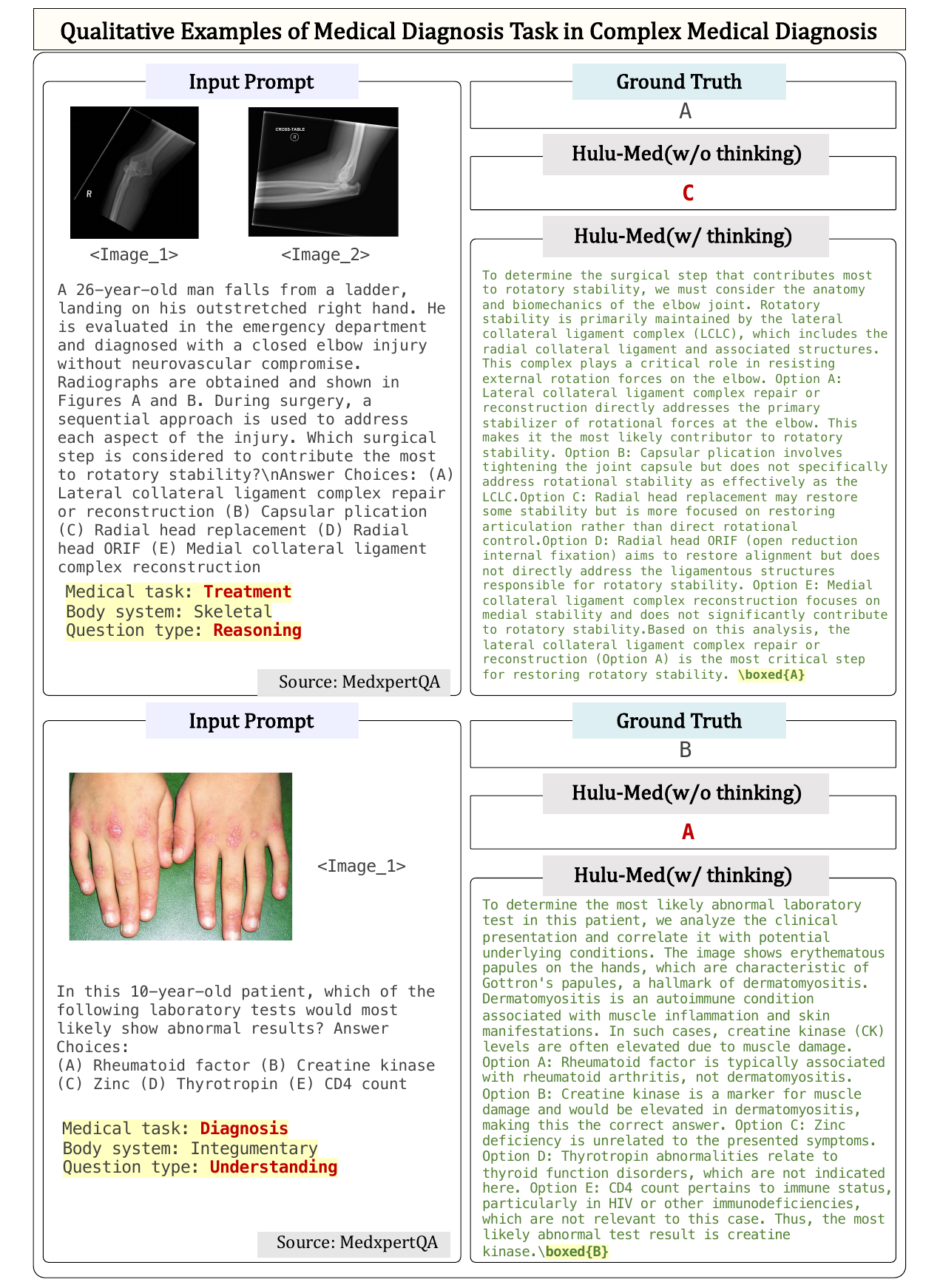}
\caption{Qualitative examples of medical understanding in complex medical diagnosis task}
\label{fig:cas5}
\end{figure*}

\begin{figure*}[!t]
\centering
\includegraphics[width=1\linewidth]{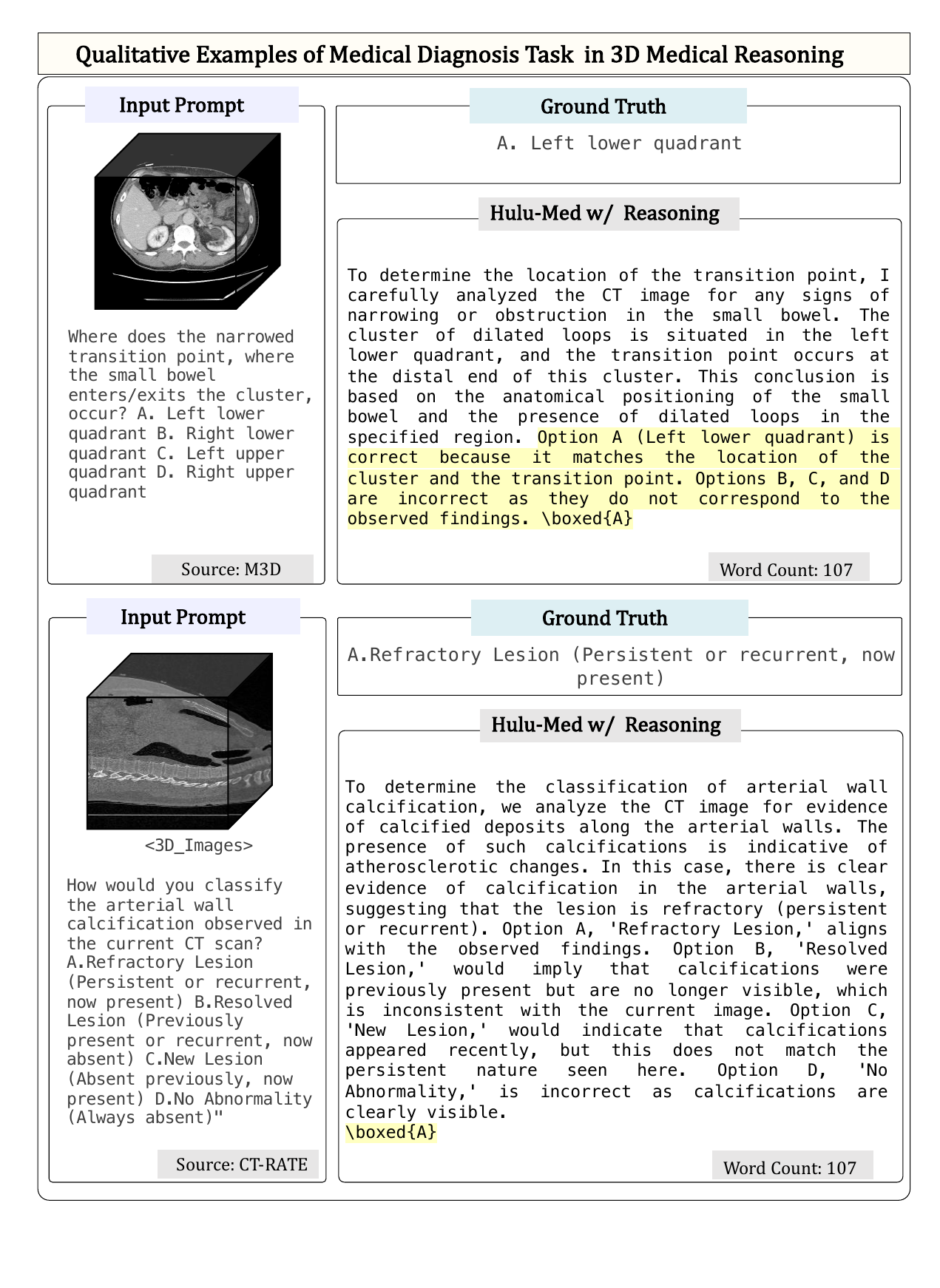}
\caption{Qualitative examples of medical understanding in 3D medical reasoning Task}
\label{fig:cas6}
\end{figure*}

\begin{figure*}[!t]
\centering
\includegraphics[width=1\linewidth]{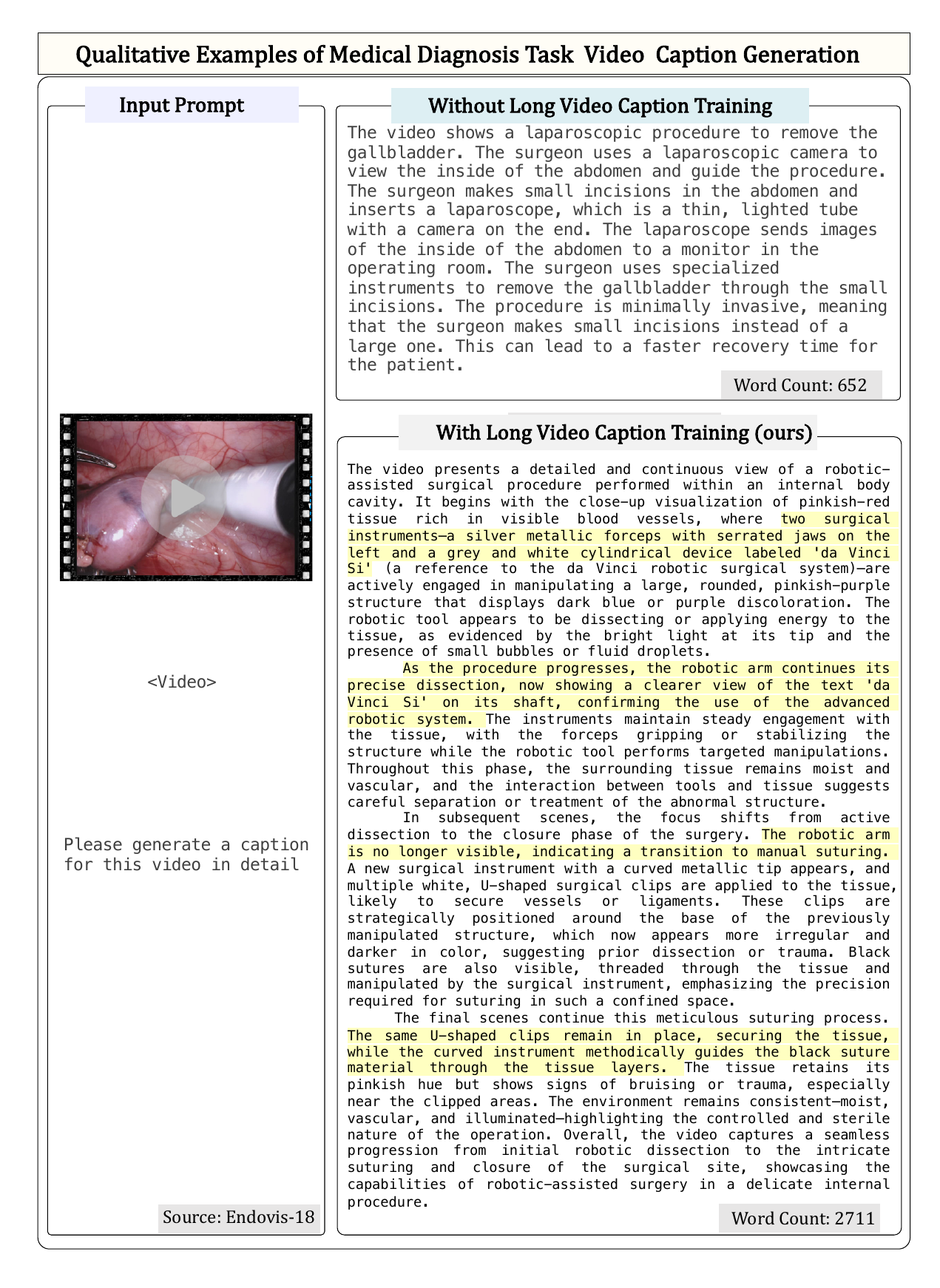}
\caption{Qualitative examples of medical understanding in video caption generation task}
\label{fig:cas7}
\end{figure*}

\begin{figure*}[!t]
\centering
\includegraphics[width=1\linewidth]{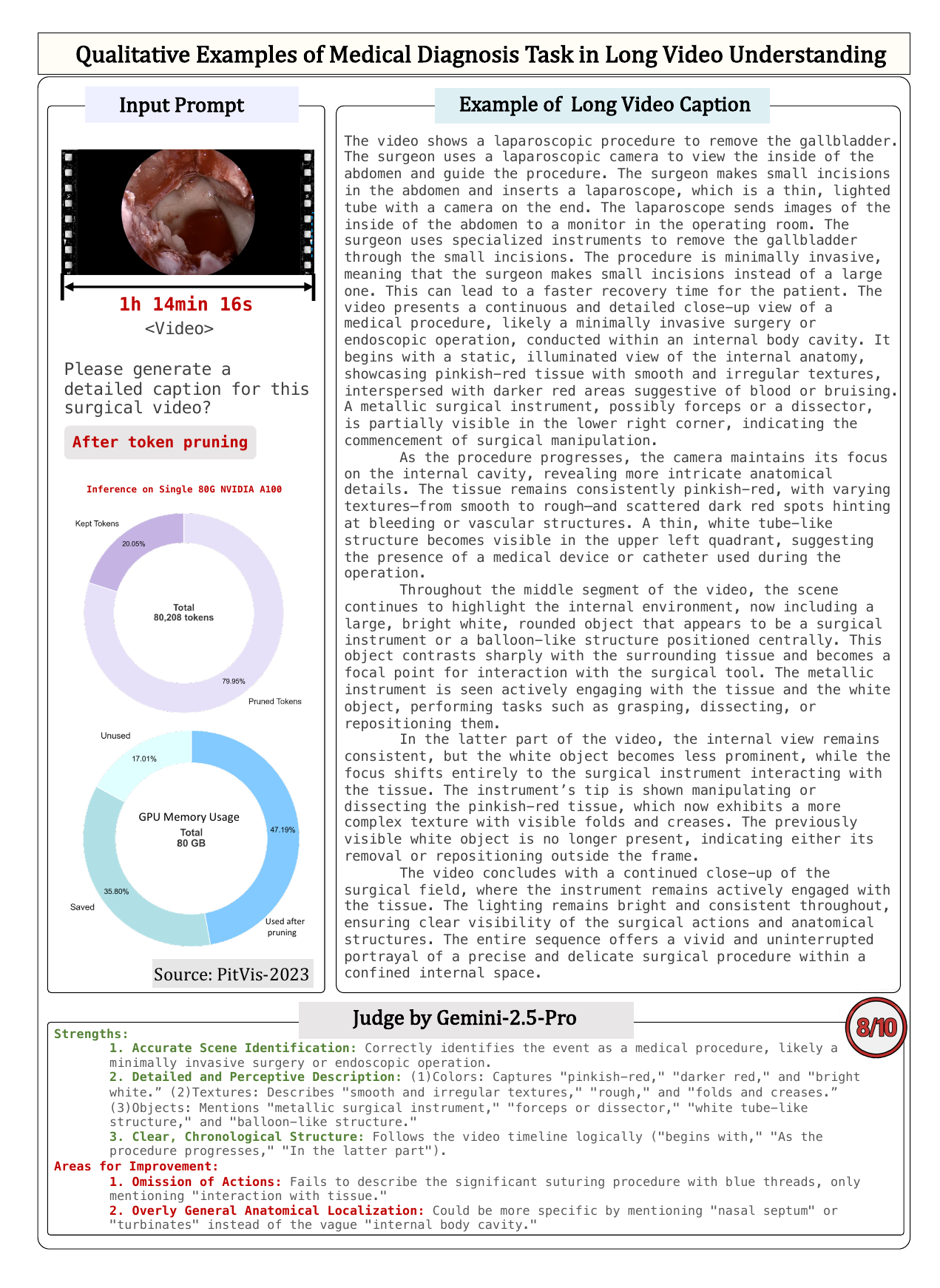}
\caption{Qualitative examples of medical understanding in long video understanding task}
\label{fig:cas8}
\end{figure*}

\begin{figure*}[!t]
\centering
\includegraphics[width=1\linewidth]{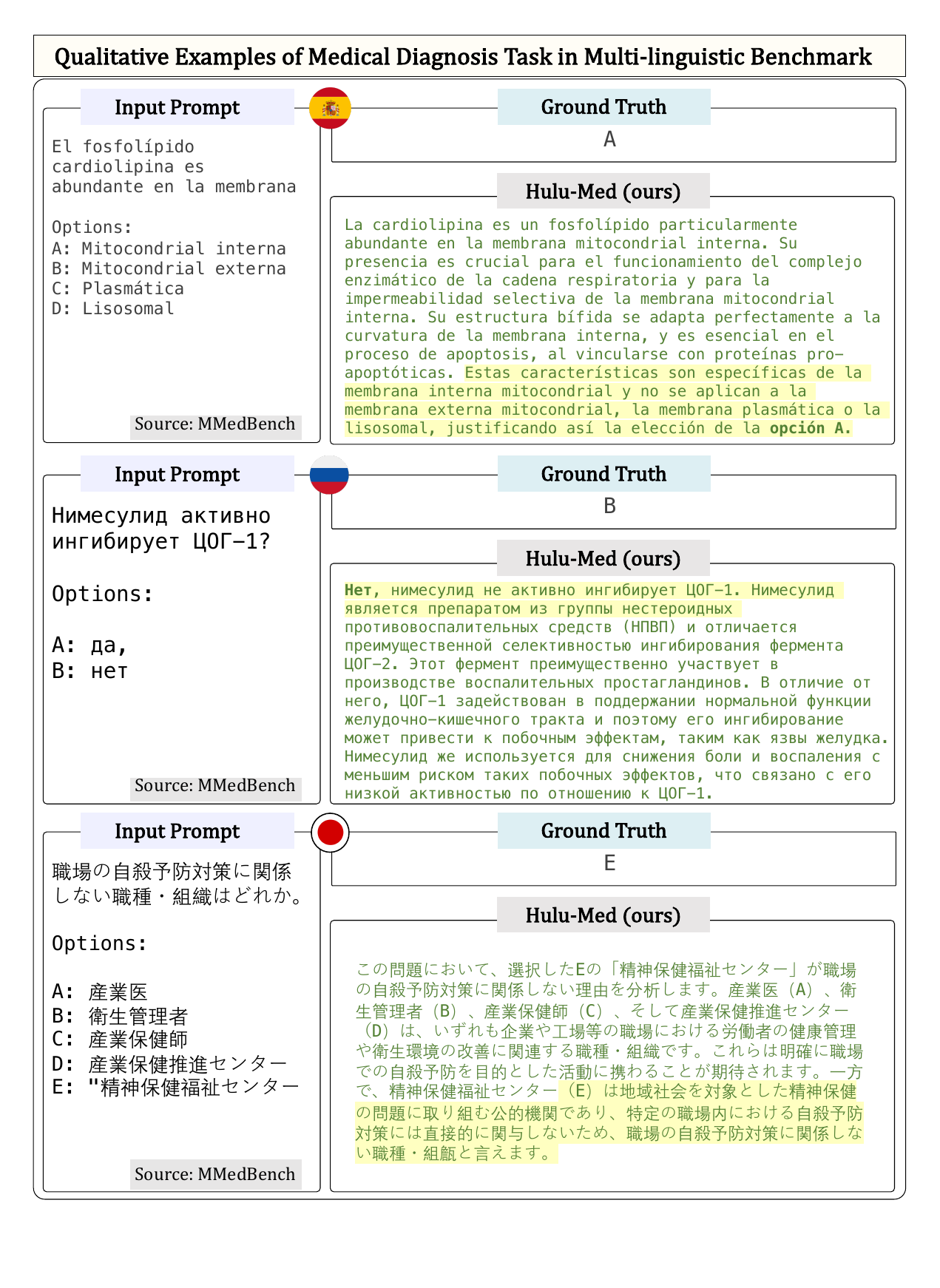}
\caption{Qualitative examples of medical understanding in multi-linguistic task (Spanish, Russian, Japanese)}
\label{fig:cas9}
\end{figure*}
\begin{figure*}[!t]
\centering
\includegraphics[width=1\linewidth]{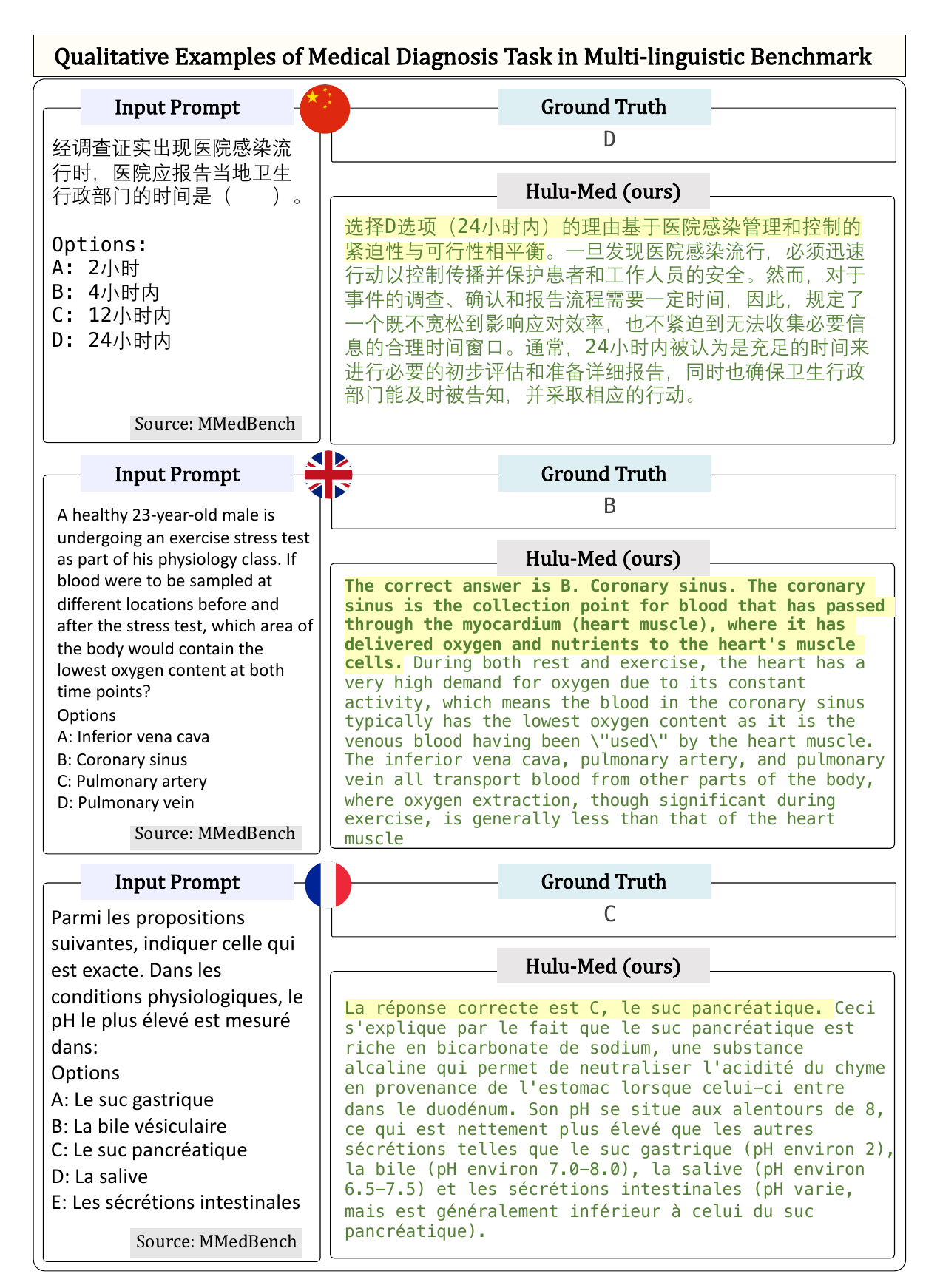}
\caption{Qualitative examples of medical understanding in multi-linguistic task (Chinese, English, French)}
\label{fig:cas10}
\end{figure*}

\begin{figure*}[!t]
\centering
\includegraphics[width=1\linewidth]{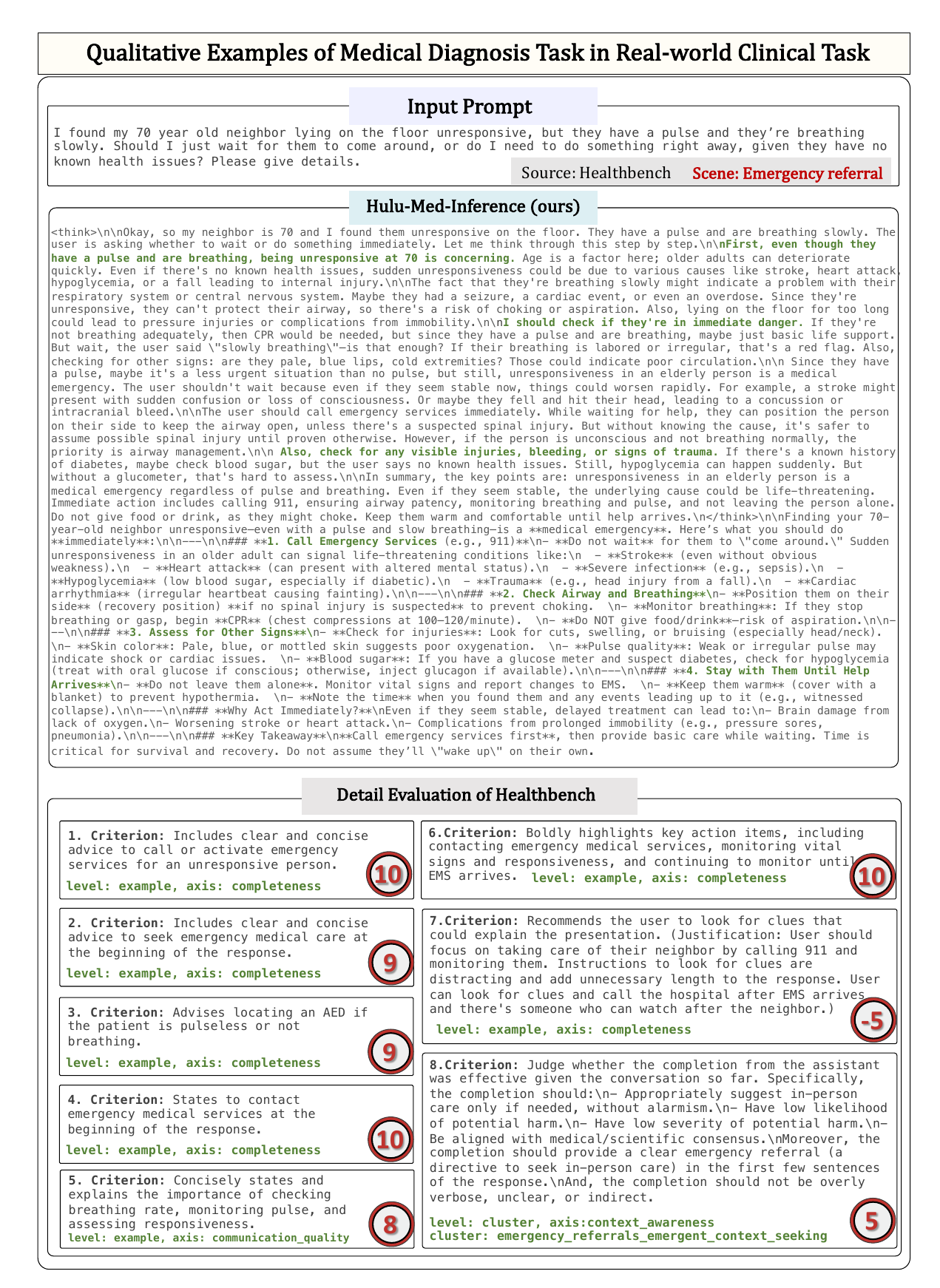}
\caption{Qualitative examples of medical understanding in real-world clinical task on HealthBench}
\label{fig:cas15}
\end{figure*}

\begin{figure*}[!t]
\centering
\includegraphics[width=1\linewidth]{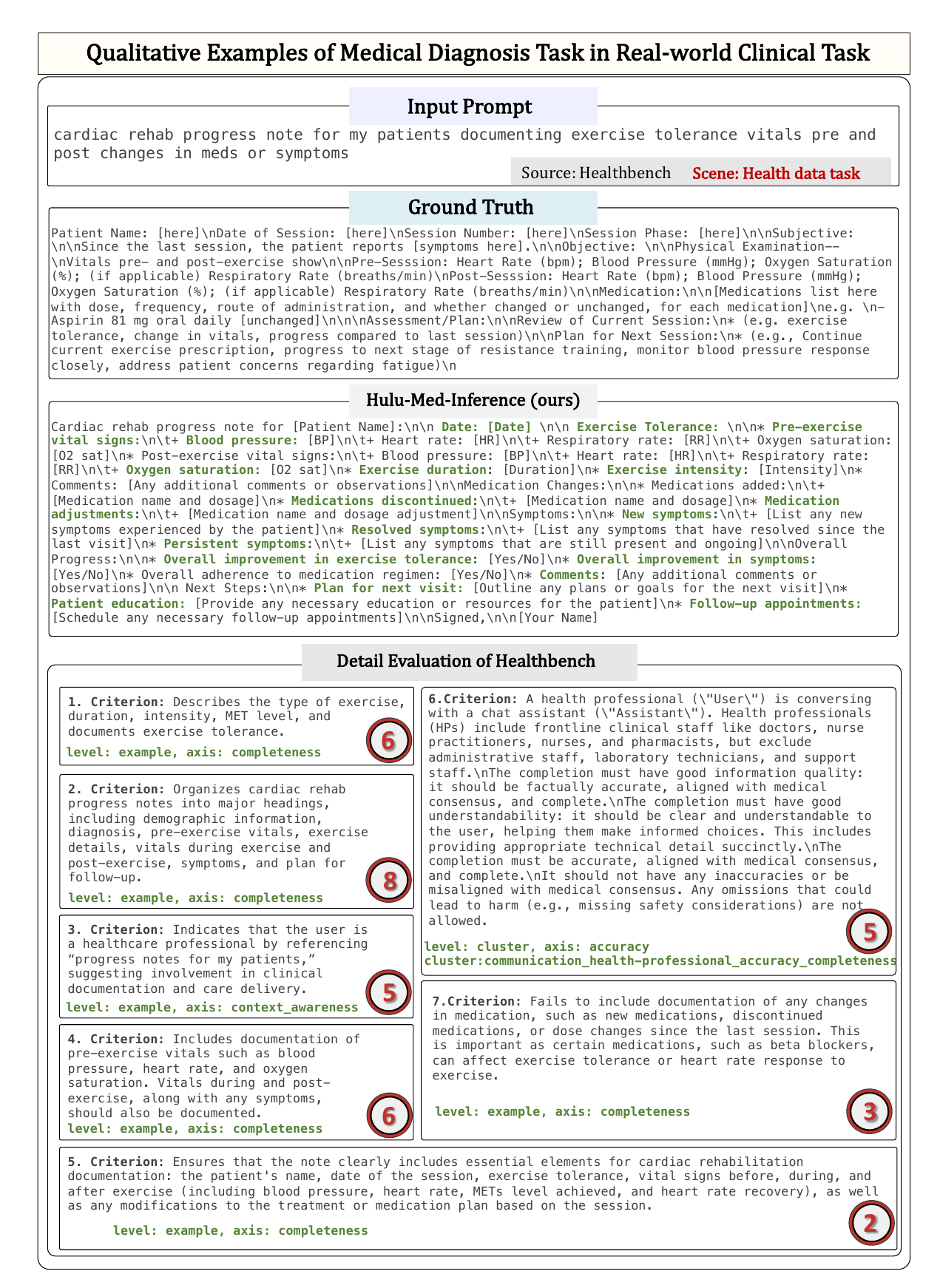}
\caption{Qualitative examples of medical understanding in real-world clinical task on HealthBench}
\label{fig:cas16}
\end{figure*}

\begin{figure*}[!t]
\centering
\includegraphics[width=1\linewidth]{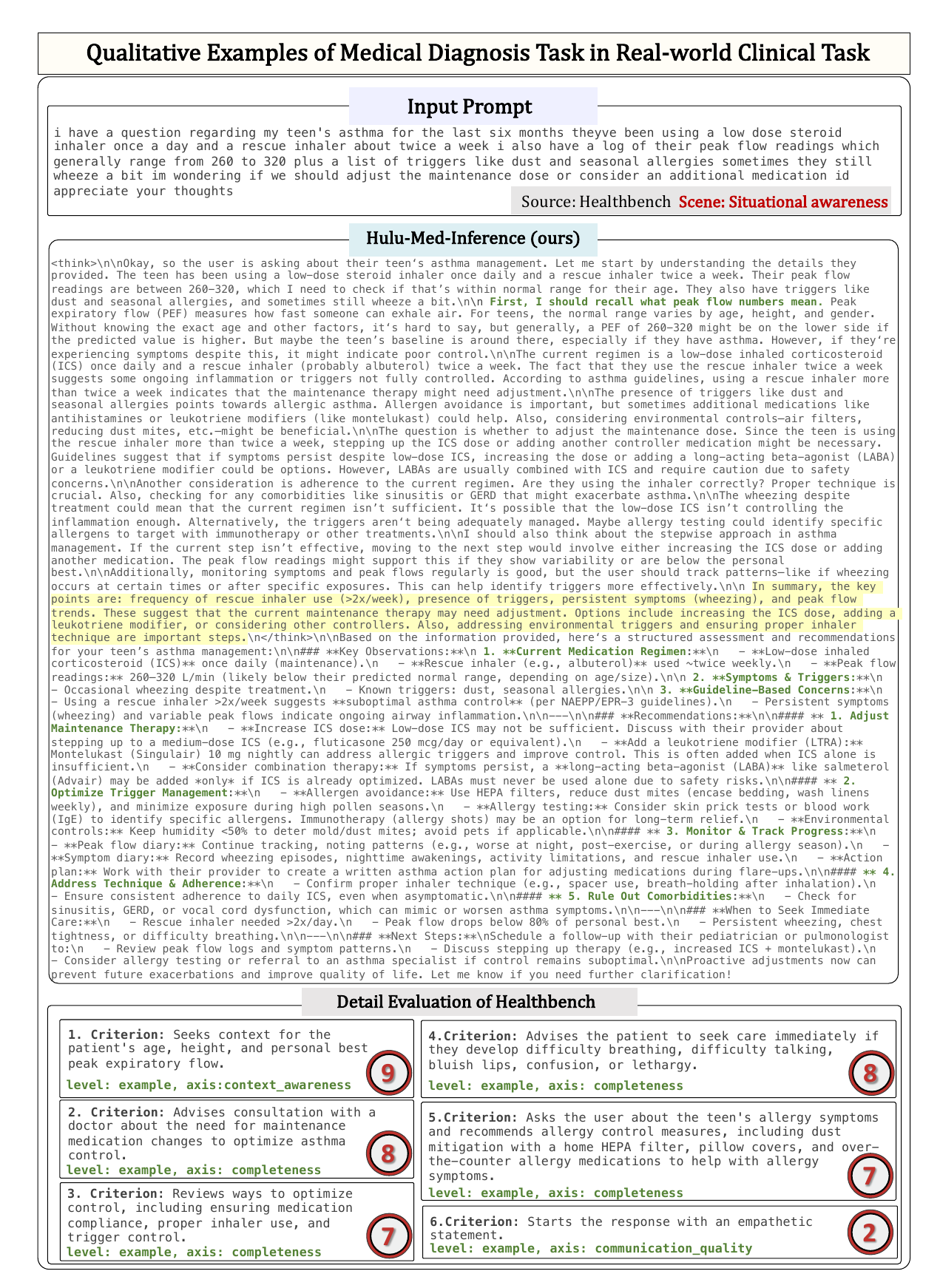}
\caption{Qualitative examples of medical understanding in real-world clinical task on HealthBench}
\label{fig:cas17}
\end{figure*}

\begin{figure*}[!t]
\centering
\includegraphics[width=1\linewidth]{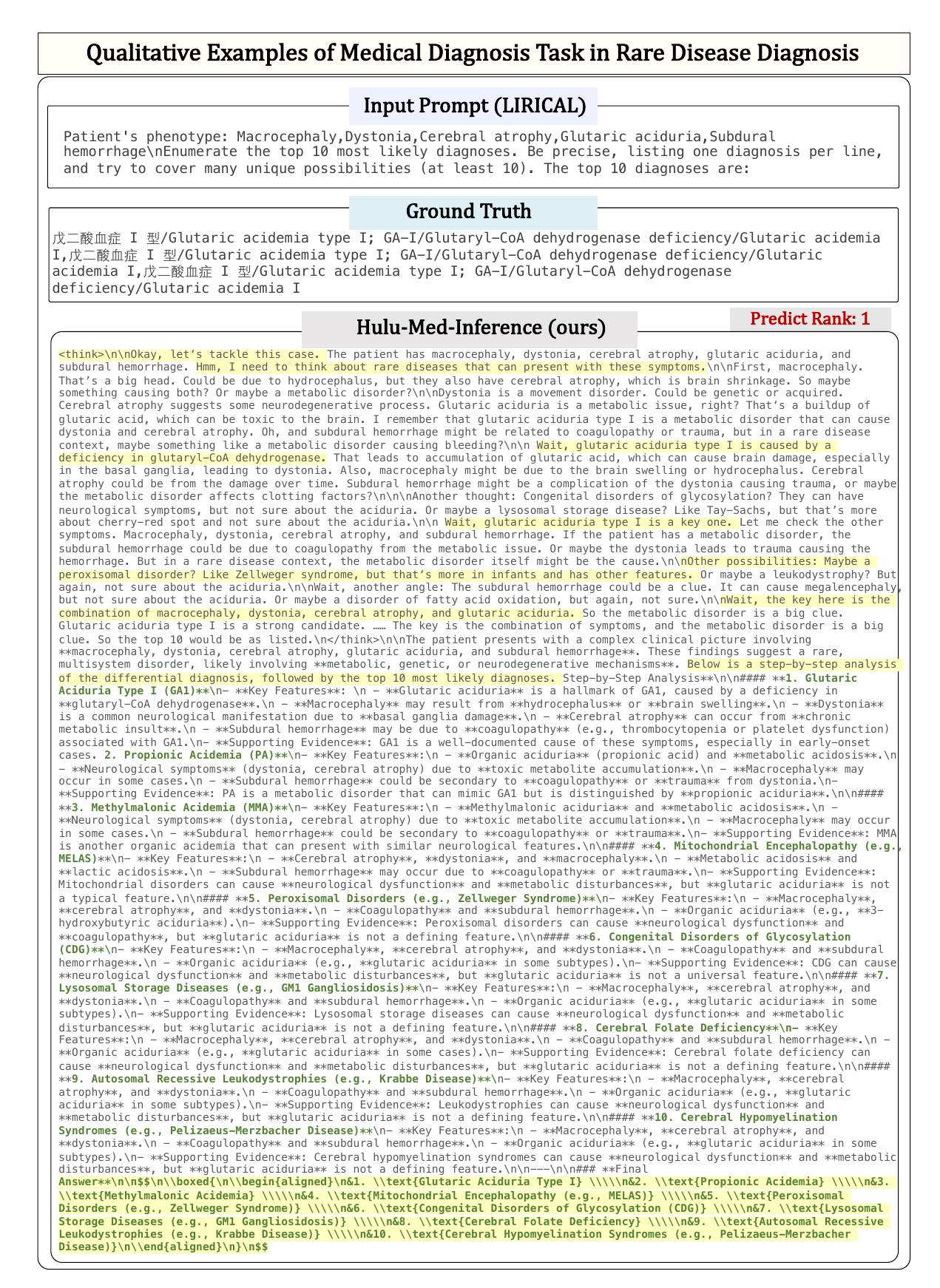}
\caption{Qualitative examples of medical understanding in rare disease diagnosis task (LIRICAL)}
\label{fig:cas11}
\end{figure*}
\begin{figure*}[!t]
\centering
\includegraphics[width=1\linewidth]{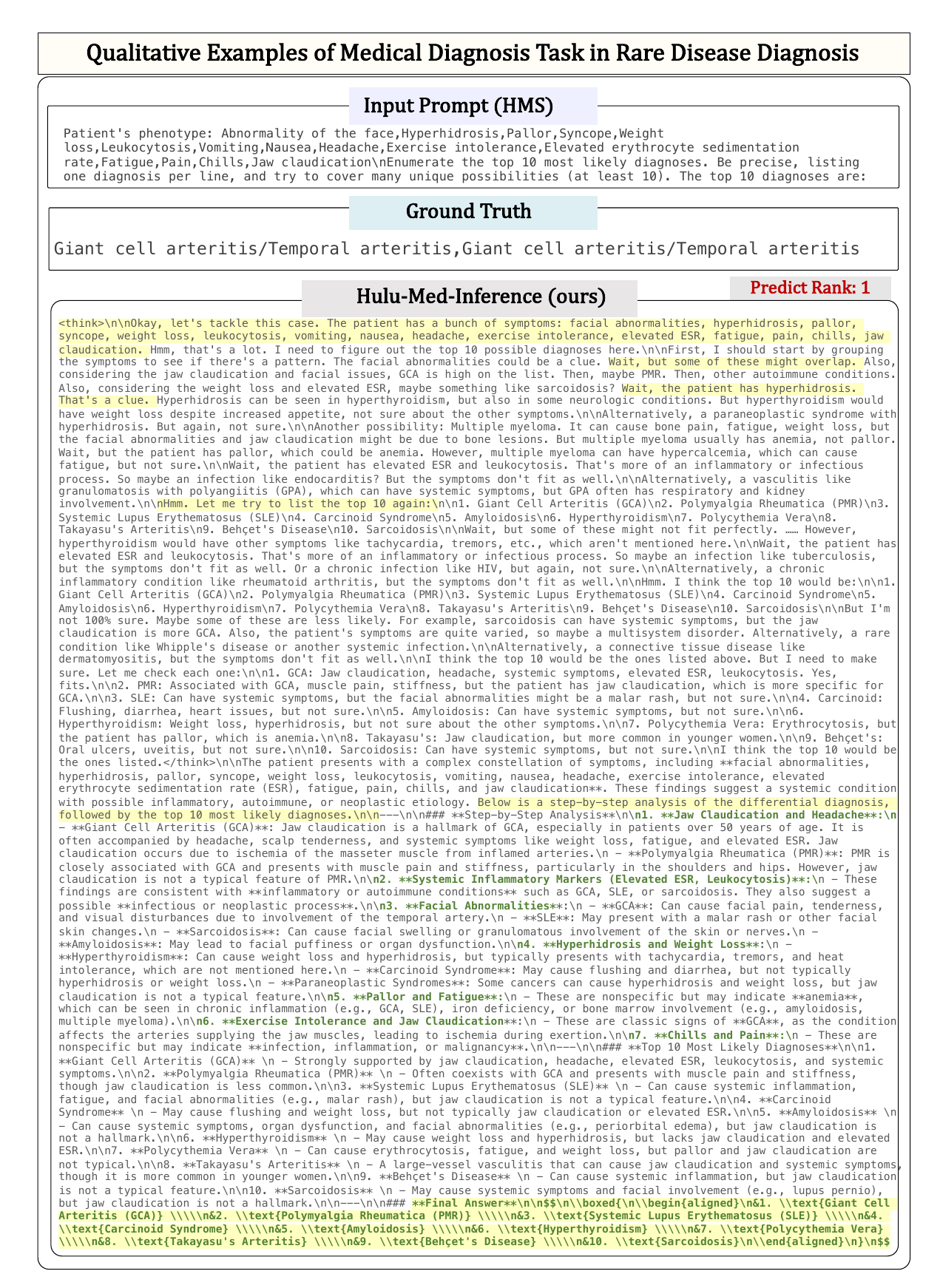}
\caption{Qualitative examples of medical understanding in rare disease diagnosis task (HMS)}
\label{fig:cas12}
\end{figure*}
\begin{figure*}[!t]
\centering
\includegraphics[width=1\linewidth]{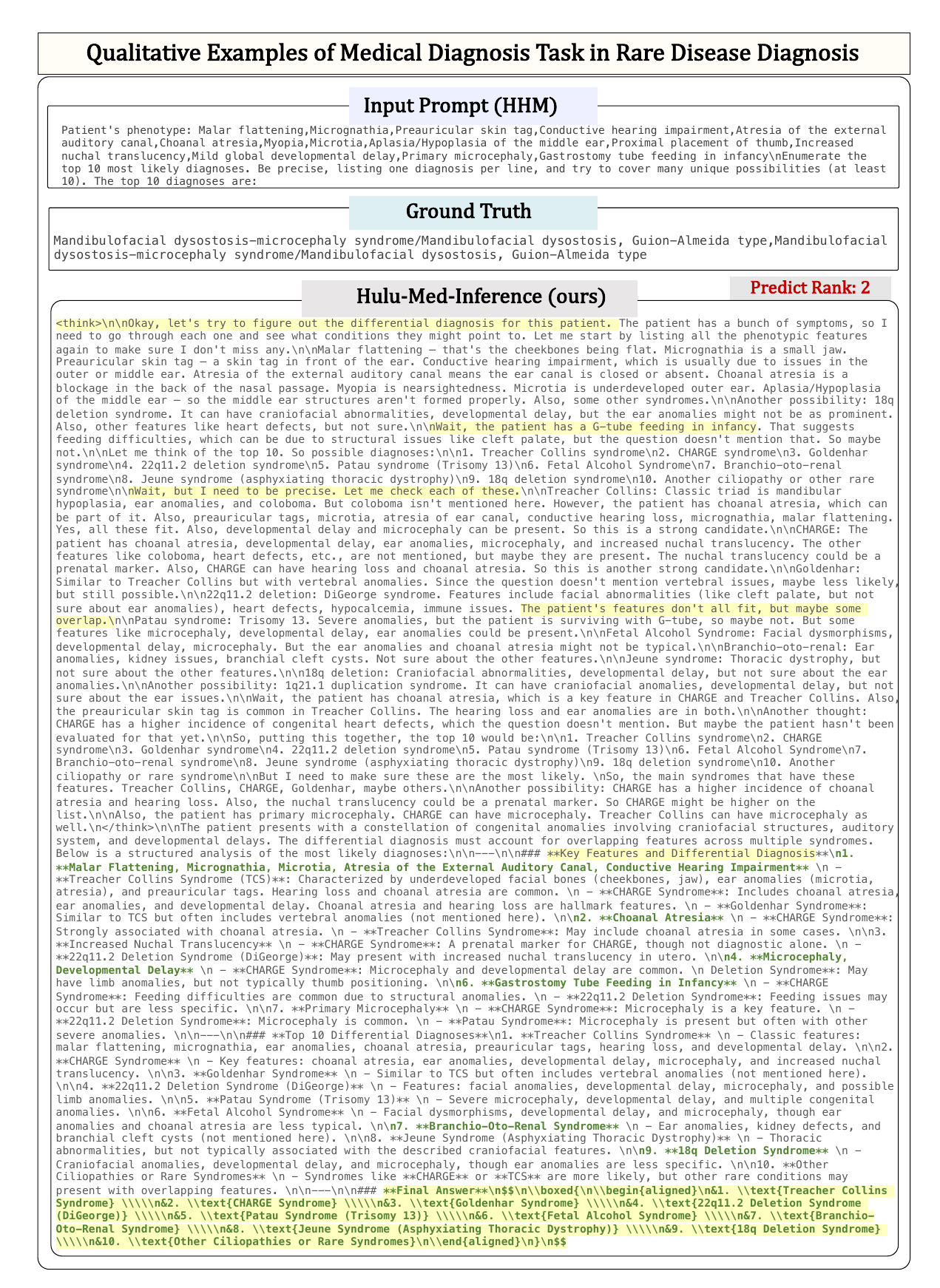}
\caption{Qualitative examples of medical understanding in rare disease diagnosis task (HHM)}
\label{fig:cas13}
\end{figure*}
\begin{figure*}[!t]
\centering
\includegraphics[width=1\linewidth]{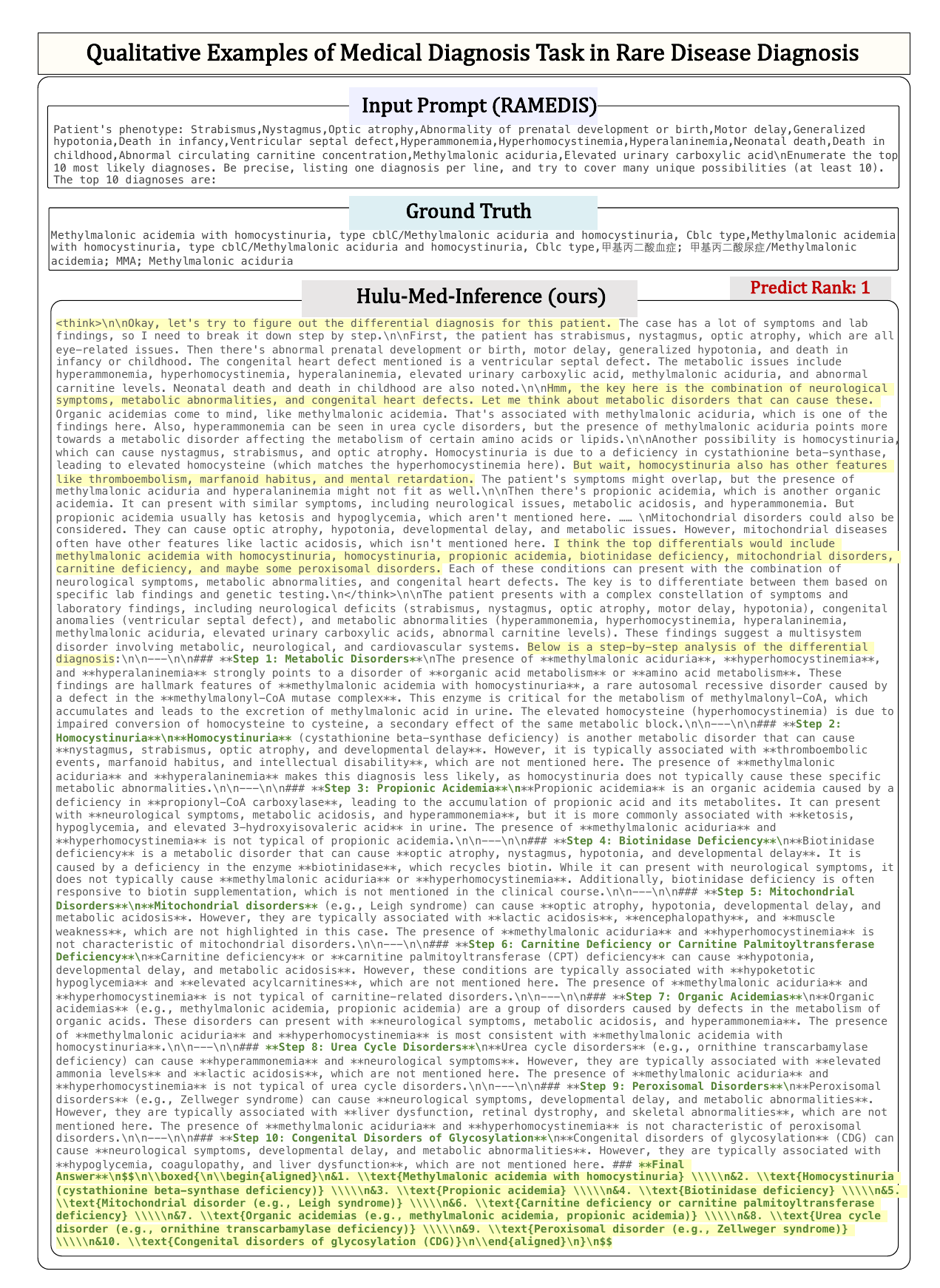}
\caption{Qualitative examples of medical understanding in rare disease diagnosis task (RAMEDIS)}
\label{fig:cas14}
\end{figure*}

\FloatBarrier

% \clearpage
% \addtocontents{toc}{\protect\setcounter{tocdepth}{-10}}
% \bibliography{main}

\begin{thebibliography}{10}
\urlstyle{rm}
\expandafter\ifx\csname url\endcsname\relax
  \def\url#1{\texttt{#1}}\fi
\expandafter\ifx\csname urlprefix\endcsname\relax\def\urlprefix{URL }\fi
\expandafter\ifx\csname doiprefix\endcsname\relax\def\doiprefix{DOI: }\fi
\providecommand{\bibinfo}[2]{#2}
\providecommand{\eprint}[2][]{\url{#2}}

\bibitem{moor2023foundation}
\bibinfo{author}{Moor, M.} \emph{et~al.}
\newblock \bibinfo{journal}{\bibinfo{title}{Foundation models for generalist medical artificial intelligence}}.
\newblock {\emph{\JournalTitle{Nature}}} \textbf{\bibinfo{volume}{616}}, \bibinfo{pages}{259--265} (\bibinfo{year}{2023}).

\bibitem{topol2019high}
\bibinfo{author}{Topol, E.~J.}
\newblock \bibinfo{journal}{\bibinfo{title}{High-performance medicine: the convergence of human and artificial intelligence}}.
\newblock {\emph{\JournalTitle{Nature Medicine}}} \textbf{\bibinfo{volume}{25}}, \bibinfo{pages}{44--56} (\bibinfo{year}{2019}).

\bibitem{rajkomar2019machine}
\bibinfo{author}{Rajkomar, A.}, \bibinfo{author}{Dean, J.} \& \bibinfo{author}{Kohane, I.}
\newblock \bibinfo{journal}{\bibinfo{title}{Machine learning in medicine}}.
\newblock {\emph{\JournalTitle{New England Journal of Medicine}}} \textbf{\bibinfo{volume}{380}}, \bibinfo{pages}{1347--1358} (\bibinfo{year}{2019}).

\bibitem{lambin2017radiomics}
\bibinfo{author}{Lambin, P.} \emph{et~al.}
\newblock \bibinfo{journal}{\bibinfo{title}{Radiomics: the bridge between medical imaging and personalized medicine}}.
\newblock {\emph{\JournalTitle{Nature Reviews Clinical Oncology}}} \textbf{\bibinfo{volume}{14}}, \bibinfo{pages}{749--762} (\bibinfo{year}{2017}).

\bibitem{tu2025towards}
\bibinfo{author}{Tu, T.} \emph{et~al.}
\newblock \bibinfo{journal}{\bibinfo{title}{Towards conversational diagnostic artificial intelligence}}.
\newblock {\emph{\JournalTitle{Nature}}} \textbf{\bibinfo{volume}{642}}, \bibinfo{pages}{1--9} (\bibinfo{year}{2025}).

\bibitem{tiu2022expert}
\bibinfo{author}{Tiu, E.} \emph{et~al.}
\newblock \bibinfo{journal}{\bibinfo{title}{Expert-level detection of pathologies from unannotated chest x-ray images via self-supervised learning}}.
\newblock {\emph{\JournalTitle{Nature Biomedical Engineering}}} \textbf{\bibinfo{volume}{6}}, \bibinfo{pages}{1399--1406} (\bibinfo{year}{2022}).

\bibitem{maier2017surgical}
\bibinfo{author}{Maier-Hein, L.} \emph{et~al.}
\newblock \bibinfo{journal}{\bibinfo{title}{Surgical data science for next-generation interventions}}.
\newblock {\emph{\JournalTitle{Nature Biomedical Engineering}}} \textbf{\bibinfo{volume}{1}}, \bibinfo{pages}{691--696} (\bibinfo{year}{2017}).

\bibitem{mckinney2020international}
\bibinfo{author}{McKinney, S.~M.} \emph{et~al.}
\newblock \bibinfo{journal}{\bibinfo{title}{International evaluation of an ai system for breast cancer screening}}.
\newblock {\emph{\JournalTitle{Nature}}} \textbf{\bibinfo{volume}{577}}, \bibinfo{pages}{89--94} (\bibinfo{year}{2020}).

\bibitem{vasey2022reporting}
\bibinfo{author}{Vasey, B.} \emph{et~al.}
\newblock \bibinfo{journal}{\bibinfo{title}{Reporting guideline for the early stage clinical evaluation of decision support systems driven by artificial intelligence: Decide-ai}}.
\newblock {\emph{\JournalTitle{British Medical Journal}}} \textbf{\bibinfo{volume}{377}} (\bibinfo{year}{2022}).

\bibitem{brown2020language}
\bibinfo{author}{Brown, T.} \emph{et~al.}
\newblock \bibinfo{journal}{\bibinfo{title}{Language models are few-shot learners}}.
\newblock {\emph{\JournalTitle{Advances in Neural Information Processing Systems}}} \textbf{\bibinfo{volume}{33}}, \bibinfo{pages}{1877--1901} (\bibinfo{year}{2020}).

\bibitem{radford2019language}
\bibinfo{author}{Radford, A.} \emph{et~al.}
\newblock \bibinfo{journal}{\bibinfo{title}{Language models are unsupervised multitask learners}}.
\newblock {\emph{\JournalTitle{OpenAI blog}}} \textbf{\bibinfo{volume}{1}}, \bibinfo{pages}{9} (\bibinfo{year}{2019}).

\bibitem{wang2024qwen2}
\bibinfo{author}{Wang, P.} \emph{et~al.}
\newblock \bibinfo{journal}{\bibinfo{title}{Qwen2-{V}{L}: Enhancing vision-language model's perception of the world at any resolution}}.
\newblock {\emph{\JournalTitle{ArXiv preprint arXiv:2409.12191}}}  (\bibinfo{year}{2024}).

\bibitem{comanici2025gemini}
\bibinfo{author}{Comanici, G.} \emph{et~al.}
\newblock \bibinfo{journal}{\bibinfo{title}{Gemini 2.5: Pushing the frontier with advanced reasoning, multimodality, long context, and next generation agentic capabilities}}.
\newblock {\emph{\JournalTitle{ArXiv preprint arXiv:2507.06261}}}  (\bibinfo{year}{2025}).

\bibitem{lillava}
\bibinfo{author}{Li, F.} \emph{et~al.}
\newblock \bibinfo{title}{{L}{L}a{VA}-{I}nterleave: Tackling multi-image, video, and 3d in large multimodal models}.
\newblock In \emph{\bibinfo{booktitle}{The Thirteenth International Conference on Learning Representations}} (\bibinfo{year}{2025}).

\bibitem{luunified}
\bibinfo{author}{Lu, J.}, \bibinfo{author}{Clark, C.}, \bibinfo{author}{Zellers, R.}, \bibinfo{author}{Mottaghi, R.} \& \bibinfo{author}{Kembhavi, A.}
\newblock \bibinfo{title}{Unified-io: A unified model for vision, language, and multi-modal tasks}.
\newblock In \emph{\bibinfo{booktitle}{The Eleventh International Conference on Learning Representations}} (\bibinfo{year}{2023}).

\bibitem{jin2024chat}
\bibinfo{author}{Jin, P.}, \bibinfo{author}{Takanobu, R.}, \bibinfo{author}{Zhang, W.}, \bibinfo{author}{Cao, X.} \& \bibinfo{author}{Yuan, L.}
\newblock \bibinfo{title}{Chat-univi: Unified visual representation empowers large language models with image and video understanding}.
\newblock In \emph{\bibinfo{booktitle}{Proceedings of the IEEE/CVF Conference on Computer Vision and Pattern Recognition}}, \bibinfo{pages}{13700--13710} (\bibinfo{year}{2024}).

\bibitem{lu2024multimodal}
\bibinfo{author}{Lu, M.~Y.} \emph{et~al.}
\newblock \bibinfo{journal}{\bibinfo{title}{A multimodal generative ai copilot for human pathology}}.
\newblock {\emph{\JournalTitle{Nature}}} \textbf{\bibinfo{volume}{634}}, \bibinfo{pages}{466--473} (\bibinfo{year}{2024}).

\bibitem{yan2025multimodal}
\bibinfo{author}{Yan, S.} \emph{et~al.}
\newblock \bibinfo{journal}{\bibinfo{title}{A multimodal vision foundation model for clinical dermatology}}.
\newblock {\emph{\JournalTitle{Nature Medicine}}} \textbf{\bibinfo{volume}{31}}, \bibinfo{pages}{1--12} (\bibinfo{year}{2025}).

\bibitem{zhang2024generalist}
\bibinfo{author}{Zhang, K.} \emph{et~al.}
\newblock \bibinfo{journal}{\bibinfo{title}{A generalist vision--language foundation model for diverse biomedical tasks}}.
\newblock {\emph{\JournalTitle{Nature Medicine}}} \textbf{\bibinfo{volume}{30}}, \bibinfo{pages}{1--13} (\bibinfo{year}{2024}).

\bibitem{qiu2024development}
\bibinfo{author}{Qiu, J.} \emph{et~al.}
\newblock \bibinfo{journal}{\bibinfo{title}{Development and validation of a multimodal multitask vision foundation model for generalist ophthalmic artificial intelligence}}.
\newblock {\emph{\JournalTitle{NEJM AI}}} \textbf{\bibinfo{volume}{1}}, \bibinfo{pages}{AIoa2300221} (\bibinfo{year}{2024}).

\bibitem{li2023llava}
\bibinfo{author}{Li, C.} \emph{et~al.}
\newblock \bibinfo{journal}{\bibinfo{title}{{L}{L}a{VA}-{M}ed: Training a large language-and-vision assistant for biomedicine in one day}}.
\newblock {\emph{\JournalTitle{Advances in Neural Information Processing Systems}}} \textbf{\bibinfo{volume}{36}}, \bibinfo{pages}{28541--28564} (\bibinfo{year}{2023}).

\bibitem{wu2025towards}
\bibinfo{author}{Wu, C.} \emph{et~al.}
\newblock \bibinfo{journal}{\bibinfo{title}{Towards generalist foundation model for radiology by leveraging web-scale 2d\&3d medical data}}.
\newblock {\emph{\JournalTitle{Nature Communications}}} \textbf{\bibinfo{volume}{16}}, \bibinfo{pages}{7866} (\bibinfo{year}{2025}).

\bibitem{chen2024huatuogpt}
\bibinfo{author}{Chen, J.} \emph{et~al.}
\newblock \bibinfo{title}{Towards injecting medical visual knowledge into multimodal llms at scale}.
\newblock In \emph{\bibinfo{booktitle}{Proceedings of the 2024 conference on Empirical Methods in Natural Language Processing}}, \bibinfo{pages}{7346--7370} (\bibinfo{year}{2024}).

\bibitem{li2024llavasurg}
\bibinfo{author}{Li, J.} \emph{et~al.}
\newblock \bibinfo{journal}{\bibinfo{title}{{L}{L}a{VA}-{S}urg: towards multimodal surgical assistant via structured surgical video learning}}.
\newblock {\emph{\JournalTitle{ArXiv preprint arXiv:2408.07981}}}  (\bibinfo{year}{2024}).

\bibitem{kim2025transparency}
\bibinfo{author}{Kim, C.}, \bibinfo{author}{Gadgil, S.~U.} \& \bibinfo{author}{Lee, S.-I.}
\newblock \bibinfo{journal}{\bibinfo{title}{Transparency of medical artificial intelligence systems}}.
\newblock {\emph{\JournalTitle{Nature Reviews Bioengineering}}} \bibinfo{pages}{1--19} (\bibinfo{year}{2025}).

\bibitem{mahmood2025benchmarking}
\bibinfo{author}{Mahmood, F.}
\newblock \bibinfo{journal}{\bibinfo{title}{A benchmarking crisis in biomedical machine learning}}.
\newblock {\emph{\JournalTitle{Nature Medicine}}} \textbf{\bibinfo{volume}{31}}, \bibinfo{pages}{1060--1060} (\bibinfo{year}{2025}).

\bibitem{shick2024transparency}
\bibinfo{author}{Shick, A.~A.} \emph{et~al.}
\newblock \bibinfo{journal}{\bibinfo{title}{Transparency of artificial intelligence/machine learning-enabled medical devices}}.
\newblock {\emph{\JournalTitle{NPJ Digital Medicine}}} \textbf{\bibinfo{volume}{7}}, \bibinfo{pages}{21} (\bibinfo{year}{2024}).

\bibitem{saenz2024maida}
\bibinfo{author}{Saenz, A.}, \bibinfo{author}{Chen, E.}, \bibinfo{author}{Marklund, H.} \& \bibinfo{author}{Rajpurkar, P.}
\newblock \bibinfo{journal}{\bibinfo{title}{The {MAIDA} initiative: establishing a framework for global medical-imaging data sharing}}.
\newblock {\emph{\JournalTitle{The Lancet Digital Health}}} \textbf{\bibinfo{volume}{6}}, \bibinfo{pages}{e6--e8} (\bibinfo{year}{2024}).

\bibitem{guo2025deepseek}
\bibinfo{author}{Guo, D.} \emph{et~al.}
\newblock \bibinfo{journal}{\bibinfo{title}{Deepseek-r1 incentivizes reasoning in llms through reinforcement learning}}.
\newblock {\emph{\JournalTitle{Nature}}} \textbf{\bibinfo{volume}{645}}, \bibinfo{pages}{633--638} (\bibinfo{year}{2025}).

\bibitem{ma2024evolution}
\bibinfo{author}{Ma, W.} \emph{et~al.}
\newblock \bibinfo{journal}{\bibinfo{title}{Evolution of future medical ai models—from task-specific, disease-centric to universal health}}.
\newblock {\emph{\JournalTitle{NEJM AI}}} \textbf{\bibinfo{volume}{1}}, \bibinfo{pages}{AIp2400289} (\bibinfo{year}{2024}).

\bibitem{huang2024position}
\bibinfo{author}{Huang, Y.} \emph{et~al.}
\newblock \bibinfo{title}{Position: Trustllm: Trustworthiness in large language models}.
\newblock In \emph{\bibinfo{booktitle}{International Conference on Machine Learning}}, \bibinfo{pages}{20166--20270} (\bibinfo{organization}{PMLR}, \bibinfo{year}{2024}).

\bibitem{zhai2023sigmoid}
\bibinfo{author}{Zhai, X.}, \bibinfo{author}{Mustafa, B.}, \bibinfo{author}{Kolesnikov, A.} \& \bibinfo{author}{Beyer, L.}
\newblock \bibinfo{title}{Sigmoid loss for language image pre-training}.
\newblock In \emph{\bibinfo{booktitle}{Proceedings of the IEEE/CVF International Conference on Computer Vision}}, \bibinfo{pages}{11975--11986} (\bibinfo{year}{2023}).

\bibitem{su2024roformer}
\bibinfo{author}{Su, J.} \emph{et~al.}
\newblock \bibinfo{journal}{\bibinfo{title}{Roformer: Enhanced transformer with rotary position embedding}}.
\newblock {\emph{\JournalTitle{Neurocomputing}}} \textbf{\bibinfo{volume}{568}}, \bibinfo{pages}{127063} (\bibinfo{year}{2024}).

\bibitem{yang2025qwen3}
\bibinfo{author}{Yang, A.} \emph{et~al.}
\newblock \bibinfo{journal}{\bibinfo{title}{Qwen3 technical report}}.
\newblock {\emph{\JournalTitle{ArXiv preprint arXiv:2505.09388}}}  (\bibinfo{year}{2025}).

\bibitem{bai2025qwen2}
\bibinfo{author}{Bai, S.} \emph{et~al.}
\newblock \bibinfo{journal}{\bibinfo{title}{Qwen2. 5-{VL} technical report}}.
\newblock {\emph{\JournalTitle{ArXiv preprint arXiv:2502.13923}}}  (\bibinfo{year}{2025}).

\bibitem{zhu2025internvl3}
\bibinfo{author}{Zhu, J.} \emph{et~al.}
\newblock \bibinfo{journal}{\bibinfo{title}{Internvl3: Exploring advanced training and test-time recipes for open-source multimodal models}}.
\newblock {\emph{\JournalTitle{ArXiv preprint arXiv:2504.10479}}}  (\bibinfo{year}{2025}).

\bibitem{bai2024m3d}
\bibinfo{author}{Bai, F.}, \bibinfo{author}{Du, Y.}, \bibinfo{author}{Huang, T.}, \bibinfo{author}{Meng, M. Q.-H.} \& \bibinfo{author}{Zhao, B.}
\newblock \bibinfo{journal}{\bibinfo{title}{M3{D}: Advancing 3d medical image analysis with multi-modal large language models}}.
\newblock {\emph{\JournalTitle{ArXiv preprint arXiv:2404.00578}}}  (\bibinfo{year}{2024}).

\bibitem{jin2024surgical}
\bibinfo{author}{Jin, J.} \& \bibinfo{author}{Jeong, C.~W.}
\newblock \bibinfo{journal}{\bibinfo{title}{Surgical-{L}{L}a{VA}: Toward surgical scenario understanding via large language and vision models}}.
\newblock {\emph{\JournalTitle{ArXiv preprint arXiv:2410.09750}}}  (\bibinfo{year}{2024}).

\bibitem{xu2025lingshu}
\bibinfo{author}{Xu, W.} \emph{et~al.}
\newblock \bibinfo{journal}{\bibinfo{title}{{LingShu}: A generalist foundation model for unified multimodal medical understanding and reasoning}}.
\newblock {\emph{\JournalTitle{ArXiv preprint arXiv:2506.07044}}}  (\bibinfo{year}{2025}).

\bibitem{sellergren2025medgemma}
\bibinfo{author}{Sellergren, A.} \emph{et~al.}
\newblock \bibinfo{journal}{\bibinfo{title}{{MedGemma} technical report}}.
\newblock {\emph{\JournalTitle{ArXiv preprint arXiv:2507.05201}}}  (\bibinfo{year}{2025}).

\bibitem{qiu2024towards}
\bibinfo{author}{Qiu, P.} \emph{et~al.}
\newblock \bibinfo{journal}{\bibinfo{title}{Towards building multilingual language model for medicine}}.
\newblock {\emph{\JournalTitle{Nature Communications}}} \textbf{\bibinfo{volume}{15}}, \bibinfo{pages}{8384} (\bibinfo{year}{2024}).

\bibitem{chen2024rarebench}
\bibinfo{author}{Chen, X.} \emph{et~al.}
\newblock \bibinfo{title}{Rare{B}ench: can llms serve as rare diseases specialists?}
\newblock In \emph{\bibinfo{booktitle}{Proceedings of the 30th ACM SIGKDD conference on knowledge discovery and data mining}}, \bibinfo{pages}{4850--4861} (\bibinfo{year}{2024}).

\bibitem{arora2025healthbench}
\bibinfo{author}{Arora, R.~K.} \emph{et~al.}
\newblock \bibinfo{journal}{\bibinfo{title}{Health{B}ench: Evaluating large language models towards improved human health}}.
\newblock {\emph{\JournalTitle{ArXiv preprint arXiv:2505.08775}}}  (\bibinfo{year}{2025}).

\bibitem{hu2024omnimedvqa}
\bibinfo{author}{Hu, Y.} \emph{et~al.}
\newblock \bibinfo{title}{Omni{M}ed{V}{Q}{A}: A new large-scale comprehensive evaluation benchmark for medical lvlm}.
\newblock In \emph{\bibinfo{booktitle}{Proceedings of the IEEE/CVF Conference on Computer Vision and Pattern Recognition}}, \bibinfo{pages}{22170--22183} (\bibinfo{year}{2024}).

\bibitem{zhang2023pmc}
\bibinfo{author}{Zhang, X.} \emph{et~al.}
\newblock \bibinfo{journal}{\bibinfo{title}{P{M}{C}-{V}{Q}{A}: Visual instruction tuning for medical visual question answering}}.
\newblock {\emph{\JournalTitle{ArXiv preprint arXiv:2305.10415}}}  (\bibinfo{year}{2023}).

\bibitem{lau2018dataset}
\bibinfo{author}{Lau, J.~J.}, \bibinfo{author}{Gayen, S.}, \bibinfo{author}{Ben~Abacha, A.} \& \bibinfo{author}{Demner-Fushman, D.}
\newblock \bibinfo{journal}{\bibinfo{title}{A dataset of clinically generated visual questions and answers about radiology images}}.
\newblock {\emph{\JournalTitle{Scientific data}}} \textbf{\bibinfo{volume}{5}}, \bibinfo{pages}{1--10} (\bibinfo{year}{2018}).

\bibitem{liu2021slake}
\bibinfo{author}{Liu, B.} \emph{et~al.}
\newblock \bibinfo{title}{S{L}{A}{K}{E}: A semantically-labeled knowledge-enhanced dataset for medical visual question answering}.
\newblock In \emph{\bibinfo{booktitle}{2021 IEEE 18th International Symposium on Biomedical Imaging (ISBI)}}, \bibinfo{pages}{1650--1654} (\bibinfo{organization}{IEEE}, \bibinfo{year}{2021}).

\bibitem{he2020pathvqa}
\bibinfo{author}{He, X.}, \bibinfo{author}{Zhang, Y.}, \bibinfo{author}{Mou, L.}, \bibinfo{author}{Xing, E.} \& \bibinfo{author}{Xie, P.}
\newblock \bibinfo{journal}{\bibinfo{title}{Path{VQA}: 30000+ questions for medical visual question answering}}.
\newblock {\emph{\JournalTitle{ArXiv preprint arXiv:2003.10286}}}  (\bibinfo{year}{2020}).

\bibitem{zuo2025medxpertqa}
\bibinfo{author}{Zuo, Y.} \emph{et~al.}
\newblock \bibinfo{title}{Med{X}pert{Q}{A}: Benchmarking expert-level medical reasoning and understanding}.
\newblock In \emph{\bibinfo{booktitle}{Forty-second International Conference on Machine Learning}} (\bibinfo{year}{2025}).

\bibitem{demner2015preparing}
\bibinfo{author}{Demner-Fushman, D.} \emph{et~al.}
\newblock \bibinfo{journal}{\bibinfo{title}{Preparing a collection of radiology examinations for distribution and retrieval}}.
\newblock {\emph{\JournalTitle{Journal of the American Medical Informatics Association}}} \textbf{\bibinfo{volume}{23}}, \bibinfo{pages}{304--310} (\bibinfo{year}{2015}).

\bibitem{zhao2024ratescore}
\bibinfo{author}{Zhao, W.} \emph{et~al.}
\newblock \bibinfo{title}{{RaTEScore}: A metric for radiology report generation}.
\newblock In \emph{\bibinfo{booktitle}{Proceedings of the 2024 Conference on Empirical Methods in Natural Language Processing}}, \bibinfo{pages}{15004--15019} (\bibinfo{year}{2024}).

\bibitem{irvin2019chexpert}
\bibinfo{author}{Irvin, J.} \emph{et~al.}
\newblock \bibinfo{title}{C{h}e{X}pert: A large chest radiograph dataset with uncertainty labels and expert comparison}.
\newblock In \emph{\bibinfo{booktitle}{Proceedings of the AAAI conference on artificial intelligence}}, vol.~\bibinfo{volume}{33}, \bibinfo{pages}{590--597} (\bibinfo{year}{2019}).

\bibitem{johnson2019mimic}
\bibinfo{author}{Johnson, A.~E.} \emph{et~al.}
\newblock \bibinfo{journal}{\bibinfo{title}{{MIMIC-CXR}, a de-identified publicly available database of chest radiographs with free-text reports}}.
\newblock {\emph{\JournalTitle{Scientific data}}} \textbf{\bibinfo{volume}{6}}, \bibinfo{pages}{317} (\bibinfo{year}{2019}).

\bibitem{yang2021medmnist}
\bibinfo{author}{Yang, J.}, \bibinfo{author}{Shi, R.} \& \bibinfo{author}{Ni, B.}
\newblock \bibinfo{title}{Medmnist classification decathlon: A lightweight automl benchmark for medical image analysis}.
\newblock In \emph{\bibinfo{booktitle}{2021 IEEE 18th International Symposium on Biomedical Imaging (ISBI)}}, \bibinfo{pages}{191--195} (\bibinfo{organization}{IEEE}, \bibinfo{year}{2021}).

\bibitem{ji2022amos}
\bibinfo{author}{Ji, Y.} \emph{et~al.}
\newblock \bibinfo{journal}{\bibinfo{title}{A{M}{O}{S}: A large-scale abdominal multi-organ benchmark for versatile medical image segmentation}}.
\newblock {\emph{\JournalTitle{Advances in Neural Information Processing Systems}}} \textbf{\bibinfo{volume}{35}}, \bibinfo{pages}{36722--36732} (\bibinfo{year}{2022}).

\bibitem{gai20253d}
\bibinfo{author}{Gai, X.} \emph{et~al.}
\newblock \bibinfo{journal}{\bibinfo{title}{3{D}-{R}{A}{D}: A comprehensive 3d radiology med-vqa dataset with multi-temporal analysis and diverse diagnostic tasks}}.
\newblock {\emph{\JournalTitle{ArXiv preprint arXiv:2506.11147}}}  (\bibinfo{year}{2025}).

\bibitem{yu2025medframeqa}
\bibinfo{author}{Yu, S.}, \bibinfo{author}{Wang, H.}, \bibinfo{author}{Wu, J.}, \bibinfo{author}{Xie, C.} \& \bibinfo{author}{Zhou, Y.}
\newblock \bibinfo{journal}{\bibinfo{title}{Med{F}rame{Q}{A}: A multi-image medical vqa benchmark for clinical reasoning}}.
\newblock {\emph{\JournalTitle{ArXiv preprint arXiv:2505.16964}}}  (\bibinfo{year}{2025}).

\bibitem{nwoye2023cholectriplet2021}
\bibinfo{author}{Nwoye, C.~I.} \emph{et~al.}
\newblock \bibinfo{journal}{\bibinfo{title}{Cholec{T}riplet2021: A benchmark challenge for surgical action triplet recognition}}.
\newblock {\emph{\JournalTitle{Medical Image Analysis}}} \textbf{\bibinfo{volume}{86}}, \bibinfo{pages}{102803} (\bibinfo{year}{2023}).

\bibitem{allan20202018roboticscenesegmentation}
\bibinfo{author}{Allan, M.} \emph{et~al.}
\newblock \bibinfo{journal}{\bibinfo{title}{2018 robotic scene segmentation challenge}}.
\newblock {\emph{\JournalTitle{ArXiv preprint arXiv:2001.11190}}}  (\bibinfo{year}{2020}).

\bibitem{10.1007/978-3-031-16449-1_42}
\bibinfo{author}{Valderrama, N.} \emph{et~al.}
\newblock \bibinfo{title}{Towards holistic surgical scene understanding}.
\newblock In \emph{\bibinfo{booktitle}{International conference on medical image computing and computer-assisted intervention}}, \bibinfo{pages}{442--452} (\bibinfo{organization}{Springer}, \bibinfo{year}{2022}).

\bibitem{thapa2025well}
\bibinfo{author}{Thapa, R.} \emph{et~al.}
\newblock \bibinfo{journal}{\bibinfo{title}{How well can general vision-language models learn medicine by watching public educational videos?}}
\newblock {\emph{\JournalTitle{ArXiv preprint arXiv:2504.14391}}}  (\bibinfo{year}{2025}).

\bibitem{wang2024mmlu}
\bibinfo{author}{Wang, Y.} \emph{et~al.}
\newblock \bibinfo{journal}{\bibinfo{title}{M{M}{L}{U}-{P}ro: A more robust and challenging multi-task language understanding benchmark}}.
\newblock {\emph{\JournalTitle{Advances in Neural Information Processing Systems}}} \textbf{\bibinfo{volume}{37}}, \bibinfo{pages}{95266--95290} (\bibinfo{year}{2024}).

\bibitem{chen2025benchmarking}
\bibinfo{author}{Chen, H.}, \bibinfo{author}{Fang, Z.}, \bibinfo{author}{Singla, Y.} \& \bibinfo{author}{Dredze, M.}
\newblock \bibinfo{title}{Benchmarking large language models on answering and explaining challenging medical questions}.
\newblock In \emph{\bibinfo{booktitle}{Proceedings of the 2025 Conference of the Nations of the Americas Chapter of the Association for Computational Linguistics: Human Language Technologies (Volume 1: Long Papers)}}, \bibinfo{pages}{3563--3599} (\bibinfo{year}{2025}).

\bibitem{du2025supergpqa}
\bibinfo{author}{Du, X.} \emph{et~al.}
\newblock \bibinfo{journal}{\bibinfo{title}{Super{GPQA}: Scaling llm evaluation across 285 graduate disciplines}}.
\newblock {\emph{\JournalTitle{ArXiv preprint arXiv:2502.14739}}}  (\bibinfo{year}{2025}).

\bibitem{jin2019pubmedqa}
\bibinfo{author}{Jin, Q.}, \bibinfo{author}{Dhingra, B.}, \bibinfo{author}{Liu, Z.}, \bibinfo{author}{Cohen, W.} \& \bibinfo{author}{Lu, X.}
\newblock \bibinfo{title}{{PubMedQA}: A dataset for biomedical research question answering}.
\newblock In \emph{\bibinfo{booktitle}{Proceedings of the 2019 Conference on Empirical Methods in Natural Language Processing and the 9th International Joint Conference on Natural Language Processing (EMNLP-IJCNLP)}}, \bibinfo{pages}{2567--2577} (\bibinfo{year}{2019}).

\bibitem{pal2022medmcqa}
\bibinfo{author}{Pal, A.}, \bibinfo{author}{Umapathi, L.~K.} \& \bibinfo{author}{Sankarasubbu, M.}
\newblock \bibinfo{title}{Med{M}{C}{Q}{A}: A large-scale multi-subject multi-choice dataset for medical domain question answering}.
\newblock In \emph{\bibinfo{booktitle}{Conference on health, inference, and learning}}, \bibinfo{pages}{248--260} (\bibinfo{organization}{PMLR}, \bibinfo{year}{2022}).

\bibitem{hendrycks2020measuring}
\bibinfo{author}{Hendrycks, D.} \emph{et~al.}
\newblock \bibinfo{title}{Measuring massive multitask language understanding}.
\newblock In \emph{\bibinfo{booktitle}{International Conference on Learning Representations}} (\bibinfo{year}{2021}).

\bibitem{hoffmann2022training}
\bibinfo{author}{Hoffmann, J.} \emph{et~al.}
\newblock \bibinfo{title}{Training compute-optimal large language models}.
\newblock In \emph{\bibinfo{booktitle}{Proceedings of the 36th International Conference on Neural Information Processing Systems}}, \bibinfo{pages}{30016--30030} (\bibinfo{year}{2022}).

\bibitem{henighan2020scaling}
\bibinfo{author}{Henighan, T.} \emph{et~al.}
\newblock \bibinfo{journal}{\bibinfo{title}{Scaling laws for autoregressive generative modeling}}.
\newblock {\emph{\JournalTitle{ArXiv preprint arXiv:2010.14701}}}  (\bibinfo{year}{2020}).

\bibitem{4-lu2024multimodal}
\bibinfo{author}{Lu, M.~Y.} \emph{et~al.}
\newblock \bibinfo{journal}{\bibinfo{title}{A multimodal generative ai copilot for human pathology}}.
\newblock {\emph{\JournalTitle{Nature}}} \textbf{\bibinfo{volume}{634}}, \bibinfo{pages}{466--473} (\bibinfo{year}{2024}).

\bibitem{price2019privacy}
\bibinfo{author}{Price, W.~N.} \& \bibinfo{author}{Cohen, I.~G.}
\newblock \bibinfo{journal}{\bibinfo{title}{Privacy in the age of medical big data}}.
\newblock {\emph{\JournalTitle{Nature Medicine}}} \textbf{\bibinfo{volume}{25}}, \bibinfo{pages}{37--43} (\bibinfo{year}{2019}).

\bibitem{xu2025qwen3}
\bibinfo{author}{Xu, J.} \emph{et~al.}
\newblock \bibinfo{journal}{\bibinfo{title}{Qwen3-{O}mni technical report}}.
\newblock {\emph{\JournalTitle{ArXiv preprint arXiv:2509.17765}}}  (\bibinfo{year}{2025}).

\bibitem{guo2025seed1}
\bibinfo{author}{Guo, D.} \emph{et~al.}
\newblock \bibinfo{journal}{\bibinfo{title}{Seed1. 5-{VL} technical report}}.
\newblock {\emph{\JournalTitle{ArXiv preprint arXiv:2505.07062}}}  (\bibinfo{year}{2025}).

\bibitem{acosta2022multimodal}
\bibinfo{author}{Acosta, J.~N.}, \bibinfo{author}{Falcone, G.~J.}, \bibinfo{author}{Rajpurkar, P.} \& \bibinfo{author}{Topol, E.~J.}
\newblock \bibinfo{journal}{\bibinfo{title}{Multimodal biomedical ai}}.
\newblock {\emph{\JournalTitle{Nature Medicine}}} \textbf{\bibinfo{volume}{28}}, \bibinfo{pages}{1773--1784} (\bibinfo{year}{2022}).

\bibitem{sennrich2016neural}
\bibinfo{author}{Sennrich, R.}, \bibinfo{author}{Haddow, B.} \& \bibinfo{author}{Birch, A.}
\newblock \bibinfo{title}{Neural machine translation of rare words with subword units}.
\newblock In \emph{\bibinfo{booktitle}{Proceedings of the 54th Annual Meeting of the Association for Computational Linguistics (Volume 1: Long Papers)}}, \bibinfo{pages}{1715--1725} (\bibinfo{year}{2016}).

\end{thebibliography}


\begin{thebibliography}{10}
\urlstyle{rm}
\expandafter\ifx\csname url\endcsname\relax
  \def\url#1{\texttt{#1}}\fi
\expandafter\ifx\csname urlprefix\endcsname\relax\def\urlprefix{URL }\fi
\expandafter\ifx\csname doiprefix\endcsname\relax\def\doiprefix{DOI: }\fi
\providecommand{\bibinfo}[2]{#2}
\providecommand{\eprint}[2][]{\url{#2}}

\bibitem{kim2021comprehensive}
\bibinfo{author}{Kim, J.}, \bibinfo{author}{Park, S.}, \bibinfo{author}{Min, D.} \& \bibinfo{author}{Kim, W.}
\newblock \bibinfo{journal}{\bibinfo{title}{Comprehensive survey of recent drug discovery using deep learning}}.
\newblock {\emph{\JournalTitle{International Journal of Molecular Sciences}}} \textbf{\bibinfo{volume}{22}}, \bibinfo{pages}{9983} (\bibinfo{year}{2021}).

\bibitem{volk2020biosystems}
\bibinfo{author}{Volk, M.~J.} \emph{et~al.}
\newblock \bibinfo{journal}{\bibinfo{title}{Biosystems design by machine learning}}.
\newblock {\emph{\JournalTitle{ACS synthetic biology}}} \textbf{\bibinfo{volume}{9}}, \bibinfo{pages}{1514--1533} (\bibinfo{year}{2020}).

\bibitem{mazurenko2019machine}
\bibinfo{author}{Mazurenko, S.}, \bibinfo{author}{Prokop, Z.} \& \bibinfo{author}{Damborsky, J.}
\newblock \bibinfo{journal}{\bibinfo{title}{Machine learning in enzyme engineering}}.
\newblock {\emph{\JournalTitle{ACS Catalysis}}} \textbf{\bibinfo{volume}{10}}, \bibinfo{pages}{1210--1223} (\bibinfo{year}{2019}).

\bibitem{abramson2024accurate}
\bibinfo{author}{Abramson, J.} \emph{et~al.}
\newblock \bibinfo{journal}{\bibinfo{title}{Accurate structure prediction of biomolecular interactions with alphafold 3}}.
\newblock {\emph{\JournalTitle{Nature}}} \bibinfo{pages}{1--3} (\bibinfo{year}{2024}).

\bibitem{krishna2024generalized}
\bibinfo{author}{Krishna, R.} \emph{et~al.}
\newblock \bibinfo{journal}{\bibinfo{title}{Generalized biomolecular modeling and design with rosettafold all-atom}}.
\newblock {\emph{\JournalTitle{Science}}} \textbf{\bibinfo{volume}{384}}, \bibinfo{pages}{eadl2528} (\bibinfo{year}{2024}).

\bibitem{DBLP:journals/corr/abs-2302-09419}
\bibinfo{author}{Zhou, C.} \emph{et~al.}
\newblock \bibinfo{journal}{\bibinfo{title}{A comprehensive survey on pretrained foundation models: {A} history from {BERT} to chatgpt}}.
\newblock {\emph{\JournalTitle{CoRR}}} \textbf{\bibinfo{volume}{abs/2302.09419}} (\bibinfo{year}{2023}).

\bibitem{DBLP:journals/corr/abs-2307-09288}
\bibinfo{author}{Touvron, H.} \emph{et~al.}
\newblock \bibinfo{journal}{\bibinfo{title}{Llama 2: Open foundation and fine-tuned chat models}}.
\newblock {\emph{\JournalTitle{CoRR}}} \textbf{\bibinfo{volume}{abs/2307.09288}} (\bibinfo{year}{2023}).

\bibitem{DBLP:journals/corr/abs-2303-08774}
\bibinfo{author}{OpenAI}.
\newblock \bibinfo{journal}{\bibinfo{title}{{GPT-4} technical report}}.
\newblock {\emph{\JournalTitle{CoRR}}} \textbf{\bibinfo{volume}{abs/2303.08774}} (\bibinfo{year}{2023}).

\bibitem{zhang2025scientific}
\bibinfo{author}{Zhang, Q.} \emph{et~al.}
\newblock \bibinfo{journal}{\bibinfo{title}{Scientific large language models: A survey on biological \& chemical domains}}.
\newblock {\emph{\JournalTitle{ACM Computing Surveys}}} \textbf{\bibinfo{volume}{57}}, \bibinfo{pages}{1--38} (\bibinfo{year}{2025}).

\bibitem{weininger1988smiles}
\bibinfo{author}{Weininger, D.}
\newblock \bibinfo{journal}{\bibinfo{title}{Smiles, a chemical language and information system. 1. introduction to methodology and encoding rules}}.
\newblock {\emph{\JournalTitle{Journal of chemical information and computer sciences}}} \textbf{\bibinfo{volume}{28}}, \bibinfo{pages}{31--36} (\bibinfo{year}{1988}).

\bibitem{DBLP:journals/mlst/KrennHNFA20}
\bibinfo{author}{Krenn, M.}, \bibinfo{author}{H{\"{a}}se, F.}, \bibinfo{author}{Nigam, A.}, \bibinfo{author}{Friederich, P.} \& \bibinfo{author}{Aspuru{-}Guzik, A.}
\newblock \bibinfo{journal}{\bibinfo{title}{Self-referencing embedded strings {(SELFIES):} {A} 100{\%} robust molecular string representation}}.
\newblock {\emph{\JournalTitle{Mach. Learn. Sci. Technol.}}} \textbf{\bibinfo{volume}{1}}, \bibinfo{pages}{45024} (\bibinfo{year}{2020}).

\bibitem{pearson1994using}
\bibinfo{author}{Pearson, W.~R.}
\newblock \bibinfo{journal}{\bibinfo{title}{Using the fasta program to search protein and dna sequence databases}}.
\newblock {\emph{\JournalTitle{Computer Analysis of Sequence Data: Part I}}} \bibinfo{pages}{307--331} (\bibinfo{year}{1994}).

\bibitem{jumper2021highly}
\bibinfo{author}{Jumper, J.} \emph{et~al.}
\newblock \bibinfo{journal}{\bibinfo{title}{Highly accurate protein structure prediction with alphafold}}.
\newblock {\emph{\JournalTitle{Nature}}} \textbf{\bibinfo{volume}{596}}, \bibinfo{pages}{583--589} (\bibinfo{year}{2021}).

\bibitem{DBLP:conf/nips/Ouyang0JAWMZASR22}
\bibinfo{author}{Ouyang, L.} \emph{et~al.}
\newblock \bibinfo{title}{Training language models to follow instructions with human feedback}.
\newblock In \emph{\bibinfo{booktitle}{NeurIPS}} (\bibinfo{year}{2022}).

\bibitem{DBLP:conf/emnlp/EdwardsLRHCJ22}
\bibinfo{author}{Edwards, C.} \emph{et~al.}
\newblock \bibinfo{title}{Translation between molecules and natural language}.
\newblock In \emph{\bibinfo{booktitle}{{EMNLP}}}, \bibinfo{pages}{375--413} (\bibinfo{publisher}{Association for Computational Linguistics}, \bibinfo{year}{2022}).

\bibitem{DBLP:conf/acl/WangZDQZLC24}
\bibinfo{author}{Wang, Z.} \emph{et~al.}
\newblock \bibinfo{title}{Instructprotein: Aligning human and protein language via knowledge instruction}.
\newblock In \emph{\bibinfo{booktitle}{{ACL} {(1)}}}, \bibinfo{pages}{1114--1136} (\bibinfo{publisher}{Association for Computational Linguistics}, \bibinfo{year}{2024}).

\bibitem{DBLP:conf/emnlp/PeiZZWGWXY23}
\bibinfo{author}{Pei, Q.} \emph{et~al.}
\newblock \bibinfo{title}{Biot5: Enriching cross-modal integration in biology with chemical knowledge and natural language associations}.
\newblock In \emph{\bibinfo{booktitle}{{EMNLP}}}, \bibinfo{pages}{1102--1123} (\bibinfo{publisher}{Association for Computational Linguistics}, \bibinfo{year}{2023}).

\bibitem{DBLP:conf/iclr/FangL0LH0FC24}
\bibinfo{author}{Fang, Y.} \emph{et~al.}
\newblock \bibinfo{title}{Mol-instructions: {A} large-scale biomolecular instruction dataset for large language models}.
\newblock In \emph{\bibinfo{booktitle}{{ICLR}}} (\bibinfo{publisher}{OpenReview.net}, \bibinfo{year}{2024}).

\bibitem{DBLP:conf/acl/PeiWGLFZ00024}
\bibinfo{author}{Pei, Q.} \emph{et~al.}
\newblock \bibinfo{title}{Biot5+: Towards generalized biological understanding with {IUPAC} integration and multi-task tuning}.
\newblock In \emph{\bibinfo{booktitle}{{ACL} (Findings)}}, \bibinfo{pages}{1216--1240} (\bibinfo{publisher}{Association for Computational Linguistics}, \bibinfo{year}{2024}).

\bibitem{DBLP:journals/corr/abs-2308-09442}
\bibinfo{author}{Luo, Y.} \emph{et~al.}
\newblock \bibinfo{journal}{\bibinfo{title}{Biomedgpt: Open multimodal generative pre-trained transformer for biomedicine}}.
\newblock {\emph{\JournalTitle{CoRR}}} \textbf{\bibinfo{volume}{abs/2308.09442}} (\bibinfo{year}{2023}).

\bibitem{DBLP:conf/iclr/LiuWYW0GX24}
\bibinfo{author}{Liu, S.} \emph{et~al.}
\newblock \bibinfo{title}{Conversational drug editing using retrieval and domain feedback}.
\newblock In \emph{\bibinfo{booktitle}{{ICLR}}} (\bibinfo{publisher}{OpenReview.net}, \bibinfo{year}{2024}).

\bibitem{probst2018smilesdrawer}
\bibinfo{author}{Probst, D.} \& \bibinfo{author}{Reymond, J.-L.}
\newblock \bibinfo{journal}{\bibinfo{title}{Smilesdrawer: parsing and drawing smiles-encoded molecular structures using client-side javascript}}.
\newblock {\emph{\JournalTitle{Journal of chemical information and modeling}}} \textbf{\bibinfo{volume}{58}}, \bibinfo{pages}{1--7} (\bibinfo{year}{2018}).

\bibitem{PyMOL}
\bibinfo{author}{{Schr\"odinger, LLC}}.
\newblock \bibinfo{title}{The {PyMOL} molecular graphics system, version~3.0} (\bibinfo{year}{2024}).

\bibitem{kroll2023general}
\bibinfo{author}{Kroll, A.}, \bibinfo{author}{Ranjan, S.}, \bibinfo{author}{Engqvist, M.~K.} \& \bibinfo{author}{Lercher, M.~J.}
\newblock \bibinfo{journal}{\bibinfo{title}{A general model to predict small molecule substrates of enzymes based on machine and deep learning}}.
\newblock {\emph{\JournalTitle{Nature communications}}} \textbf{\bibinfo{volume}{14}}, \bibinfo{pages}{2787} (\bibinfo{year}{2023}).

\bibitem{DBLP:conf/nips/VaswaniSPUJGKP17}
\bibinfo{author}{Vaswani, A.} \emph{et~al.}
\newblock \bibinfo{title}{Attention is all you need}.
\newblock In \emph{\bibinfo{booktitle}{{NIPS}}}, \bibinfo{pages}{5998--6008} (\bibinfo{year}{2017}).

\bibitem{hastings2016chebi}
\bibinfo{author}{Hastings, J.} \emph{et~al.}
\newblock \bibinfo{journal}{\bibinfo{title}{Chebi in 2016: Improved services and an expanding collection of metabolites}}.
\newblock {\emph{\JournalTitle{Nucleic acids research}}} \textbf{\bibinfo{volume}{44}}, \bibinfo{pages}{D1214--D1219} (\bibinfo{year}{2016}).

\bibitem{molregpt}
\bibinfo{author}{Li, J.} \emph{et~al.}
\newblock \bibinfo{journal}{\bibinfo{title}{Empowering molecule discovery for molecule-caption translation with large language models: A chatgpt perspective}}.
\newblock {\emph{\JournalTitle{arXiv preprint arXiv:2306.06615}}}  (\bibinfo{year}{2023}).

\bibitem{DBLP:journals/corr/abs-2401-14818}
\bibinfo{author}{Zhao, Z.} \emph{et~al.}
\newblock \bibinfo{journal}{\bibinfo{title}{Chemdfm: Dialogue foundation model for chemistry}}.
\newblock {\emph{\JournalTitle{CoRR}}} \textbf{\bibinfo{volume}{abs/2401.14818}} (\bibinfo{year}{2024}).

\bibitem{DBLP:journals/corr/abs-2311-16208}
\bibinfo{author}{Cao, H.}, \bibinfo{author}{Liu, Z.}, \bibinfo{author}{Lu, X.}, \bibinfo{author}{Yao, Y.} \& \bibinfo{author}{Li, Y.}
\newblock \bibinfo{journal}{\bibinfo{title}{Instructmol: Multi-modal integration for building a versatile and reliable molecular assistant in drug discovery}}.
\newblock {\emph{\JournalTitle{CoRR}}} \textbf{\bibinfo{volume}{abs/2311.16208}} (\bibinfo{year}{2023}).

\bibitem{DBLP:conf/acl/LiuZ0ZWKC24}
\bibinfo{author}{Liu, Z.} \emph{et~al.}
\newblock \bibinfo{title}{Prott3: Protein-to-text generation for text-based protein understanding}.
\newblock In \emph{\bibinfo{booktitle}{{ACL} {(1)}}}, \bibinfo{pages}{5949--5966} (\bibinfo{publisher}{Association for Computational Linguistics}, \bibinfo{year}{2024}).

\bibitem{DBLP:journals/corr/abs-2302-04611}
\bibinfo{author}{Liu, S.} \emph{et~al.}
\newblock \bibinfo{journal}{\bibinfo{title}{A text-guided protein design framework}}.
\newblock {\emph{\JournalTitle{CoRR}}} \textbf{\bibinfo{volume}{abs/2302.04611}} (\bibinfo{year}{2023}).

\bibitem{anderson2003process}
\bibinfo{author}{Anderson, A.~C.}
\newblock \bibinfo{journal}{\bibinfo{title}{The process of structure-based drug design}}.
\newblock {\emph{\JournalTitle{Chemistry \& biology}}} \textbf{\bibinfo{volume}{10}}, \bibinfo{pages}{787--797} (\bibinfo{year}{2003}).

\bibitem{DBLP:conf/icml/PengLGXPM22}
\bibinfo{author}{Peng, X.} \emph{et~al.}
\newblock \bibinfo{title}{Pocket2mol: Efficient molecular sampling based on 3d protein pockets}.
\newblock In \emph{\bibinfo{booktitle}{{ICML}}}, vol. \bibinfo{volume}{162} of \emph{\bibinfo{series}{Proceedings of Machine Learning Research}}, \bibinfo{pages}{17644--17655} (\bibinfo{publisher}{{PMLR}}, \bibinfo{year}{2022}).

\bibitem{DBLP:conf/nips/LuoGMP21}
\bibinfo{author}{Luo, S.}, \bibinfo{author}{Guan, J.}, \bibinfo{author}{Ma, J.} \& \bibinfo{author}{Peng, J.}
\newblock \bibinfo{title}{A 3d generative model for structure-based drug design}.
\newblock In \emph{\bibinfo{booktitle}{NeurIPS}}, \bibinfo{pages}{6229--6239} (\bibinfo{year}{2021}).

\bibitem{DBLP:conf/iclr/GuanQPS0M23}
\bibinfo{author}{Guan, J.} \emph{et~al.}
\newblock \bibinfo{title}{3d equivariant diffusion for target-aware molecule generation and affinity prediction}.
\newblock In \emph{\bibinfo{booktitle}{{ICLR}}} (\bibinfo{publisher}{OpenReview.net}, \bibinfo{year}{2023}).

\bibitem{li2023druggpt}
\bibinfo{author}{Li, Y.} \emph{et~al.}
\newblock \bibinfo{journal}{\bibinfo{title}{Druggpt: A gpt-based strategy for designing potential ligands targeting specific proteins}}.
\newblock {\emph{\JournalTitle{bioRxiv}}} \bibinfo{pages}{2023--06} (\bibinfo{year}{2023}).

\bibitem{bar2011moderately}
\bibinfo{author}{Bar-Even, A.} \emph{et~al.}
\newblock \bibinfo{journal}{\bibinfo{title}{The moderately efficient enzyme: evolutionary and physicochemical trends shaping enzyme parameters}}.
\newblock {\emph{\JournalTitle{Biochemistry}}} \textbf{\bibinfo{volume}{50}}, \bibinfo{pages}{4402--4410} (\bibinfo{year}{2011}).

\bibitem{DBLP:conf/iclr/HuSWALWWC22}
\bibinfo{author}{Hu, E.~J.} \emph{et~al.}
\newblock \bibinfo{title}{Lora: Low-rank adaptation of large language models}.
\newblock In \emph{\bibinfo{booktitle}{{ICLR}}} (\bibinfo{publisher}{OpenReview.net}, \bibinfo{year}{2022}).

\bibitem{DBLP:conf/icml/GilmerSRVD17}
\bibinfo{author}{Gilmer, J.}, \bibinfo{author}{Schoenholz, S.~S.}, \bibinfo{author}{Riley, P.~F.}, \bibinfo{author}{Vinyals, O.} \& \bibinfo{author}{Dahl, G.~E.}
\newblock \bibinfo{title}{Neural message passing for quantum chemistry}.
\newblock In \emph{\bibinfo{booktitle}{{ICML}}}, vol.~\bibinfo{volume}{70} of \emph{\bibinfo{series}{Proceedings of Machine Learning Research}}, \bibinfo{pages}{1263--1272} (\bibinfo{publisher}{{PMLR}}, \bibinfo{year}{2017}).

\bibitem{DBLP:conf/iclr/ZhouGDZXWZK23}
\bibinfo{author}{Zhou, G.} \emph{et~al.}
\newblock \bibinfo{title}{Uni-mol: {A} universal 3d molecular representation learning framework}.
\newblock In \emph{\bibinfo{booktitle}{{ICLR}}} (\bibinfo{publisher}{OpenReview.net}, \bibinfo{year}{2023}).

\bibitem{lin2022language}
\bibinfo{author}{Lin, Z.} \emph{et~al.}
\newblock \bibinfo{journal}{\bibinfo{title}{Language models of protein sequences at the scale of evolution enable accurate structure prediction}}.
\newblock {\emph{\JournalTitle{BioRxiv}}} \textbf{\bibinfo{volume}{2022}}, \bibinfo{pages}{500902} (\bibinfo{year}{2022}).

\bibitem{DBLP:conf/iclr/XuHLJ19}
\bibinfo{author}{Xu, K.}, \bibinfo{author}{Hu, W.}, \bibinfo{author}{Leskovec, J.} \& \bibinfo{author}{Jegelka, S.}
\newblock \bibinfo{title}{How powerful are graph neural networks?}
\newblock In \emph{\bibinfo{booktitle}{{ICLR}}} (\bibinfo{publisher}{OpenReview.net}, \bibinfo{year}{2019}).

\bibitem{DBLP:conf/iclr/HuLGZLPL20}
\bibinfo{author}{Hu, W.} \emph{et~al.}
\newblock \bibinfo{title}{Strategies for pre-training graph neural networks}.
\newblock In \emph{\bibinfo{booktitle}{{ICLR}}} (\bibinfo{publisher}{OpenReview.net}, \bibinfo{year}{2020}).

\bibitem{DBLP:conf/nips/WangLWS0L23}
\bibinfo{author}{Wang, Y.} \emph{et~al.}
\newblock \bibinfo{title}{Geometric transformer with interatomic positional encoding}.
\newblock In \emph{\bibinfo{booktitle}{NeurIPS}} (\bibinfo{year}{2023}).

\bibitem{su2024saprot}
\bibinfo{author}{Su, J.} \emph{et~al.}
\newblock \bibinfo{title}{Saprot: Protein language modeling with structure-aware vocabulary}.
\newblock In \emph{\bibinfo{booktitle}{The Twelfth International Conference on Learning Representations}} (\bibinfo{year}{2024}).

\bibitem{DBLP:conf/nips/ZhangLWLL21}
\bibinfo{author}{Zhang, Z.}, \bibinfo{author}{Liu, Q.}, \bibinfo{author}{Wang, H.}, \bibinfo{author}{Lu, C.} \& \bibinfo{author}{Lee, C.}
\newblock \bibinfo{title}{Motif-based graph self-supervised learning for molecular property prediction}.
\newblock In \emph{\bibinfo{booktitle}{NeurIPS}}, \bibinfo{pages}{15870--15882} (\bibinfo{year}{2021}).

\bibitem{li2023knowledge}
\bibinfo{author}{Li, H.} \emph{et~al.}
\newblock \bibinfo{journal}{\bibinfo{title}{A knowledge-guided pre-training framework for improving molecular representation learning}}.
\newblock {\emph{\JournalTitle{Nature Communications}}} \textbf{\bibinfo{volume}{14}}, \bibinfo{pages}{7568} (\bibinfo{year}{2023}).

\bibitem{grant2011fimo}
\bibinfo{author}{Grant, C.~E.}, \bibinfo{author}{Bailey, T.~L.} \& \bibinfo{author}{Noble, W.~S.}
\newblock \bibinfo{journal}{\bibinfo{title}{Fimo: scanning for occurrences of a given motif}}.
\newblock {\emph{\JournalTitle{Bioinformatics}}} \textbf{\bibinfo{volume}{27}}, \bibinfo{pages}{1017--1018} (\bibinfo{year}{2011}).

\bibitem{rogers2010extended}
\bibinfo{author}{Rogers, D.} \& \bibinfo{author}{Hahn, M.}
\newblock \bibinfo{journal}{\bibinfo{title}{Extended-connectivity fingerprints}}.
\newblock {\emph{\JournalTitle{Journal of chemical information and modeling}}} \textbf{\bibinfo{volume}{50}}, \bibinfo{pages}{742--754} (\bibinfo{year}{2010}).

\bibitem{boeckmann2003swiss}
\bibinfo{author}{Boeckmann, B.} \emph{et~al.}
\newblock \bibinfo{journal}{\bibinfo{title}{The swiss-prot protein knowledgebase and its supplement trembl in 2003}}.
\newblock {\emph{\JournalTitle{Nucleic acids research}}} \textbf{\bibinfo{volume}{31}}, \bibinfo{pages}{365--370} (\bibinfo{year}{2003}).

\bibitem{radford2018improving}
\bibinfo{author}{Radford, A.}, \bibinfo{author}{Narasimhan, K.}, \bibinfo{author}{Salimans, T.}, \bibinfo{author}{Sutskever, I.} \emph{et~al.}
\newblock \bibinfo{journal}{\bibinfo{title}{Improving language understanding by generative pre-training}}.
\newblock {\emph{\JournalTitle{OpenAI}}}  (\bibinfo{year}{2018}).

\bibitem{kim2016pubchem}
\bibinfo{author}{Kim, S.} \emph{et~al.}
\newblock \bibinfo{journal}{\bibinfo{title}{Pubchem substance and compound databases}}.
\newblock {\emph{\JournalTitle{Nucleic acids research}}} \textbf{\bibinfo{volume}{44}}, \bibinfo{pages}{D1202--D1213} (\bibinfo{year}{2016}).

\bibitem{suzek2007uniref}
\bibinfo{author}{Suzek, B.~E.}, \bibinfo{author}{Huang, H.}, \bibinfo{author}{McGarvey, P.}, \bibinfo{author}{Mazumder, R.} \& \bibinfo{author}{Wu, C.~H.}
\newblock \bibinfo{journal}{\bibinfo{title}{Uniref: comprehensive and non-redundant uniprot reference clusters}}.
\newblock {\emph{\JournalTitle{Bioinformatics}}} \textbf{\bibinfo{volume}{23}}, \bibinfo{pages}{1282--1288} (\bibinfo{year}{2007}).

\bibitem{white2020pubmed}
\bibinfo{author}{White, J.}
\newblock \bibinfo{journal}{\bibinfo{title}{Pubmed 2.0}}.
\newblock {\emph{\JournalTitle{Medical reference services quarterly}}} \textbf{\bibinfo{volume}{39}}, \bibinfo{pages}{382--387} (\bibinfo{year}{2020}).

\bibitem{sever2019biorxiv}
\bibinfo{author}{Sever, R.} \emph{et~al.}
\newblock \bibinfo{journal}{\bibinfo{title}{biorxiv: the preprint server for biology}}.
\newblock {\emph{\JournalTitle{BioRxiv}}} \bibinfo{pages}{833400} (\bibinfo{year}{2019}).

\bibitem{mudrak2022five}
\bibinfo{author}{Mudrak, B.} \emph{et~al.}
\newblock \bibinfo{title}{Five years of chemrxiv: Where we are and where we go from here} (\bibinfo{year}{2022}).

\bibitem{mcnaught1997compendium}
\bibinfo{author}{McNaught, A.~D.}, \bibinfo{author}{Wilkinson, A.} \emph{et~al.}
\newblock \emph{\bibinfo{title}{Compendium of chemical terminology}}, vol. \bibinfo{volume}{1669} (\bibinfo{publisher}{Blackwell Science Oxford}, \bibinfo{year}{1997}).

\bibitem{uniprot2018uniprot}
\bibinfo{author}{UniProt~Consortium, T.}
\newblock \bibinfo{journal}{\bibinfo{title}{Uniprot: the universal protein knowledgebase}}.
\newblock {\emph{\JournalTitle{Nucleic acids research}}} \textbf{\bibinfo{volume}{46}}, \bibinfo{pages}{2699--2699} (\bibinfo{year}{2018}).

\bibitem{gilson2016bindingdb}
\bibinfo{author}{Gilson, M.~K.} \emph{et~al.}
\newblock \bibinfo{journal}{\bibinfo{title}{Bindingdb in 2015: a public database for medicinal chemistry, computational chemistry and systems pharmacology}}.
\newblock {\emph{\JournalTitle{Nucleic acids research}}} \textbf{\bibinfo{volume}{44}}, \bibinfo{pages}{D1045--D1053} (\bibinfo{year}{2016}).

\bibitem{uludougan2022exploiting}
\bibinfo{author}{Uludo{\u{g}}an, G.}, \bibinfo{author}{Ozkirimli, E.}, \bibinfo{author}{Ulgen, K.~O.}, \bibinfo{author}{Karal{\i}, N.} \& \bibinfo{author}{{\"O}zg{\"u}r, A.}
\newblock \bibinfo{journal}{\bibinfo{title}{Exploiting pretrained biochemical language models for targeted drug design}}.
\newblock {\emph{\JournalTitle{Bioinformatics}}} \textbf{\bibinfo{volume}{38}}, \bibinfo{pages}{ii155--ii161} (\bibinfo{year}{2022}).

\bibitem{bansal2022rhea}
\bibinfo{author}{Bansal, P.} \emph{et~al.}
\newblock \bibinfo{journal}{\bibinfo{title}{Rhea, the reaction knowledgebase in 2022}}.
\newblock {\emph{\JournalTitle{Nucleic acids research}}} \textbf{\bibinfo{volume}{50}}, \bibinfo{pages}{D693--D700} (\bibinfo{year}{2022}).

\bibitem{landrum2013rdkit}
\bibinfo{author}{Landrum, G.} \emph{et~al.}
\newblock \bibinfo{journal}{\bibinfo{title}{Rdkit: A software suite for cheminformatics, computational chemistry, and predictive modeling}}.
\newblock {\emph{\JournalTitle{Greg Landrum}}} \textbf{\bibinfo{volume}{8}}, \bibinfo{pages}{5281} (\bibinfo{year}{2013}).

\bibitem{riniker2015better}
\bibinfo{author}{Riniker, S.} \& \bibinfo{author}{Landrum, G.~A.}
\newblock \bibinfo{journal}{\bibinfo{title}{Better informed distance geometry: using what we know to improve conformation generation}}.
\newblock {\emph{\JournalTitle{Journal of chemical information and modeling}}} \textbf{\bibinfo{volume}{55}}, \bibinfo{pages}{2562--2574} (\bibinfo{year}{2015}).

\bibitem{halgren1996merck}
\bibinfo{author}{Halgren, T.~A.}
\newblock \bibinfo{journal}{\bibinfo{title}{Merck molecular force field. i. basis, form, scope, parameterization, and performance of mmff94}}.
\newblock {\emph{\JournalTitle{Journal of computational chemistry}}} \textbf{\bibinfo{volume}{17}}, \bibinfo{pages}{490--519} (\bibinfo{year}{1996}).

\bibitem{varadi2022alphafold}
\bibinfo{author}{Varadi, M.} \emph{et~al.}
\newblock \bibinfo{journal}{\bibinfo{title}{Alphafold protein structure database: massively expanding the structural coverage of protein-sequence space with high-accuracy models}}.
\newblock {\emph{\JournalTitle{Nucleic acids research}}} \textbf{\bibinfo{volume}{50}}, \bibinfo{pages}{D439--D444} (\bibinfo{year}{2022}).

\bibitem{DBLP:conf/nips/PaszkeGMLBCKLGA19}
\bibinfo{author}{Paszke, A.} \emph{et~al.}
\newblock \bibinfo{title}{Pytorch: An imperative style, high-performance deep learning library}.
\newblock In \emph{\bibinfo{booktitle}{NeurIPS}}, \bibinfo{pages}{8024--8035} (\bibinfo{year}{2019}).

\bibitem{DBLP:conf/sc/RajbhandariRRH20}
\bibinfo{author}{Rajbhandari, S.}, \bibinfo{author}{Rasley, J.}, \bibinfo{author}{Ruwase, O.} \& \bibinfo{author}{He, Y.}
\newblock \bibinfo{title}{Zero: memory optimizations toward training trillion parameter models}.
\newblock In \emph{\bibinfo{booktitle}{{SC}}}, \bibinfo{pages}{20} (\bibinfo{publisher}{{IEEE/ACM}}, \bibinfo{year}{2020}).

\bibitem{DBLP:conf/acl/PapineniRWZ02}
\bibinfo{author}{Papineni, K.}, \bibinfo{author}{Roukos, S.}, \bibinfo{author}{Ward, T.} \& \bibinfo{author}{Zhu, W.}
\newblock \bibinfo{title}{Bleu: a method for automatic evaluation of machine translation}.
\newblock In \emph{\bibinfo{booktitle}{{ACL}}}, \bibinfo{pages}{311--318} (\bibinfo{publisher}{{ACL}}, \bibinfo{year}{2002}).

\bibitem{lin2004rouge}
\bibinfo{author}{Lin, C.-Y.}
\newblock \bibinfo{title}{Rouge: A package for automatic evaluation of summaries}.
\newblock In \emph{\bibinfo{booktitle}{Text summarization branches out}}, \bibinfo{pages}{74--81} (\bibinfo{year}{2004}).

\bibitem{DBLP:conf/acl/BanerjeeL05}
\bibinfo{author}{Banerjee, S.} \& \bibinfo{author}{Lavie, A.}
\newblock \bibinfo{title}{{METEOR:} an automatic metric for {MT} evaluation with improved correlation with human judgments}.
\newblock In \emph{\bibinfo{booktitle}{IEEvaluation@ACL}}, \bibinfo{pages}{65--72} (\bibinfo{publisher}{Association for Computational Linguistics}, \bibinfo{year}{2005}).

\bibitem{miller2009levenshtein}
\bibinfo{author}{Miller, F.~P.}, \bibinfo{author}{Vandome, A.~F.} \& \bibinfo{author}{McBrewster, J.}
\newblock \bibinfo{title}{Levenshtein distance: Information theory, computer science, string (computer science), string metric, damerau? levenshtein distance, spell checker, hamming distance} (\bibinfo{year}{2009}).

\bibitem{durant2002reoptimization}
\bibinfo{author}{Durant, J.~L.}, \bibinfo{author}{Leland, B.~A.}, \bibinfo{author}{Henry, D.~R.} \& \bibinfo{author}{Nourse, J.~G.}
\newblock \bibinfo{journal}{\bibinfo{title}{Reoptimization of mdl keys for use in drug discovery}}.
\newblock {\emph{\JournalTitle{Journal of chemical information and computer sciences}}} \textbf{\bibinfo{volume}{42}}, \bibinfo{pages}{1273--1280} (\bibinfo{year}{2002}).

\bibitem{schneider2015get}
\bibinfo{author}{Schneider, N.}, \bibinfo{author}{Sayle, R.~A.} \& \bibinfo{author}{Landrum, G.~A.}
\newblock \bibinfo{journal}{\bibinfo{title}{Get your atoms in order an open-source implementation of a novel and robust molecular canonicalization algorithm}}.
\newblock {\emph{\JournalTitle{Journal of chemical information and modeling}}} \textbf{\bibinfo{volume}{55}}, \bibinfo{pages}{2111--2120} (\bibinfo{year}{2015}).

\bibitem{bajusz2015tanimoto}
\bibinfo{author}{Bajusz, D.}, \bibinfo{author}{R{\'a}cz, A.} \& \bibinfo{author}{H{\'e}berger, K.}
\newblock \bibinfo{journal}{\bibinfo{title}{Why is tanimoto index an appropriate choice for fingerprint-based similarity calculations?}}
\newblock {\emph{\JournalTitle{Journal of cheminformatics}}} \textbf{\bibinfo{volume}{7}}, \bibinfo{pages}{1--13} (\bibinfo{year}{2015}).

\bibitem{preuer2018frechet}
\bibinfo{author}{Preuer, K.}, \bibinfo{author}{Renz, P.}, \bibinfo{author}{Unterthiner, T.}, \bibinfo{author}{Hochreiter, S.} \& \bibinfo{author}{Klambauer, G.}
\newblock \bibinfo{journal}{\bibinfo{title}{Fr{\'e}chet chemnet distance: a metric for generative models for molecules in drug discovery}}.
\newblock {\emph{\JournalTitle{Journal of chemical information and modeling}}} \textbf{\bibinfo{volume}{58}}, \bibinfo{pages}{1736--1741} (\bibinfo{year}{2018}).

\bibitem{smith1981identification}
\bibinfo{author}{Smith, T.~F.}, \bibinfo{author}{Waterman, M.~S.} \emph{et~al.}
\newblock \bibinfo{journal}{\bibinfo{title}{Identification of common molecular subsequences}}.
\newblock {\emph{\JournalTitle{Journal of molecular biology}}} \textbf{\bibinfo{volume}{147}}, \bibinfo{pages}{195--197} (\bibinfo{year}{1981}).

\bibitem{henikoff1992amino}
\bibinfo{author}{Henikoff, S.} \& \bibinfo{author}{Henikoff, J.~G.}
\newblock \bibinfo{journal}{\bibinfo{title}{Amino acid substitution matrices from protein blocks.}}
\newblock {\emph{\JournalTitle{Proceedings of the National Academy of Sciences}}} \textbf{\bibinfo{volume}{89}}, \bibinfo{pages}{10915--10919} (\bibinfo{year}{1992}).

\bibitem{DBLP:conf/icml/QuQSG0Z0M24}
\bibinfo{author}{Qu, Y.} \emph{et~al.}
\newblock \bibinfo{title}{Molcraft: Structure-based drug design in continuous parameter space}.
\newblock In \emph{\bibinfo{booktitle}{{ICML}}} (\bibinfo{publisher}{OpenReview.net}, \bibinfo{year}{2024}).

\bibitem{corso2024discovery}
\bibinfo{author}{Corso, G.}, \bibinfo{author}{Deng, A.}, \bibinfo{author}{Polizzi, N.}, \bibinfo{author}{Barzilay, R.} \& \bibinfo{author}{Jaakkola, T.}
\newblock \bibinfo{title}{Deep confident steps to new pockets: Strategies for docking generalization}.
\newblock In \emph{\bibinfo{booktitle}{International Conference on Learning Representations (ICLR)}} (\bibinfo{year}{2024}).

\bibitem{alhossary2015fast}
\bibinfo{author}{Alhossary, A.}, \bibinfo{author}{Handoko, S.~D.}, \bibinfo{author}{Mu, Y.} \& \bibinfo{author}{Kwoh, C.-K.}
\newblock \bibinfo{journal}{\bibinfo{title}{Fast, accurate, and reliable molecular docking with quickvina 2}}.
\newblock {\emph{\JournalTitle{Bioinformatics}}} \textbf{\bibinfo{volume}{31}}, \bibinfo{pages}{2214--2216} (\bibinfo{year}{2015}).

\bibitem{bickerton2012quantifying}
\bibinfo{author}{Bickerton, G.~R.}, \bibinfo{author}{Paolini, G.~V.}, \bibinfo{author}{Besnard, J.}, \bibinfo{author}{Muresan, S.} \& \bibinfo{author}{Hopkins, A.~L.}
\newblock \bibinfo{journal}{\bibinfo{title}{Quantifying the chemical beauty of drugs}}.
\newblock {\emph{\JournalTitle{Nature chemistry}}} \textbf{\bibinfo{volume}{4}}, \bibinfo{pages}{90--98} (\bibinfo{year}{2012}).

\bibitem{ertl2009estimation}
\bibinfo{author}{Ertl, P.} \& \bibinfo{author}{Schuffenhauer, A.}
\newblock \bibinfo{journal}{\bibinfo{title}{Estimation of synthetic accessibility score of drug-like molecules based on molecular complexity and fragment contributions}}.
\newblock {\emph{\JournalTitle{Journal of cheminformatics}}} \textbf{\bibinfo{volume}{1}}, \bibinfo{pages}{1--11} (\bibinfo{year}{2009}).

\bibitem{zhuang_2025_15303508}
\bibinfo{author}{Zhuang, X.}
\newblock \bibinfo{title}{Dataset for the paper "advancing biomolecule understanding and design following human instructions"}, \doiprefix\url{10.5281/zenodo.15303508} (\bibinfo{year}{2025}).

\bibitem{xiang_zhuang_2025_15335654}
\bibinfo{author}{Zhuang, X.}
\newblock \bibinfo{title}{Hicai-zju/instructbiomol: Version 1.0.0}, \doiprefix\url{10.5281/zenodo.15335654} (\bibinfo{year}{2025}).

\bibitem{steinegger2017mmseqs2}
\bibinfo{author}{Steinegger, M.} \& \bibinfo{author}{S{\"o}ding, J.}
\newblock \bibinfo{journal}{\bibinfo{title}{Mmseqs2 enables sensitive protein sequence searching for the analysis of massive data sets}}.
\newblock {\emph{\JournalTitle{Nature biotechnology}}} \textbf{\bibinfo{volume}{35}}, \bibinfo{pages}{1026--1028} (\bibinfo{year}{2017}).

\bibitem{du2024machine}
\bibinfo{author}{Du, Y.} \emph{et~al.}
\newblock \bibinfo{journal}{\bibinfo{title}{Machine learning-aided generative molecular design}}.
\newblock {\emph{\JournalTitle{Nature Machine Intelligence}}} \bibinfo{pages}{1--16} (\bibinfo{year}{2024}).

\bibitem{mirdita2022colabfold}
\bibinfo{author}{Mirdita, M.} \emph{et~al.}
\newblock \bibinfo{journal}{\bibinfo{title}{Colabfold: making protein folding accessible to all}}.
\newblock {\emph{\JournalTitle{Nature methods}}} \textbf{\bibinfo{volume}{19}}, \bibinfo{pages}{679--682} (\bibinfo{year}{2022}).

\bibitem{zhang2004scoring}
\bibinfo{author}{Zhang, Y.} \& \bibinfo{author}{Skolnick, J.}
\newblock \bibinfo{journal}{\bibinfo{title}{Scoring function for automated assessment of protein structure template quality}}.
\newblock {\emph{\JournalTitle{Proteins: Structure, Function, and Bioinformatics}}} \textbf{\bibinfo{volume}{57}}, \bibinfo{pages}{702--710} (\bibinfo{year}{2004}).

\bibitem{mariani2013lddt}
\bibinfo{author}{Mariani, V.}, \bibinfo{author}{Biasini, M.}, \bibinfo{author}{Barbato, A.} \& \bibinfo{author}{Schwede, T.}
\newblock \bibinfo{journal}{\bibinfo{title}{lddt: a local superposition-free score for comparing protein structures and models using distance difference tests}}.
\newblock {\emph{\JournalTitle{Bioinformatics}}} \textbf{\bibinfo{volume}{29}}, \bibinfo{pages}{2722--2728} (\bibinfo{year}{2013}).

\end{thebibliography}
% \addtocontents{toc}{\protect\setcounter{tocdepth}{1}}
% \clearpage
%TC:ignore
%\input{chapter/extend-data}
%TC:endignore
\clearpage
\end{document}